\documentclass[10pt,twocolumn,letterpaper]{article}

\usepackage{cvpr}              

\DeclareMathOperator*{\argmax}{arg\,max}

\definecolor{cvprblue}{rgb}{0.21,0.49,0.74}
\usepackage[pagebackref,breaklinks,colorlinks,allcolors=cvprblue]{hyperref}

\usepackage[accsupp]{axessibility}

\usepackage{url}
\usepackage{hyperref}

\usepackage{multirow} 
\usepackage{makecell} 

\usepackage{adjustbox}

\usepackage{colortbl}
\usepackage{xcolor}
\usepackage{color}
\usepackage{array}   

\usepackage{float}

\usepackage{nicematrix}  
\usepackage{tabularx}

\usepackage{algorithmic}
\usepackage{algorithm}
\usepackage{caption}

\usepackage{pifont}

\newcommand{\highlight}[1]{{\cellcolor[rgb]{0.925,0.957,1}}{#1}}

\definecolor{green}{rgb}{0, 0.8, 0.2}
\newcommand{\cmark}{\textcolor{green}{\ding{51}}} %

\definecolor{red}{rgb}{1.0, 0.0, 0.0}
\newcommand{\xmark}{\textcolor{red}{\ding{55}}} %

\definecolor{grey}{rgb}{0.9, 0.9, 0.9}
\newcommand{\ccol}{\cellcolor{grey}}

\definecolor{lightblue}{rgb}{0.68, 0.85, 0.9}

\definecolor{softgray}{gray}{0.96}

\definecolor{Template}{RGB}{84, 130, 53}
\definecolor{Description}{RGB}{197, 90, 17}

\definecolor{removed}{RGB}{192, 0, 0}
\definecolor{retained}{RGB}{0, 112, 192}

\def\paperID{4381} 
\def\confName{CVPR}
\def\confYear{2025}

\title{ProAPO: Progressively Automatic Prompt Optimization for Visual Classification}

\author{%
  Xiangyan Qu$^{12}$ \quad Gaopeng Gou$^{12}$\thanks{Corresponding author} \quad Jiamin Zhuang$^{12}$ \quad Jing Yu$^{3}$  \\
   Kun Song$^{4}$ \quad Qihao Wang$^{12}$ \quad Yili Li$^{12}$ \quad Gang Xiong$^{12}$ \\ 
  \small $^1$Institute of Information Engineering, Chinese Academy of Sciences \quad 
  \small $^2$School of Cyber Security, University of Chinese Academy of Sciences \\ 
  \small $^3$School of Information Engineering, Minzu University of China \quad 
  \small $^4$University of Science and Technology Beijing \\
  \tt\small \{quxiangyan, gougaopeng, zhuangjiamin\}@iie.ac.cn, jing.yu@muc.edu.cn \\
  \tt\small songkun@xs.ustb.edu.cn, wangqihao22@mails.ucas.ac.cn, \{liyili, xionggang\}@iie.ac.cn
  \vspace{-4mm}
}

\begin{document}
\maketitle

\begin{abstract}
Vision-language models (VLMs) have made significant progress in image classification by training with large-scale paired image-text data. Their performances largely depend on the prompt quality. While recent methods show that visual descriptions generated by large language models (LLMs) enhance the generalization of VLMs, class-specific prompts may be inaccurate or lack discrimination due to the hallucination in LLMs. In this paper, we aim to find visually discriminative prompts for fine-grained categories with minimal supervision and no human-in-the-loop. An evolution-based algorithm is proposed to progressively optimize language prompts from task-specific templates to class-specific descriptions. Unlike optimizing templates, the search space shows an explosion in class-specific candidate prompts. This increases prompt generation costs, iterative times, and the overfitting problem. To this end, we first introduce several simple yet effective edit-based and evolution-based operations to generate diverse candidate prompts by one-time query of LLMs. Then, two sampling strategies are proposed to find a better initial search point and reduce traversed categories, saving iteration costs. Moreover, we apply a novel fitness score with entropy constraints to mitigate overfitting. In a challenging one-shot image classification setting, our method outperforms existing textual prompt-based methods and improves LLM-generated description methods across 13 datasets. Meanwhile, we demonstrate that our optimal prompts improve adapter-based methods and transfer effectively across different backbones. Our code is available at \href{https://github.com/MorningStarOvO/ProAPO}{here}.
\end{abstract}

\vspace{-4mm}



\section{Introduction}
\label{sec:intro}

In recent years, vision-language models (VLMs)~\cite{CLIP, Align, BLIP, Flamingo, SigLIP, SLIP, EVA-01, LLaVa_Grounding} pre-trained on large-scale paired image-text data, such as CLIP~\cite{CLIP}, have shown strong generalization on various image classification tasks. These models classify images by computing the similarity between the image and the prompt (a human-readable natural language) associated with a category. The prediction is the category with the highest similarity to the query image. Their performance highly relies on prompt quality, especially in fine-grained categories. Finding the optimal prompts for the downstream task becomes an urgent challenge~\cite{CoOp, CuPL, P_N}.

Recent works have made efforts to improve prompt quality~\cite{CLIP, GPT3, prompt_influence_ACL_2021}.
\textbf{Manual prompt engineering}~\cite{CLIP, FILIP, DEFILIP} is a standard approach, writing several templates to include task-specific information, \textit{e.g.}, ``\texttt{a photo of a \{class\}, a type of bird.}'' for bird recognition. However, designing templates requires domain expertise, making it costly and challenging to scale~\cite{DCLIP, CuPL, P_N, CoOp, VDT_2023_ICCV}. Moreover, the template may lack details to recognize fine-grained categories as only the class name provides the distinct information in prompts~\cite{CoOp, VDT_2023_ICCV, CuPL}. 
\textbf{Prompt tuning} methods~\cite{CoOp, CoCoOp, PLOT, ProGrad} introduce a set of learnable tokens in the prompt to represent task-specific context, optimizing through gradient updates. While improving the performance of VLMs, they require additional training and lack interpretability~\cite{DCLIP, P_N, WaffleCLIP}. 
In contrast, \textbf{LLM-generated description} methods~\cite{DCLIP, CuPL, GPT4Vis, VDT_2023_ICCV, VDT_2023_NIPS_Hierarchical, AWT} leverage the implicit knowledge of large language models (LLMs)~\cite{GPT3, GPT4_Tech} to generate descriptions with class-specific details, enhancing generalization ability of VLMs. However, due to the hallucination in LLMs~\cite{Hallucination_in_LLM, Hallucination_in_LLM_EMNLP}, generated descriptions are suboptimal due to inaccurate, \textit{e.g.}, ``\texttt{feet}'' for the food Peking Duck, or lack discrimination for fine-grained recognition, \textit{e.g.}, the same descriptions ``\texttt{hooked bill}'' and ``\texttt{webbed feet}'' appear in Laysan Albatross and Sooty Albatross (see~\cref{fig: Problem}(a)), or exhibit non-visual descriptions, \textit{e.g.}, ``\texttt{strong smell}'' for Jackfruit. To this end, we propose the following key questions:


\begin{figure*}[t]
\centering
\includegraphics[width=0.85\linewidth]{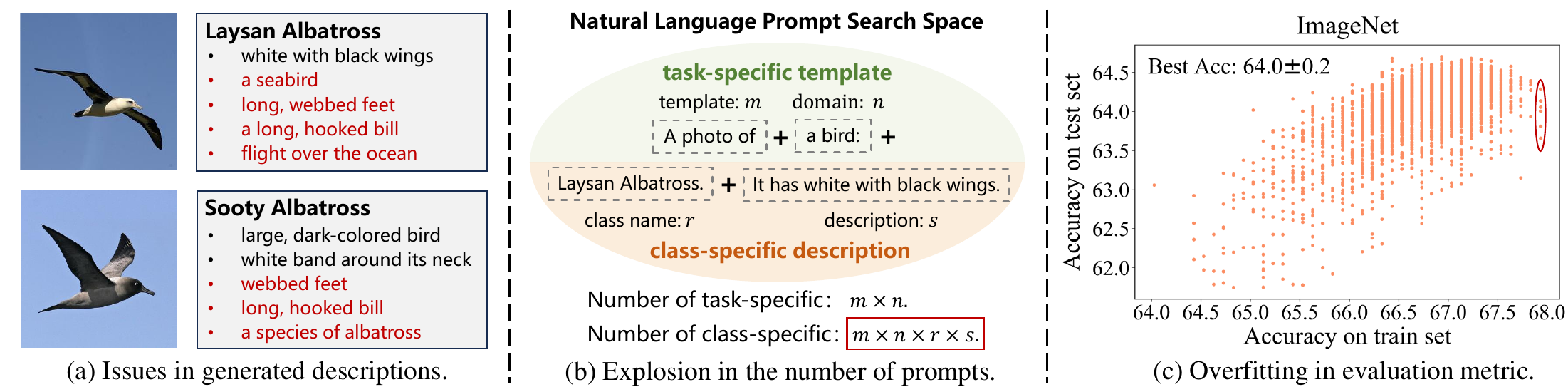}
\vspace{-7pt}
\caption{
\textbf{Issues of optimizing class-specific prompts.}
\textbf{(a)} Due to the hallucination in LLMs, generated descriptions may be inaccurate and lack discrimination between fine-grained categories (see \textbf{\textcolor{removed}{red words}}).
\textbf{(b)} Compared to task-specific templates, we see an explosion in the number of class-specific prompts (see \textbf{\textcolor{removed}{red rectangle}}). This leads to higher generation costs, iteration times, and the overfitting problem.
\textbf{(c)} Overfitting problem: Multiple candidate prompts have the same best training accuracy but variable and low test results (see \textbf{\textcolor{removed}{red circle}}).
}
\label{fig: Problem}
\vspace{-11pt}
\end{figure*}

\textit{How can we find the optimal class-specific prompts that are visually discriminative for fine-grained categories with minimal supervision and no human intervention?}

\noindent
Inspired by recent automatic prompt optimization (APO) methods~\cite{autoprompt, APE_NLP, GA_NLP, ProTeGi} in language tasks, we aim to optimize class-specific prompts by an evolution-based process, removing confused prompts while retaining discriminative ones. We consider an approximate zero-shot setting, \textit{i.e.}, one-shot classification, where minimal supervision is introduced to evaluate the prompt quality. However, previous APO methods focus solely on task-specific template optimization. In contrast, except for templates, a class-specific prompt also contains a description associated with visual details in a category, as shown in~\cref{fig: Problem}(b).
Compared with templates, class-specific optimization introduces new challenges due to the expanded search space (see~\cref{fig: Problem}(b)), which leads to:
(1) \textbf{High generation costs}. As the search space grows, it is costly and time-consuming to generate class-specific prompts exclusively with LLMs at each iteration.
(2) \textbf{Long iteration times}. Evaluating each candidate prompt in the search space may exponentially increase the iteration time, which is impractical in this scenario.
(3) \textbf{Overfitting problem}. As shown in~\cref{fig: Problem}(c), multiple candidate prompts achieve variable and low test results (from 63.6\% to 64.3\%) at the best training accuracy.







To address these issues, we propose a \textbf{Pro}gressively \textbf{A}utomatic \textbf{P}rompt \textbf{O}ptimization (ProAPO) algorithm to iteratively optimize prompts from task-specific to class-specific levels. In each iteration, we use several edited-based (\textit{i.e.}, add, remove, and replace) and evolution-based operators (\textit{i.e.}, crossover, and mutation) to generate diverse candidate prompts from a prompt library. We query LLMs to generate this library in the initialization stage. Compared to querying LLMs at each iteration, these simple operations can effectively save generation costs. Then, we introduce a novel fitness score to evaluate generated candidate prompts and retain several top-scoring ones for the next iteration generation. An entropy constraint is added to the score to increase the soft prediction score and reduce overfitting. After several iterations, we return the best prompt for classification. To reduce iteration times in class-specific descriptions, we propose a prompt sampling strategy to find a better initial point in the search space and a group sampling strategy to explore a few salient classes instead of all for optimization. Our key contributions are: 
\begin{itemize}
    \item We propose an evolution-based algorithm to progressively optimize prompts from task-specific to class-specific levels with one-shot supervision and no human intervention. It solves issues of inaccuracy and lack of discrimination in LLM-generated descriptions.
    \item We address challenges in class-specific prompt optimization by an offline generation algorithm to reduce LLM querying costs, an entropy-constrained fitness score to prevent overfitting, and two sampling strategies to find an optimal initial point and reduce iteration times.
    \item Extensive experiments on thirteen datasets show that our proposed ProAPO consistently outperforms SOTA textual prompt-based methods and improves description-based methods in a challenging one-shot supervision. Moreover, our optimal prompts improve adapter-based methods and transfer effectively across different backbones.
\end{itemize}

\section{Related Work}
\label{sec: related_work}



\noindent 
\textbf{Adaptation of VLMs for image classification tasks.} 
Inspired by successes in VLMs, recent works aim to adapt them for image classification tasks. Some works utilize lightweight linear layers~\cite{CLIP, AMU-Tuning, LP++, LFA, FD-Align, TMM_SelfAlign, T2VIndexer}, adapters~\cite{CLIP_Adapter, VDT_2023_ICCV}, visual prompting~\cite{VP, BlackVIP, VPT}, or cache models~\cite{Tip, APE, Tip-X} to enhance visual features. Other works aim to improve the quality of prompts. Manual prompt engineering~\cite{CLIP, DeClip, DEFILIP} applies task-specific information in prompt templates to enhance performance. However, templates need to be hand-written and lack fine-grained details~\cite{DCLIP, CuPL, P_N, CoOp, VDT_2023_ICCV}. Prompt tuning methods~\cite{CoOp, CoCoOp, PLOT, ProGrad, CPT, prompt_distribution, prompt_variation, TPT, MaPLe, PromptSRC, Visual_in_context_learn} learn task-specific context by a set of learnable tokens. However, they need additional training and lack interpretability~\cite{DCLIP, P_N, WaffleCLIP}. In contrast, LLM-generated description methods~\cite{DCLIP, CuPL, GPT4Vis, VDT_2023_ICCV, VDT_2023_NIPS_Hierarchical, AWT, MPVR, AdaptCLIP, EmDepart, I2MVFormer} exploit implicit knowledge in LLMs to generate visual descriptions for each category. They enrich semantics in prompts and offer interpretable predictions. In this work, we aim to further improve description quality through an evolution process.



\noindent 
\textbf{LLM-generated description methods} apply class-specific descriptions to language prompts to adapt VLMs for classification. DCLIP~\cite{DCLIP} and CuPL~\cite{CuPL} design prompts such as ``\texttt{What does a \{class\} look like?}'' to instruct LLMs to generate category descriptions. GPT4Vis~\cite{GPT4Vis} and VDT~\cite{VDT_2023_ICCV} use GPT-4~\cite{GPT4_Tech} for rich and diverse descriptions. Some work utilizes LLM-generated hierarchy labels~\cite {CHiLS} or descriptions~\cite{VDT_2023_NIPS_Hierarchical} to recognize images from coarse to fine-grained levels. However, due to the hallucination in LLMs, generated descriptions might be inaccurate, non-visual, and lack discrimination~\cite{DCLIP, WaffleCLIP, VDT_2023_NIPS_Hierarchical}. To this end, we propose ProAPO to iteratively remove ambiguous and retain discriminative prompts. Moreover, category names, ignored in previous methods, are equally crucial for improving accuracy and are introduced in our optimization.


\noindent \textbf{Automatic prompt optimization} aims to automatically find optimal prompts for language tasks, overcoming the time-consuming issue in manual prompt methods~\cite{survey_APO}. Early methods~\cite{autoprompt, RLPrompt, GrIPS} generate discrete prompts by filling templates with trigger tokens, iteratively refining prompts by mining, generation, or paraphrasing. Recently, chat-based LLMs have been applied as prompt engineers~\cite{ProTeGi, GA_NLP, APE_NLP, Promptbreeder, OPRO, EVOPrompt}, using a meta-prompt with task information to find optimal prompt design. PN~\cite{P_N} is the first to instruct LLMs to optimize templates on multimodal tasks, which feeds visual feedback of the best and worst templates to guide. However, these methods only optimize task-specific templates. iCM~\cite{iCM} is somewhat similar to ours, optimizing class-specific prompts with chat-based LLMs. However, it uses the whole validation set as supervision. In contrast, we optimize prompts with one-shot supervision and propose solutions to solve high generation costs, long iteration times, and overfitting in class-specific optimization.






\section{Method}
\label{sec: method}

\begin{figure*}[t]
\centering
\includegraphics[width=0.88\linewidth]{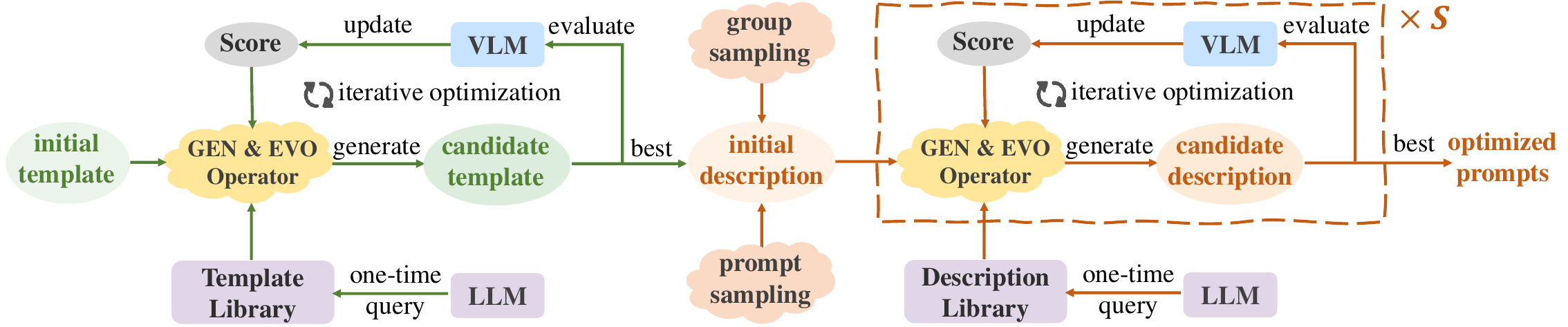}
\vspace{-0.05 in}
\caption{
\textbf{Overview of our ProAPO algorithm}. We progressively refine prompts from task-specific (\textbf{\textcolor{Template}{green lines}}) to class-specific (\textbf{\textcolor{Description}{brown lines}}) levels. Specifically, we first explore the best template by an iterative optimization process (\cref{sec: template_optim}). For each iteration, ProAPO generates a set of candidate templates by several operators (\cref{sec: prompt_generate}) and filters/refines templates by a fitness score (\cref{sec: score_function}). After several iterations, we choose the top-scoring template for description initialization. Subsequently, we introduce two sampling strategies to find a better initial point and reduce traversed categories (\cref{sec: sample_strategy}). Similar iterative optimization is then applied to class-specific descriptions.
}
\label{fig: Method}
\vspace{-10pt}
\end{figure*}


Our ProAPO algorithm is shown in~\cref{fig: Method} and summarized in Alg.~\ref{alg: ProAPO}. We first describe iterative optimization in templates (\cref{sec: template_optim}). For each iteration, candidates are generated by several operators (\cref{sec: prompt_generate}) and then evaluated by a fitness score (\cref{sec: score_function}). Afterward, similar iterative optimization is applied in descriptions, where we introduce two sampling strategies to save iteration costs (\cref{sec: sample_strategy}).


\begin{algorithm}[t]
\caption{Our \textit{Progressively Automatic Prompt Optimization} (\texttt{ProAPO}) for visual classification, which iteratively refines prompts from task-specific to class-specific. }
\label{alg: ProAPO}
\begin{algorithmic}[1]
\REQUIRE  $\mathcal{D} \leftarrow \{{(x, y)}\}_n$: training samples, $F:  \mathcal{D} \times P \to \mathbb{R}$: score function
\STATE \textbf{Initialize Template}: $\mathcal{U}_t \leftarrow \{P_0\}$ 
\STATE \textbf{Build Template Library}: $B_t \leftarrow \{ {d}_1, \cdots, {d}_n \}$ 
\STATE \textbf{Iterative Optimization}: $\mathcal{U}_t \leftarrow \texttt{APO}(\mathcal{D}, F, \mathcal{U}_t, B_t)$  
\STATE $P_t^* \leftarrow \arg \max_{P \in \mathcal{U}_t} F(\mathcal{D}, P)$
\STATE \textbf{Initialize Description}: $\mathcal{U}_c \leftarrow \{\hat{P}_0 \} $, initializing with $P_t^*$ and prompt sampling strategy
\STATE \textbf{Group Sampling}: Sample $S$ groups by class salience 
\FOR{$s = 1$ to $S$} 
    \STATE \textbf{Build Description Library}: $B_s \leftarrow \{ \hat{d}_1, \cdots, \hat{d}_q \}$
    \STATE \textbf{Iterative Optimization}: $\mathcal{U}_c \leftarrow \texttt{APO}(D, F, \mathcal{U}_c, B_s)$
\ENDFOR
\STATE ${P}^* \leftarrow \arg\max_{{P} \in \mathcal{U}_c} F(\mathcal{D}, {P})$
\RETURN candidate prompt with the highest score ${P}^*$
\end{algorithmic}
\end{algorithm}

\textbf{Preliminaries.}
Given an image, CLIP~\cite{CLIP} predicts by selecting the highest similarity between the image and category prompt. The category prompt is a human-readable natural language associated with a category, which contains a template, \textit{e.g.}, ``\texttt{a photo of a bird: \{class\}. \{description\}.}'', and class-specific descriptions, \textit{e.g.}, class name ``\texttt{Laysan Albatross}'' and its descriptions ``\texttt{It has white with black wings.}''. Some categories may use multiple prompts with varied templates or descriptions, \textit{i.e.}, prompt ensembling. We denote the prompt set that includes all categories in the task as a candidate prompt $P$. A score function $F: \mathcal{D} \times P \rightarrow \mathbb{R}$ is introduced in~\cref{sec: score_function} to evaluate the candidate prompt $P$ on a limited training set $\mathcal{D} = \{{(x, y)}\}_n$ with one-shot supervision, where $x$ is an image and $y$ is its label. Finally, we use a test set to evaluate optimized prompts.
\textbf{Our goal} is to refine prompts in the natural language space to achieve superior performance in per training sample $(x, y)$: 
\setlength{\abovedisplayskip}{3pt}
\setlength{\belowdisplayskip}{3pt}
\begin{equation}
   \argmax_P F(\mathcal{D}, P) = \argmax_P \mathbb{E}_{(x, y)} [F(\{ (x, y) \}, P)].
\end{equation}
\vspace{-15pt}
\setlength{\textfloatsep}{10pt} 
\begin{algorithm}[t]
\caption{\textit{Automatic Prompt Optimization} (\texttt{APO}) algorithm for VLMs - Lines 3 and 9 of Alg.~\ref{alg: ProAPO}, $\texttt{APO}(\mathcal{D}, F, \mathcal{U}, B)$.}
\label{alg: APO}
\begin{algorithmic}[1]
\REQUIRE  $\mathcal{D} \leftarrow \{{(x, y)}\}_n$: training samples, $F: \mathcal{D} \times P \to \mathbb{R}$: score function, $\mathcal{U}$ : candidate prompt set,  $B$: template or description library
\STATE \textbf{Initialize Evaluation Score}: $\mathcal{S} \leftarrow \{F(\mathcal{D}, P)\}_{P \in \mathcal{U}}$
\FOR{$t=1$ to $T$}
\STATE $ \mathcal{U}_{g} \leftarrow \emptyset $
\FORALL{$P \in \mathcal{U}$}
  \STATE \textbf{Edit-based Generate}: $\mathcal{U}_{g} \leftarrow \mathcal{U}_{g} \cup \texttt{GEN}(P, B) $ 
\ENDFOR
    \STATE \textbf{Evaluate}: $\mathcal{S}_{g} \leftarrow \{ F(\mathcal{D}, P') \}_{P' \in \mathcal{U}_{g}}$ 
    \STATE \textbf{Update}: $\mathcal{U} \leftarrow \{ \mathcal{U}, \mathcal{U}_{g} \}$ and $\mathcal{S} \leftarrow \{\mathcal{S}, \mathcal{S}_{g} \}$, retaining the top-$k$ of candidate prompts with high scores 
  \STATE \textbf{Evolution-based Generate}: $\mathcal{U}_e \leftarrow \texttt{EVO} (\mathcal{U}, B)$  
  \STATE \textbf{Evaluate}: $\mathcal{S}_{e} \leftarrow \{ F(\mathcal{D}, P') \}_{P' \in \mathcal{U}_{e}}$ 
  \STATE \textbf{Update}: $\mathcal{U} \leftarrow \{ \mathcal{U}, \mathcal{U}_{e} \}$ and $\mathcal{S} \leftarrow \{\mathcal{S}, \mathcal{S}_{e} \}$, retaining the top-$k$ of candidate prompts with high scores
\ENDFOR
\RETURN the latest candidate prompt set $\mathcal{U}$
\end{algorithmic}
\end{algorithm}

\subsection{Automatic Template Optimization}
\label{sec: template_optim}
To mitigate issues of semantic ambiguity caused by class names, we first iteratively optimize the template to provide better task-specific contextual information. 


\textbf{Template initialization} offers a candidate prompt $P_0$ as a starting point in the language search space (Line 1 of Alg.~\ref{alg: ProAPO}). Similar to PN~\cite{P_N}, we use ``\texttt{a photo of a \{class\}}.'' filling with class names in the dataset as $P_0$.

\textbf{Building template library} aims to provide a set of templates (Line 2 of Alg.~\ref{alg: ProAPO}) for subsequent prompt generation. We denote template library as $B_t = \{d_1, \cdots, d_n\}$, where $d$ is a single template. Similar to PN~\cite{P_N}, we can instruct LLMs with a one-time query to generate a set of templates as the library. Pre-defined templates, \textit{e.g.}, Template-80 provided in CLIP~\cite{CLIP}, can also be used. Inspired by description-based methods~\cite{WaffleCLIP, VDT_2023_ICCV}, we also supplement templates with dataset domain information generated by LLMs, such as ``\texttt{flower}'' for FLO and ``\texttt{bird}'' for CUB. 



\textbf{Iterative optimization}.
Automatic prompt optimization (APO) algorithm is introduced to refine template set $\mathcal{U}_t$ by an evolution-based process (Line 3 of Alg.~\ref{alg: ProAPO}). As summarized in Alg.~\ref{alg: APO}, each iteration of APO contains the process of: 
(1) \textbf{Generate new candidate prompts} $\mathcal{U}_g$ and $\mathcal{U}_e$ by several edit-based (Lines 4-6) and evolution-based (Line 9) operations based on the set $\mathcal{U}$ and library $B$. 
(2) \textbf{Evaluate each new candidate prompt} (Lines 7 and 10) by a score function. We regard this score as an implied ``gradient''. 
(3) \textbf{Update the prompt set} $\mathcal{U}$ based on score (Lines 8 and 11). This process retains the top-$k$ of prompts with high scores to explore the search space around the current best candidates. Finally, we return the latest candidate set $\mathcal{U}$.


\begin{algorithm}[t]
\caption{Edit-based prompt generation algorithm - Line 5 of Alg.~\ref{alg: APO}, $\texttt{GEN}(P, B)$.}
\label{alg: prompt_generate}
\begin{algorithmic}[1]
\REQUIRE  $P \leftarrow \{ \hat{d}_1, \cdots, \hat{d}_i \}$: a candidate prompt, $B \leftarrow \{ d_1, \cdots, d_j \}$: template or description library
\STATE $ \mathcal{U}_{g} \leftarrow \emptyset $
\FOR{$m = 1$ to $M$} 
    \STATE $d_{add} \leftarrow \texttt{Select}(B)$ 
    \STATE $P_{add} \leftarrow P \cup \{ d_{add} \} $ \hfill $\triangleright$ Add
    \STATE $d_{del} \leftarrow \texttt{Select}(P)$ 
    \STATE $P_{del} \leftarrow P \setminus \{ d_{del} \} $ \hfill $\triangleright$ Delete
    \STATE $d_{in} \leftarrow \texttt{Select} (B)$
    \STATE $d_{out} \leftarrow \texttt{Select} (P)$ 
    \STATE $P_{rep} \leftarrow (P \cup  \{ d_{in} \} ) \setminus \{ d_{out} \} $ \hfill $\triangleright$ Replace
    \STATE $\mathcal{U}_g \leftarrow \mathcal{U}_g \cup \{ P_{add}, P_{del}, P_{rep} \}$ 
\ENDFOR
\RETURN the generated candidate prompt set $\mathcal{U}_g$
\end{algorithmic}
\end{algorithm}

\subsection{Prompt Generation by Several Operators}
\label{sec: prompt_generate}

Due to the explosion of class-specific prompts, it is costly and time-intensive to generate new prompts exclusively with LLMs like previous methods~\cite{P_N, iCM}. To this end, we introduce edit-based and evolution-based operators to generate diverse candidate prompts from the library. 

\textbf{Edit-based generation} is introduced to create a set of new candidate prompts by several simple arithmetic operators. As shown in Alg.~\ref{alg: prompt_generate}, we iteratively operate a candidate prompt $P$ with the library $B$, where $\hat{d}$ in $P$ and $d$ in $B$ denotes a single template or description. The $\texttt{Select}(\cdot)$ operator samples an element from a given prompt set. Three operations are then applied to $P$ in each iteration: (1) \textbf{Add} a new element $d_{add}$ to $P$. (2) \textbf{Remove} an existing element $d_{del}$ from $P$. (3) \textbf{Replace} an element $d_{out}$ in $P$ with a new element $d_{in}$ in $B$. These new candidates are ensembled around the current high-score candidate $P$, which makes them more likely to succeed. Moreover, selecting elements from the library ensures differences between generated candidates. After $M$-times steps, we return the latest set $\mathcal{U}_g$.

\textbf{Evolution-based generation} is introduced to improve search efficiency over random steps, as shown in Alg.~\ref{alg: prompt_evolution}. Inspired by widely used Generic Algorithm~\cite{GA_1975, GA_1992, GA_1998}, we introduce crossover and mutation operators to enhance candidate generation. \textbf{Crossover} operator aims to find the optimal direction quickly by combining high-scoring candidate prompts. We randomly take the concatenation of two candidates $P_{c1}$ and $P_{c2}$ sampled from the current optimal set $\mathcal{U}$, yielding a new candidate $P_c$. \textbf{Mutation} operator aims to prevent convergence to local optima by introducing variations. We randomly add new elements selected from library $B$ into the candidate $P_c$, yielding mutation candidate $P_m$. After $N$-times steps, we return the generated set $\mathcal{U}_e$.




\begin{algorithm}[t]
\caption{Evolution-based generation algorithm - Line 9 of Alg.~\ref{alg: APO}, $\texttt{EVO}(\mathcal{U}, B)$.}
\label{alg: prompt_evolution}
\begin{algorithmic}[1]
\REQUIRE $\mathcal{U} \leftarrow \{ P_1, \cdots, P_n \}$: candidate prompt set, $B \leftarrow \{ d_1, \cdots, d_j \}$: template or description library
\STATE $\mathcal{U}_e \leftarrow \emptyset $
\FOR{$n = 1$ to $N$} 
\STATE $P_{c1} \leftarrow \texttt{Select}(\mathcal{U})$
\STATE $P_{c2} \leftarrow \texttt{Select}(\mathcal{U})$
\STATE $P_{c} \leftarrow P_{c1} \cup P_{c2}$ \hfill $\triangleright$ Crossover
\STATE $P_{all} \leftarrow B \cup P_c$
\STATE $P_m \leftarrow \textsc{RandomSelect}(P_{all}, \text{len}(P_c) )$\hfill $\triangleright$ Mutation
\STATE $\mathcal{U}_e \leftarrow \mathcal{U}_e \cup \{ P_c, P_m \}$
\ENDFOR
\RETURN the generated candidate prompt set $\mathcal{U}_e$
\end{algorithmic}
\end{algorithm}

\subsection{Score Functions}
\label{sec: score_function}

To measure candidate prompts, we introduce a fitness score to approximately obtain the ``gradient'' used for optimization. It contains the accuracy and an entropy constraint.

\textbf{Accuracy.}
Given an image $x$, we obtain a prediction $\text{pred}(x)$ that yields the highest cosine similarity:
\setlength{\abovedisplayskip}{3pt}
\setlength{\belowdisplayskip}{3pt}
\begin{align}
    s(x, c) &= \frac{1}{|D(c)|} \sum_{d \in D(c)} \text{cos}(I(x), T(d)). \\
    \text{pred}(x) &= \argmax_{c \in C} s(x, c),
\end{align}
where $D(c) \in P$ is a set with all prompts describing the category $c$ in candidate prompt $P$, and $C$ is the entire category in the dataset. $I$ and $T$ are image encoder and text encoder, respectively. The accuracy is formulated as follows:
\begin{equation}
    \text{Acc} = \mathbb{E}_{(x, y) \in \mathcal{D}} [\mathbb{I}(\text{pred}(x) = y)], 
\end{equation}
where $\mathcal{D}$ is the training set and $\mathbb{I}(\cdot)$ is an indicator function.

\textbf{Entropy constrain.} Previous methods~\cite{P_N, iCM} only use accuracy as the evaluation metric. However, the overfitting problem appears as shown in~\cref{fig: Problem}(c). To this end, we include a simple entropy constrain, which penalizes the model to predict a higher probabilistic score in the true label $y$:
\begin{equation}
    H = \mathbb{E}_{(x, y) \in \mathcal{D}}[\log(s(x, y))].
\end{equation}
The final score is formulated as follows: 
\begin{equation}
    F(\mathcal{D}, P) = \text{Acc} + \alpha H.
    \label{eq: score_function}
\end{equation}
where $\alpha$ is a scalar to balance them. In~\cref{sec: ablation_study}, we empirically show that this score effectively reduces overfitting.



\subsection{Sampling Strategy for Initialization}
\label{sec: sample_strategy}
After exploring the best template, we continue to optimize the class-specific descriptions. To save iteration costs, we introduce a prompt sampling strategy to find a better start point and a grouping sampling strategy to select some salient categories instead of all for iterative optimization.


\textbf{Prompt sampling strategy} aims to initialize the class-specific candidate prompt $\hat{P}_0$ with the optimized template $P_t^*$ and high-score descriptions (Line 5 of Alg.~\ref{alg: ProAPO}). Similar to description-based methods~\cite{DCLIP, CuPL}, we instruct LLM to generate a set of visual descriptions for each class. Besides, we replace the class name with its synonyms generated by LLMs to increase the number of descriptions. To obtain a better initial point, we randomly sample descriptions of each category to obtain multiple candidate prompts and select the one with the highest score as $\hat{P}_0$. It ensures that subsequent optimization is around candidates with relatively high scores. More details are shown in~\cref{sec_supp: prompt_sampling}.




\textbf{Group sampling strategy} aims to explore several salient categories into groups for optimization (Line 6 of Alg.~\ref{alg: ProAPO}). We consider two ways to sample categories. First, we select the groups with the lowest top-$n_{wst}$ accuracy and its misclassified categories. Moreover, we also choose salient groups by the category with the top-$n_{sln}$ result gains after adding descriptions and its misclassified categories. In the end, We use $S = n_{wst} + n_{sln}$ groups for optimization. More details are shown in~\cref{sec_supp: group_sampling}. In~\cref{sec: ablation_study}, experiments show that optimizing these selected categories can effectively improve performance and save costs.





\textbf{Building Library and Iterative Optimization for Descriptions.} In Lines 7-10 of Alg.~\ref{alg: ProAPO}, we iteratively optimize the class-specific description in selected groups. Similar to template optimization, we first build the description library, which contains visual descriptions for categories in the specific group. Then, we apply \texttt{APO} algorithm to refine descriptions automatically. After several iterations, we use the candidate $P^*$ with the highest score as the final prompt.

\section{Experiments}
\label{sec: exp}

\noindent 
\textbf{Datasets.}
Following prompt tuning~\cite{CoOp, CoCoOp} and LLM-generated description works~\cite{DCLIP}, we evaluate our ProAPO on thirteen downstream tasks, including ImageNet-1K (IN-1K)~\cite{Imagenet}, Caltech101~\cite{caltech101} (object recognition), StandfordCars~\cite{Cars}, CUB200-2011~\cite{CUB} (bird classification), DTD~\cite{DTD} (Textures), EuroSAT (ESAT)~\cite{EuroSAT} (satellite images), FGVCAircraft~\cite{FGVC}, Flowers102~\cite{FLO}, Food101~\cite{Food101}, OxfordPets~\cite{oxford_pets}, Places365~\cite{Places365}, SUN397~\cite{SUN} (scene recognition), UCF101~\cite{UCF101} (human action).

{
\begin{table*}[t]
  \centering
  \resizebox{0.86\linewidth}{!}
    {
    \begin{tabular}
        {l | c | ccccc ccccc ccc | c | c}
            
        \toprule
        \textbf{Module} & \textbf{TF} & \rotatebox{90}{\textbf{IN-1K}} & \rotatebox{90}{\textbf{Caltech}} & \rotatebox{90}{\textbf{Cars}} & \rotatebox{90}{\textbf{CUB}} & \rotatebox{90}{\textbf{DTD}}  & \rotatebox{90}{\textbf{ESAT}} & \rotatebox{90}{\textbf{FGVC}} & \rotatebox{90}{\textbf{FLO}} & \rotatebox{90}{\textbf{Food}}  &  \rotatebox{90}{\textbf{Pets}} & \rotatebox{90}{\textbf{Places}} & \rotatebox{90}{\textbf{SUN}} & \rotatebox{90}{\textbf{UCF}} & \rotatebox{90}{\textbf{Avg (11)}} & \rotatebox{90}{\textbf{Avg (13)}} \\
        \midrule

         \multicolumn{17}{c}{\textit{\textbf{\ccol{ResNet-50 Backbone}}}} \\
         \midrule
        CLIP (a photo of a \{\}) & \cmark & 57.9 & 84.5 & 53.9 & 44.7 & 38.8 & 28.6 & 15.9 & 60.2 & 74.0 & 83.2 & 38.2 & 58.0 & 56.9 & 55.6 & 53.4 \\ 

        \midrule

        \multicolumn{17}{l}{\textit{\textbf{prompt tuning methods}}} \\
        {CoOp}~\cite{CoOp} & \xmark & {57.2} & {87.5} & {55.6} & {-} & {44.4} & {50.6} & {9.6} & {68.1}  & {74.3} & {85.9} & {-} & {60.3} & {61.9} & {59.6} & {-} \\ 
        {PLOT}~\cite{PLOT} & \xmark & 59.5 &  \underline{89.8} & 56.6 & - & \underline{46.6} & \underline{54.1} & 17.9 & 71.7 & 77.7 & 87.5 & - & \underline{62.5} & 64.5  & 62.6 & - \\
        {ProGrad}~\cite{ProGrad} & \xmark & {57.8} & {88.7} & {\textbf{58.4}} & {-} & {46.1} & {\textbf{56.3}} & \underline{18.8} & {\underline{73.2}} & {76.0} & {\underline{88.4}} & {-} & {60.5} & \underline{65.6} & {62.7} & {-} \\
        
        \midrule

        \multicolumn{17}{l}{\textit{\textbf{automatic prompt optimization methods}}} \\
        
        PN~\cite{P_N} &  \cmark & \underline{59.6} & 89.1 & 56.2 & - & 44.8 & 49.0 & 18.1 & 67.2 & \underline{78.3} & 88.1 & - & 61.0 & 60.2 & 61.1 & - \\

        \highlight{\textbf{ProAPO} (ours)} & \highlight{\cmark} & \highlight{\textbf{61.5}} & \highlight{\textbf{90.3}} & \highlight{\underline{58.0}} & \highlight{\textbf{50.7}} & \highlight{\textbf{52.3}} & \highlight{51.7} & \highlight{\textbf{21.1}} & \highlight{\textbf{75.1}} & \highlight{\textbf{81.8}} & \highlight{\textbf{88.7}} & \highlight{\textbf{41.8}} & \highlight{\textbf{63.7}} & \highlight{\textbf{66.0}} & \highlight{\textbf{64.6}}  & \highlight{\textbf{61.8}} \\
        
        \midrule
        
        \multicolumn{17}{c}{\textit{\textbf{\ccol{ViT-B/32 Backbone}}}} \\
        
        \midrule

        CLIP (a photo of a \{\}) & \cmark & 62.1  & 91.2  & 60.4  & 51.7 & 42.9  & 43.9  & 20.2  & 66.0  & 83.2  & 86.8 & 39.9 & 62.1  & 60.9 & 61.8 & 59.3 \\ 

        \midrule

        \multicolumn{17}{l}{\textit{\textbf{hand-engineered methods}}} \\

        
        Template-80~\cite{CLIP} & \cmark & 63.5 & 91.6 & 60.4 & 51.2 & 42.8 & 52.6 & 19.5 & 66.1 & 84.2 & 87.4 & 41.6 & 63.5 & 62.9 & 63.1  & 60.6  \\ 
        FILIP-8~\cite{FILIP} & \cmark & 63.8 & 91.4 & 60.7 & 52.7 & 43.4 & 54.3 & 18.9 & 67.0 & 84.6 & 87.5 & 41.2 & 63.9 & 65.0 & 63.7  & 61.1 \\ 
        DEFILIP-6~\cite{DEFILIP} & \cmark & 62.5 & 91.0 & 59.9 & 51.1 & 41.3 & 46.4 & 18.8 & 66.5 & 84.3 & 87.5 & 40.2 & 62.3 & 63.6 & 62.2  & 59.6 \\      

        \midrule
        \multicolumn{17}{l}{\textit{\textbf{description-based methods}}} \\

        DCLIP~\cite{DCLIP} & \cmark & 63.3  & 92.7  & 59.4  & 52.7  & 44.1  & 38.4  & 19.4  & 66.1  & 83.9  & 88.1  & 41.2  & 65.0  & 65.8  & 62.4  & 60.0 \\
        Waffle~\cite{WaffleCLIP} & \cmark & 63.3 & 92.1 & 59.3 & 52.9 & 43.2 & 51.6 & 19.6 & 66.3 & \underline{84.9} & 87.7 & 41.5 & 65.0 & 64.5 & 63.4 & 60.9 \\ 
        CuPL~\cite{CuPL} &\cmark & \underline{64.4}  & 92.9  & 60.7  & 53.3  & \underline{50.6}  & 50.5  & 20.9  & 69.5  & 84.2  & 87.0  & \underline{43.1}  & \underline{66.3}  & 66.4  & 64.9  & 62.3 \\
        GPT4Vis~\cite{GPT4Vis} & \cmark & 63.5  & 93.1  & \underline{61.4}  & 52.7  & 48.5  & 47.0  & 21.4  & 69.8  & 84.3  & 88.1  & 42.7  & 64.2  & 65.7  & 64.3  & 61.7 \\ 
        AdaptCLIP~\cite{AdaptCLIP} & \cmark & 63.3  & 92.7  & 59.7  & 53.6  & 47.4  & 51.3  & 20.8  & 67.2  & 84.2  & 87.6  & 41.9  & 66.1  & 66.5  & 64.2  & 61.7 \\

        \midrule
        \highlight{\textbf{ProAPO} w/ DCLIP} & \highlight{\cmark} & \highlight{64.1} & \highlight{\underline{93.2}} & \highlight{60.6} & \highlight{\underline{53.6}} & \highlight{48.2} & \highlight{\underline{59.4}} & \highlight{\underline{22.6}} & \highlight{\underline{71.5}} & \highlight{84.2} & \highlight{\underline{88.7}} & \highlight{42.7} & \highlight{66.0} & \highlight{\underline{68.0}} & \highlight{\underline{66.0}}  & \highlight{\underline{63.3}} \\
        
        \highlight{\textbf{ProAPO} (ours)} & \highlight{\cmark} & \highlight{\textbf{64.7}} & \highlight{\textbf{94.4}} & \highlight{\textbf{61.7}} & \highlight{\textbf{55.4}} & \highlight{\textbf{53.5}} & \highlight{\textbf{63.0}} & \highlight{\textbf{23.0}} & \highlight{\textbf{74.3}} & \highlight{\textbf{85.3}} & \highlight{\textbf{91.0}} & \highlight{\textbf{43.3}} & \highlight{\textbf{66.6}} & \highlight{\textbf{69.0}} & \highlight{\textbf{67.9}}  & \highlight{\textbf{65.0}} \\

        
        \bottomrule
    \end{tabular}
}
\vskip -6pt
  \caption{\textbf{Comparison of our ProAPO with SOTA textual prompt-based methods.} We report the top-1 accuracy (\%) on the test set. The best and second best results of the same backbone for each dataset are \textbf{bolded} and \underline{underlined}, respectively. \textbf{Avg (11)} and \textbf{Avg (13)} denote average results across 11 datasets (excluding CUB~\cite{CUB} and Places~\cite{Places365}) and all 13 datasets. \textbf{TF} denotes training-free approaches.}
  \label{tab: main_result}
  \vskip -0.15in
\end{table*}
}

\noindent 
\textbf{Implementation details.}
In the default setting, we use template-80 pre-defined in CLIP~\cite{CLIP} as the template library and CuPL~\cite{CuPL} as the description library for searching the hyperparameters in a fixed language space. Dataset domain and synonym labels are generated by LLMs. This setting ensures a one-time query of LLMs, and no human intervention is required. If not explicitly stated, we set iteration times $T=4$ in both template and description optimization, generated number $M=N=8$, $\alpha = 1e3$, $n_{wst} = n_{sln} = \text{log}(|C|)$ for all datasets, where $|C|$ is the number of classes. All results are average with four seeds. Besides, our ProAPO is implemented in PyTorch and runs with an RTX 3090 GPU. More details are shown in Supp.4.




\subsection{Comparison with SOTA Methods}

\noindent \textbf{Compared methods.}
We compare our results with state-of-the-art (SOTA) textual prompt-based methods, including vanilla CLIP with ``\texttt{a photo of a \{\}}'' template, prompt tuning methods~\cite{CoOp, PLOT, ProGrad}, hand-engineered methods (\textit{i.e.}, best templates released by~\cite{CLIP, FILIP, DEFILIP}), LLM-generated description methods~\cite{CuPL, DCLIP, WaffleCLIP, GPT4Vis, AdaptCLIP}, and automatic prompt optimization methods~\cite{P_N}. We test our ProAPO on two popular backbones (ResNet50~\cite{ResNet} and ViT-B/32~\cite{ViT}). Prompt tuning and automatic prompt optimization methods are evaluated under one-shot supervision.


\textbf{Results.}
In~\cref{tab: main_result}, we see that our ProAPO consistently outperforms previous prompt-based methods across diverse datasets on ResNet50 and ViT-B/32 backbones. Our optimized prompts improve vanilla CLIP by an average of 8.4\% (from 53.4\% to 61.8\%) on ResNet50 and 5.7\% (from 59.3\% to 65.0\%) on ViT-B/32. Moreover, our training-free ProAPO surpasses gradient-based prompt-tuning methods (CoOp, PLOT, ProGrad) by at least 1.9\% average accuracy in eleven datasets. It shows optimizing prompts in a natural language space is more effective in low-shot tasks. Compared to template-optimized methods (\textit{i.e.}, PN and hand-engineered methods), our ProAPO achieves remarkable performance on fine-grained datasets, \textit{e.g.}, CUB, FLO, ESAT, and UCF. This is because class-specific descriptions provide fine-grained discriminative details. Our ProAPO also outperforms description-based methods, even with DCLIP as initialization. We improve performances of DCLIP and CuPL by 3.3\% and 2.7\% on average across thirteen datasets. Notably, it improves DCLIP and CuPL by 5.4\% and 4.8\% on FLO and by 21\% and 12.5\% on ESAT, confirming that iterative optimization of class-specific prompts enhances fine-grained recognition.



\textbf{Critical analysis.}
On datasets like ImageNet~\cite{Imagenet}, Caltech~\cite{caltech101}, Places~\cite{Places365}, and SUN~\cite{SUN}, performance gains are relatively small. We attribute this to two reasons: First, datasets like ImageNet and Caltech contain coarse-grained categories. The differences between categories have become clearer with LLM-generated descriptions, eliminating the need for our method. Second, fine-grained datasets, such as Places and SUN for scene recognition, are easily confusing between different images, limiting performance from only textual side optimization. In~\cref{sec: more_benefit}, we show that adapter-based methods with our optimized prompt improve performance again after addressing issues of image confusion. Moreover, we observe that the language search space impacts performance. Since CuPL has better descriptions than DCLIP, our ProAPO with CuPL performs well.



{
\begin{figure}[t]
\centering
{
    \hfill
    \subfloat[Training-free methods.]{\includegraphics[width=0.23\textwidth]{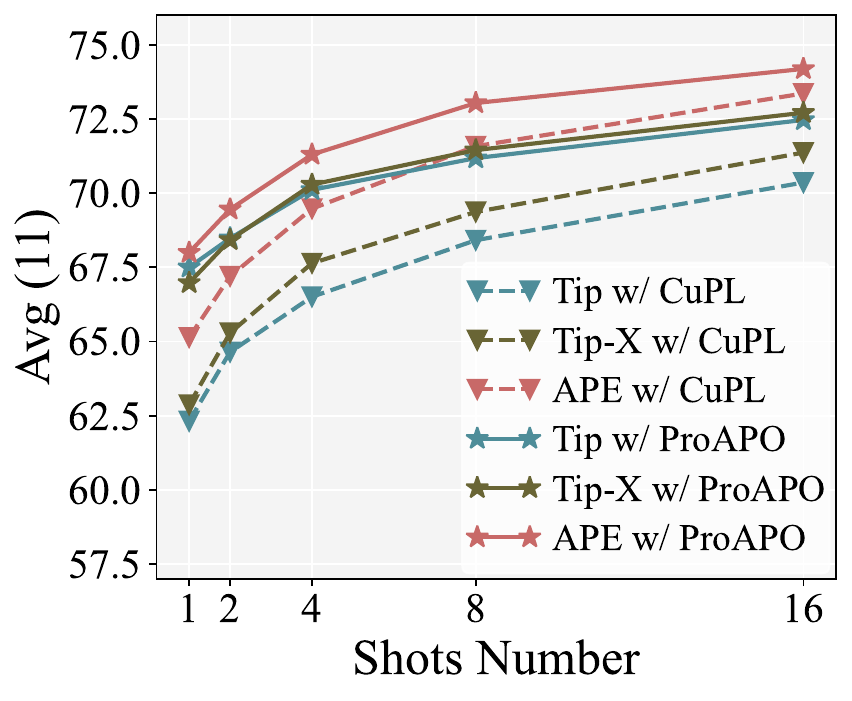}}
    \hfill
    \subfloat[Fine-tuning methods.]{\includegraphics[width=0.23\textwidth]{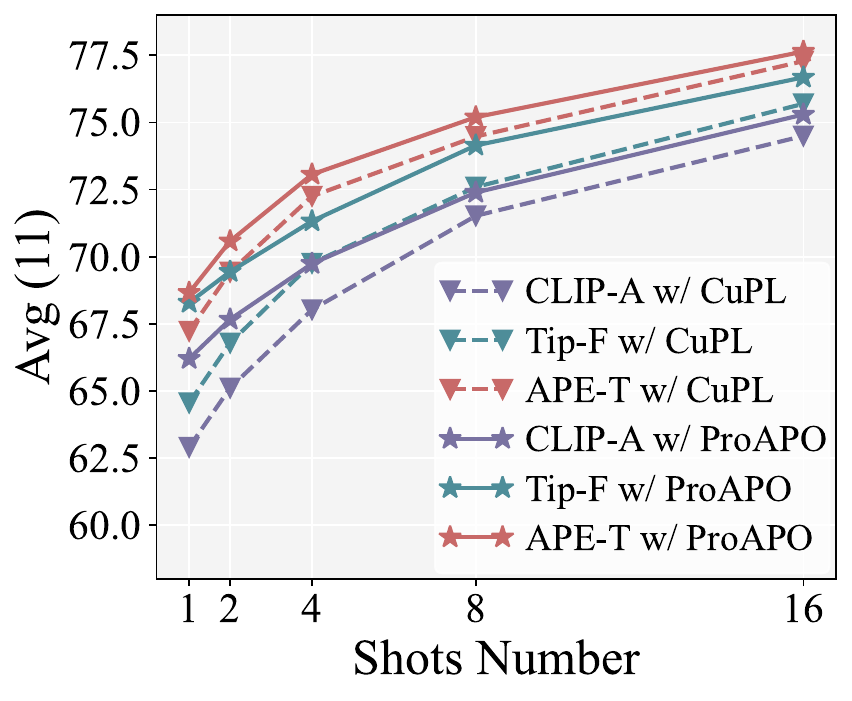}}
}
\vspace{-7pt}
\caption{\textbf{Results of adapter-based methods with different initial prompts.} Solid and dotted lines denote prompt initialization with ProAPO and CuPL, respectively.
}
\label{fig: improve_adapter_based}
\vspace{-5pt}
\end{figure}
}

\subsection{More Benefits by Optimal Prompts}
\label{sec: more_benefit}


\noindent 
\textbf{Transfer to adapter-based methods.}
In~\cref{fig: improve_adapter_based}, we show the results of popular adapter-based methods~\cite{Tip, Tip-X, APE, CLIP_Adapter} with different prompt initialization, \textit{i.e.}, SOTA method CuPL~\cite{CuPL} and our ProAPO. Adapter-based methods with ProAPO (solid lines) consistently surpass those with CuPL (dotted lines). It reveals that high-quality prompts make adapters perform better. Even in low shots, training with ProAPO achieves notable performance gains, which verifies its effectiveness. As the number of shots increases, adapters further improve results by enhancing image features.


{
\setlength{\tabcolsep}{4pt}
\begin{table}[t]
  \centering
  \resizebox{0.81\linewidth}{!}
    {
    \begin{tabular}
        {l | c | c | ccc }

        \toprule
         &  & {\textbf{Source}} & \multicolumn{3}{c}{\textbf{Target}} \\
        \cmidrule(lr){3-3} \cmidrule(lr){4-6}
        \textbf{Module} & \textbf{Shots} & RN50 & RN101 & ViT-B/32 & ViT-B/16  \\
        \midrule
        CLIP~\cite{CLIP} & 0 &  57.9 & 60.6 & 61.9 & 66.6  \\
        CoOp~\cite{CoOp} & 16 & \textbf{63.0} & 20.6 & 31.7 & 39.5 \\
        PN~\cite{P_N} & 1 & 59.9 & 60.7 & 62.2 & 67.0 \\
       \highlight{\textbf{ProAPO}} & \highlight{1} & \highlight{61.5} & \highlight{\textbf{62.1}} & \highlight{\textbf{64.6}} & \highlight{\textbf{69.9}} \\
        \bottomrule
    \end{tabular}
}
\vspace{-4pt}
\caption{\textbf{Results of prompt transfer from ResNet-50 to other architectures.} We report the top-1 accuracy in ImageNet-1K.}
\vspace{-8pt}
\label{tab: transfer_backbone}
\end{table}
}

\begin{figure}[t]
\centering
\includegraphics[width=0.83\linewidth]{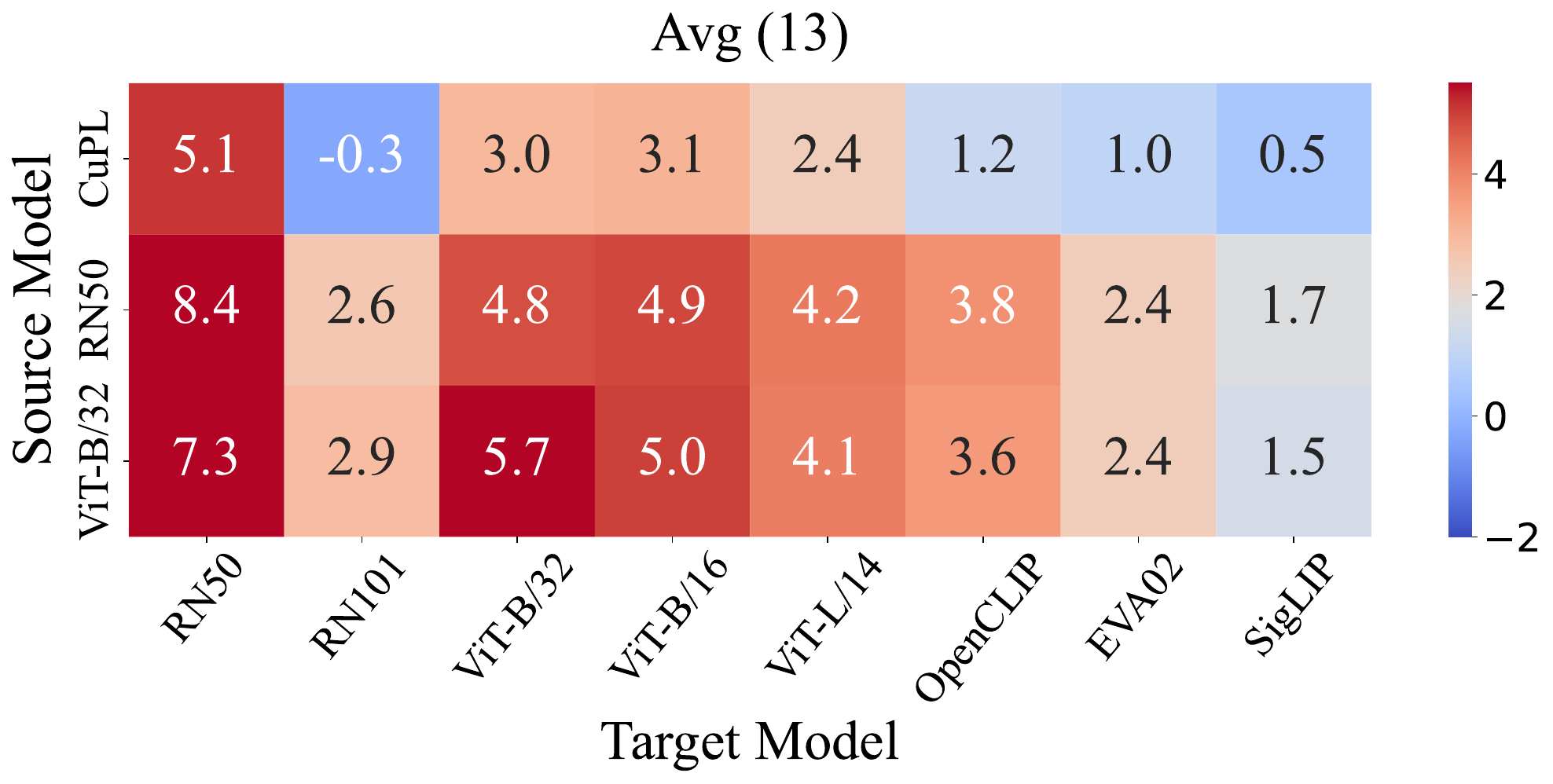}
\vspace{-8pt}
\caption{\textbf{Results of prompt transfer to different backbones.} The value denotes performance gains compared to vanilla VLMs. Our optimized prompts of ResNet50 and ViT-B/32 are reported.
}
\label{fig: transfer_to_backbones}
\vspace{-6pt}
\end{figure}

\noindent 
\textbf{Transfer to different backbones.}
In~\cref{tab: transfer_backbone}, we report accuracy in ImageNet across different backbones, with prompts optimized on a source backbone (ResNet50) adapted to target backbones (ResNet101, ViT-B/32, ViT-B/16).
We observe that CoOp~\cite{CoOp} obtains a significant drop in accuracy on target backbones while our ProAPO maintains performance. It verifies that discrete prompts searched in natural language spaces transfer better than continuous prompts.
ProAPO also outperforms PN, which reveals the effectiveness of class-specific optimization. Moreover, we achieve stable performance gains compared to CuPL~\cite{CuPL} from source to target models in~\cref{fig: transfer_to_backbones}. It further verifies that ProAPO transfers easily across different backbones.



{
\begin{figure}[t]
\centering
{
    \hfill
    \subfloat[]{\includegraphics[height=0.16\textwidth]{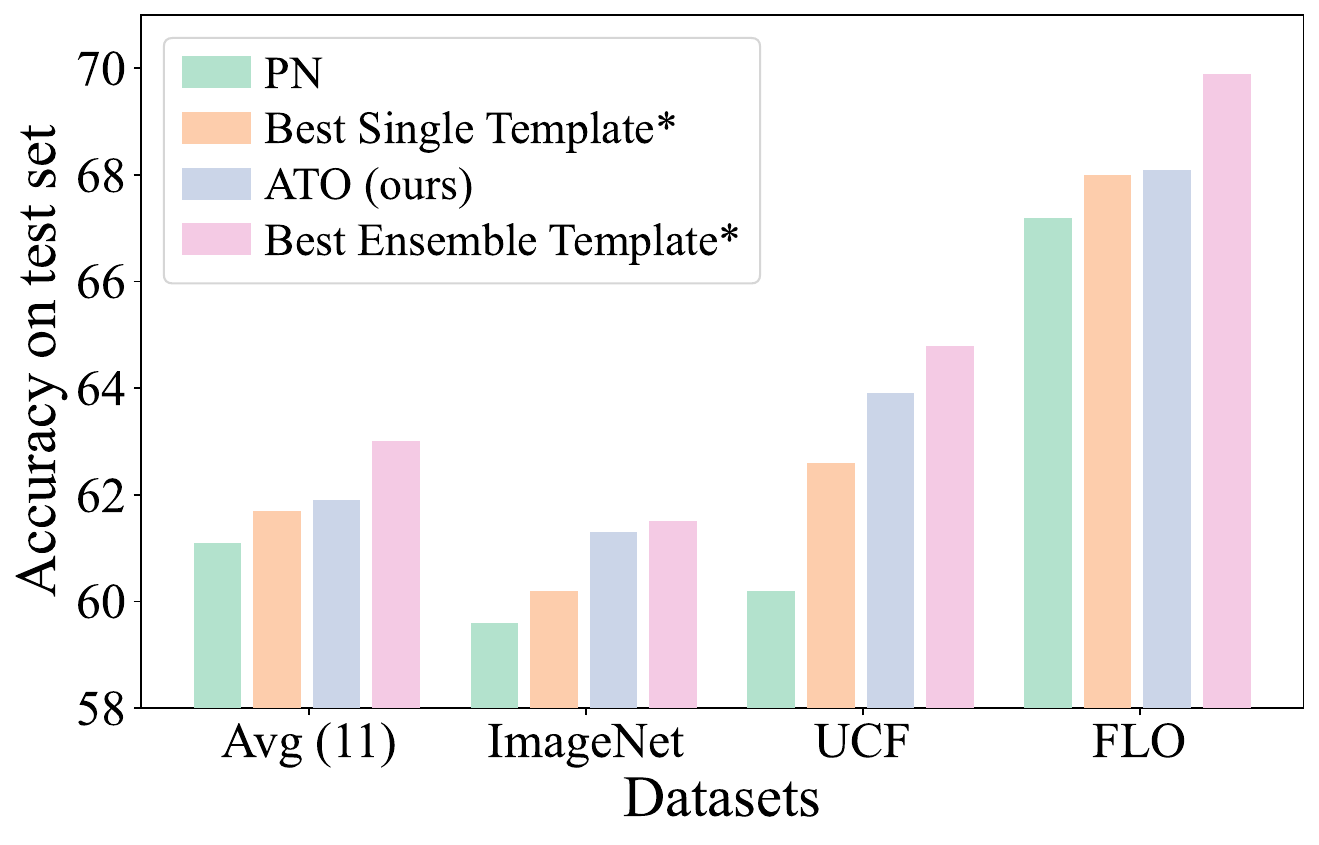}}
    \hfill
    \subfloat[]{\includegraphics[height=0.16\textwidth]{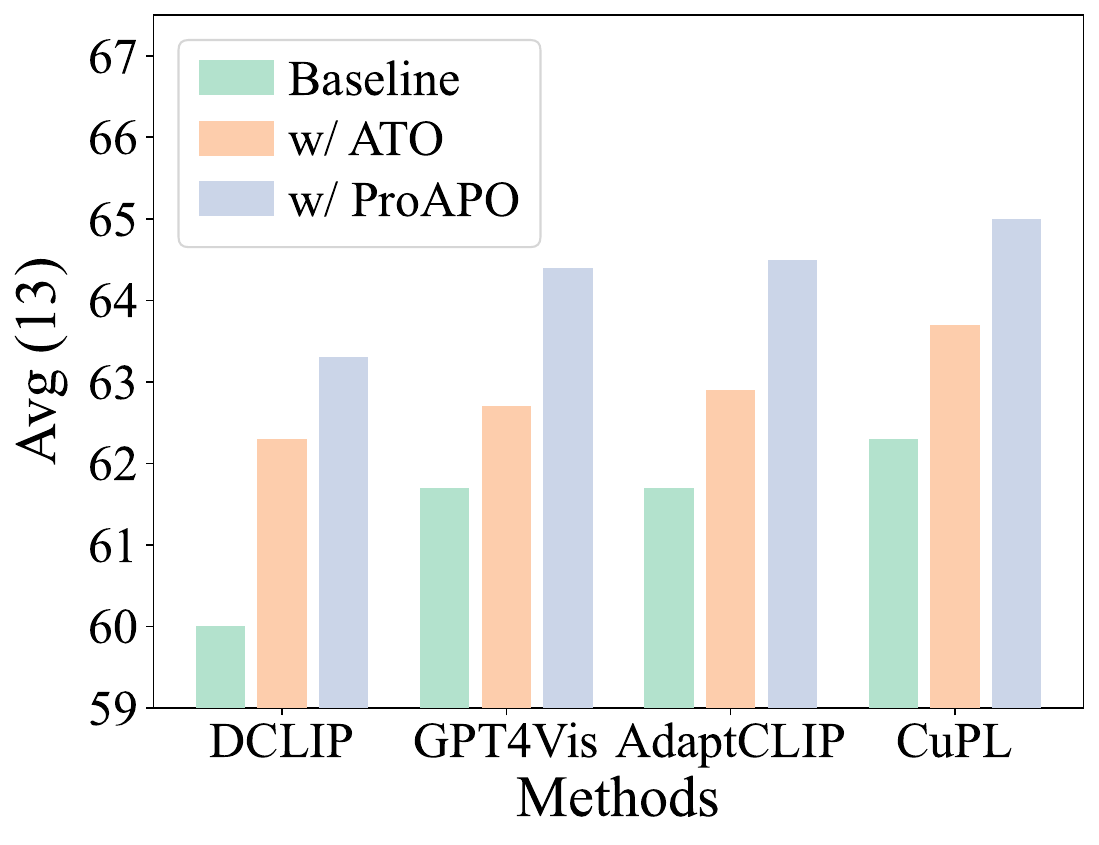}}
}
\vspace{-10pt}
\caption{\textbf{Performance improvement analysis.} (a) Analysis of the effect of single vs. ensemble prompts. * denotes results evaluated in the test set. ATO is our automatic template optimization algorithm. (b) Results of previous description-based methods with prompt optimization by our ATO and ProAPO algorithms.}
\label{fig: improve_description_methods}
\vspace{-5pt}
\end{figure}
}

\begin{figure}[t]
\centering
\includegraphics[width=1.0\linewidth]{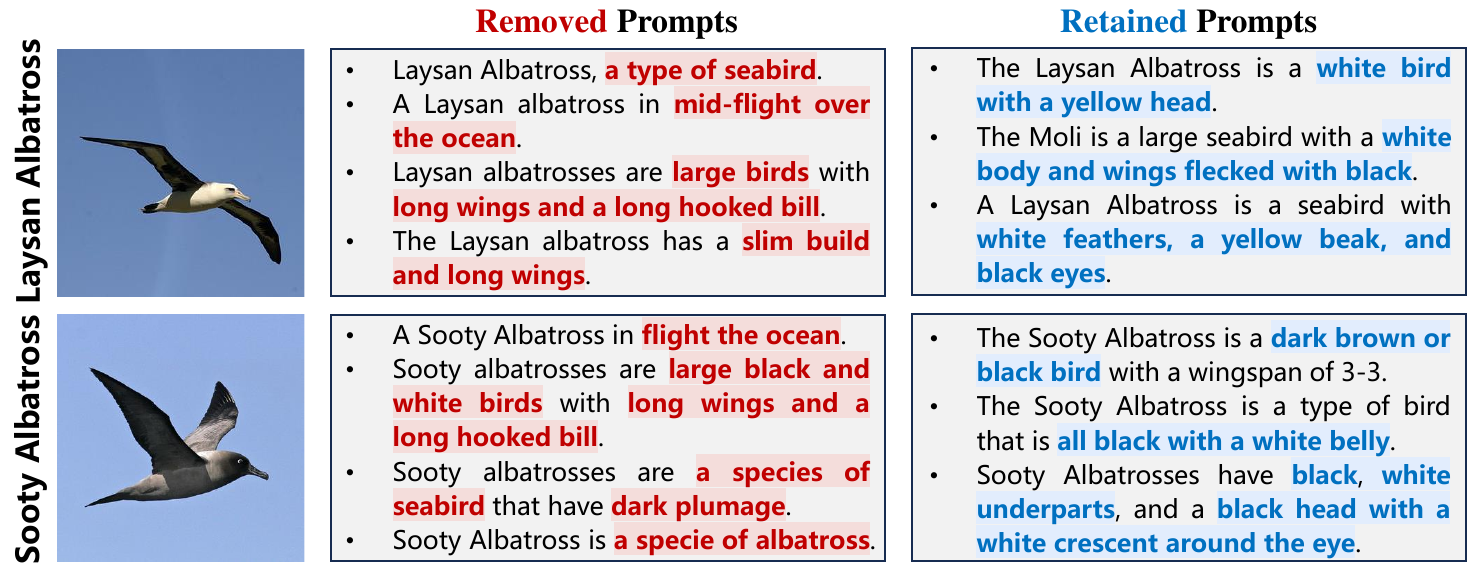}
\vspace{-6pt}
\caption{\textbf{Qualitative analysis of class-specific prompt optimization by ProAPO.} Shaded \textbf{\textcolor{removed}{red}} and \textbf{\textcolor{retained}{blue}} words denote common and discriminative descriptions in two confused categories.
}
\label{fig: qualitative_result}
\vspace{-6pt}
\end{figure}

\subsection{Performance Improvement Analysis}
In this section, we analyze the key reasons for the performance improvement of our ProAPO.

\textbf{Prompt ensembling is better than a single prompt.}
Compared to PN~\cite{P_N}, we utilize prompt ensembling instead of a single prompt to optimize the template and description. To evaluate the effectiveness of prompt ensembling, we use Template-80~\cite{CLIP} as the template library and denote the prompts searched by the test set as the upper bound. As shown in~\cref{fig: improve_description_methods}(a), we observe that ensemble templates have a higher upper bound than the single template, consistent with prior work~\cite{FILIP, DEFILIP, DCLIP}. Similarly, we observe that our optimized templates achieve higher performance than PN~\cite{P_N}, even better than the best single template, further verifying the effectiveness of our method.




\textbf{Iterative optimization improves prompt quality.}
In \cref{fig: qualitative_result}, we show the changes in descriptions with our ProAPO. After iterative optimization, common descriptions such as ``\texttt{flight over ocean}'' and ``\texttt{long wings and hooked bill}'' are removed. Discriminative descriptions are also retained in candidate prompts, \textit{e.g.}, ``\texttt{white body}'' for Laysan Albatross, and ``\texttt{black underpants}'' for Sooty Albatross. As such, we see a notable improvement in description-based methods~\cite{DCLIP, AdaptCLIP, CuPL, GPT4Vis} with our ATO and ProAPO in~\cref{fig: improve_description_methods}(b) by at least 2.7\% average in thirteen datasets. It further verifies the effectiveness of our progressive optimization.

 

{
\begin{table}[t]
  \centering
  \resizebox{0.9\linewidth}{!}
    {
    \begin{tabular}
        {l c c c c c | c c c }    
        \toprule
        \multicolumn{6}{c}{\textbf{Component}}  &   \\
        \cmidrule(lr){1-6} 
        & \texttt{Add} & \texttt{Del} & \texttt{Rep} & \texttt{Cross} & \texttt{Mut}  & \textbf{IN-1K} & {\textbf{Avg (11)}} & {\textbf{Avg (13)}} \\
        \midrule
       \multicolumn{5}{l}{CLIP (Baseline)} & & 62.1 & 61.8 & 59.3  \\ 
       \midrule
       \multicolumn{5}{l}{\textbf{\textit{edit-based generation}}} \\
       \texttt{a)} & \cmark & & & & & 63.8 & 66.0  & 63.3 \\ 
       \texttt{b)} & \cmark & \cmark & & & & 64.6 & 66.4  & 63.8 \\
       \texttt{c)} & \cmark &  & \cmark & & & 64.4 & 66.5  & 63.8 \\
       \texttt{d)} & \cmark & \cmark & \cmark & & & 64.6 & 66.7  & 64.0  \\ 
        \midrule
        \multicolumn{5}{l}{\textbf{\textit{evolution-based generation}}} \\
        \texttt{e)} & \cmark & \cmark & \cmark & \cmark & & 64.6 & 67.3  & 64.5 \\ 
        \texttt{f)} & \cmark & \cmark & \cmark & & \cmark &  64.7 & 67.1  & 64.3  \\ 
        \highlight{\texttt{g)}} & \highlight{\cmark} & \highlight{\cmark} & \highlight{\cmark} & \highlight{\cmark} & \highlight{\cmark} & \highlight{\textbf{64.7}} & \highlight{\textbf{67.9}}  & \highlight{\textbf{65.0}} \\
        \bottomrule
    \end{tabular}
}
\vspace{-6pt}
\caption{\textbf{Ablation of edit- and evolution-based operators.}}
\vspace{-5pt}
\label{tab: ablation_generate}
\end{table}
}


\subsection{Ablation Study}
\label{sec: ablation_study}

\noindent \textbf{Are both} \texttt{GEN} \textbf{and} \texttt{EVO} \textbf{algorithms necessary?}
In~\cref{tab: ablation_generate}, we ablate edit- and evolution-based operators. For edit-based operators, we observe that the model with add, delete, and replace operations achieves a higher result in row d). After introducing evolution-based operators, \textit{i.e.}, crossover operator to combine advantages of high-scoring candidates, and mutation operator to avoid locally optimal solutions, we see an increase in performance in rows e)-g). It confirms that evolution-based operators make the model search the optimal prompt faster with limited iterations.




{
\renewcommand{\arraystretch}{1.02} 
\begin{table}[t]
  \centering
  \resizebox{0.88\linewidth}{!}
    {
    \begin{tabular}
        {l | c c c | c}  
        \toprule
        {\textbf{Module} (ViT-B/32)}  & \textbf{IN-1K} & {\textbf{Avg (11)}} & {\textbf{Avg (13)}} & \textbf{Times} \\
        \midrule
         
        \texttt{a)} w/o prompt sampling & 64.4 & 67.3 & 64.5 & 12 min \\
        \texttt{b)} w/o group sampling & \textbf{64.8} & \textbf{68.1} & \textbf{65.2} & \textbf{306 min} \\ 
        \texttt{c)} w/o sampling strategies & 64.5 & 67.2 & 64.4 & \underline{302 min} \\

        \midrule
        
        \highlight{\textbf{ProAPO} (full model)} & \highlight{\underline{64.7}} & \highlight{\underline{67.9}} & \highlight{\underline{65.0}} & \highlight{15 min} \\
        \bottomrule
    \end{tabular}
}
\vspace{-6pt}
  \caption{\textbf{Ablation of two sampling strategies.}}
\vspace{-8pt}
  \label{tab: ablation_sample}
\end{table}
}

\noindent \textbf{Does sample strategies degrade performance?}
In~\cref{tab: ablation_sample}, we ablate two sampling strategies for description optimization. Without the prompt sampling, we see a slight decrease in times while results drop in row a). It verifies the effectiveness of the prompt sampling in finding a better initial point. Without the group sampling to select salient categories for optimization, we observe a notable increase in time costs (from 15 min to 300+ min, 20 times) yet similar results in row b) and the full model. It reveals that group sampling simultaneously improves performance and efficiency.


\noindent \textbf{Which score function is better?}
In~\cref{fig: score_result}, we compare the different score functions, \textit{i.e.}, accuracy (used in PN~\cite{P_N}) and our fitness score, to evaluate the quality of the candidate prompt. PCCs (Pearson Correlation Coefficients) are introduced to evaluate the linear relationship between training metrics and test performance. The high PCC values mean a strong correlation.
We see that the model with our fitness score achieves stable and high test results compared to previous score functions when achieving the best score. Besides, a higher PPC value further verifies that our fitness score effectively alleviates the overfitting problem.



{
\begin{figure}[t]
\centering
{
    \hfill
    \subfloat[Previous score function.]{\includegraphics[width=0.23\textwidth]{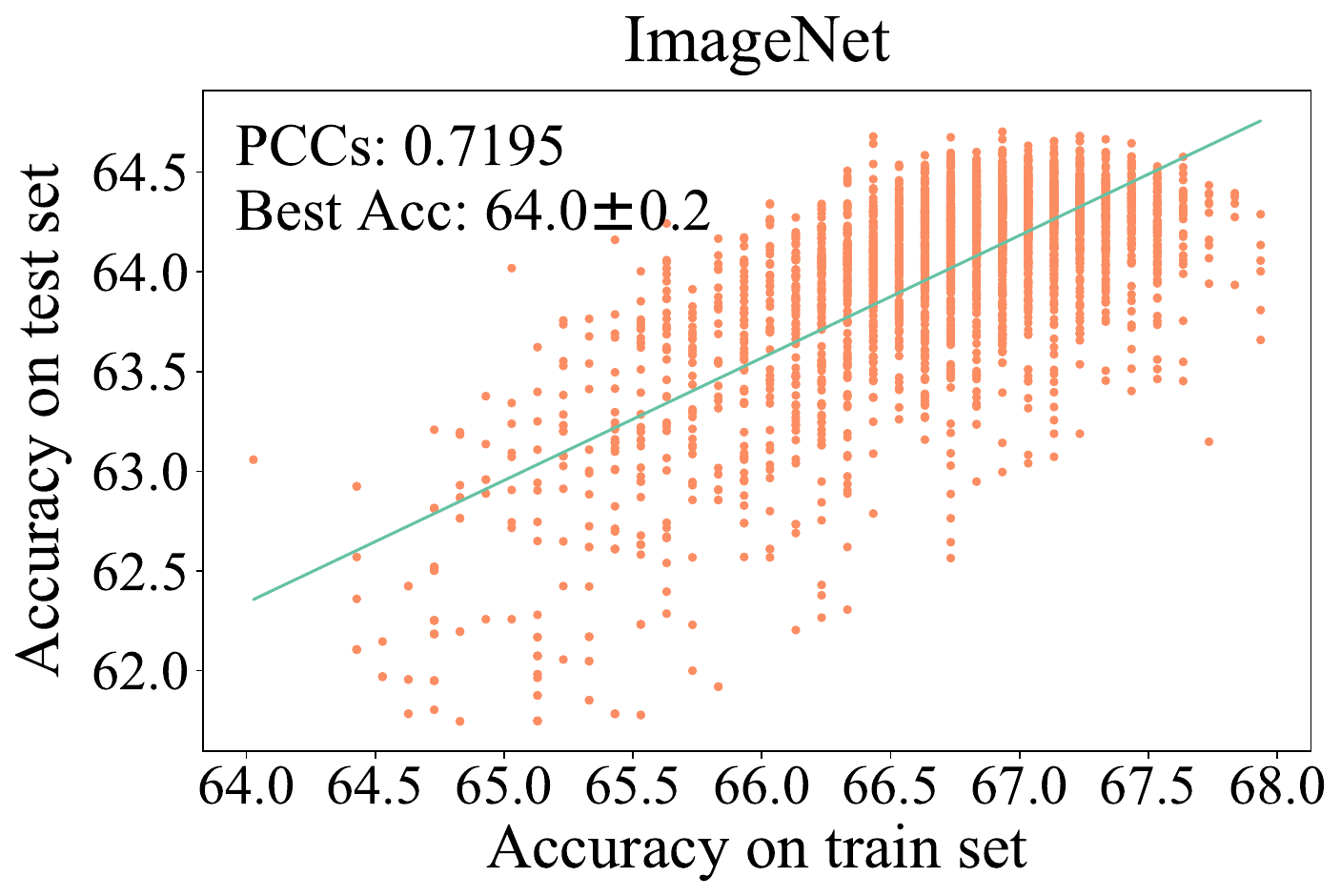}}
    \hfill
    \subfloat[Our fitness score]{\includegraphics[width=0.23\textwidth]{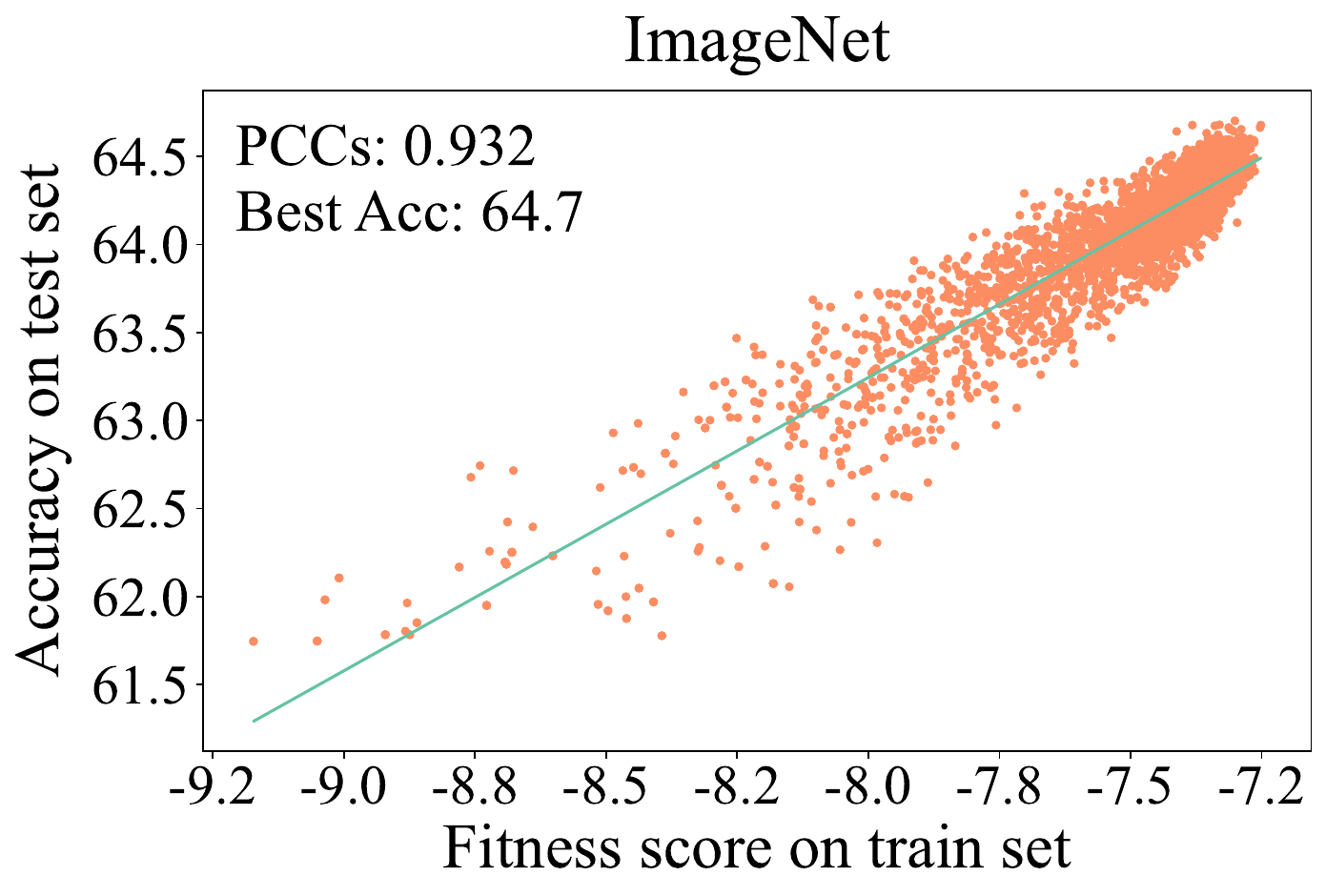}}
}
\vspace{-7pt}
\caption{\textbf{Effect of different score functions}. Higher PCC values mean stronger correlations between training metrics and test results. Best Acc is the test result when achieving the best score.}
\label{fig: score_result}
\vspace{-5pt}
\end{figure}
}



{
\begin{figure}[t]
\centering
{
    \hfill
    \subfloat[Iteration times $T$.]{\includegraphics[width=0.154\textwidth]{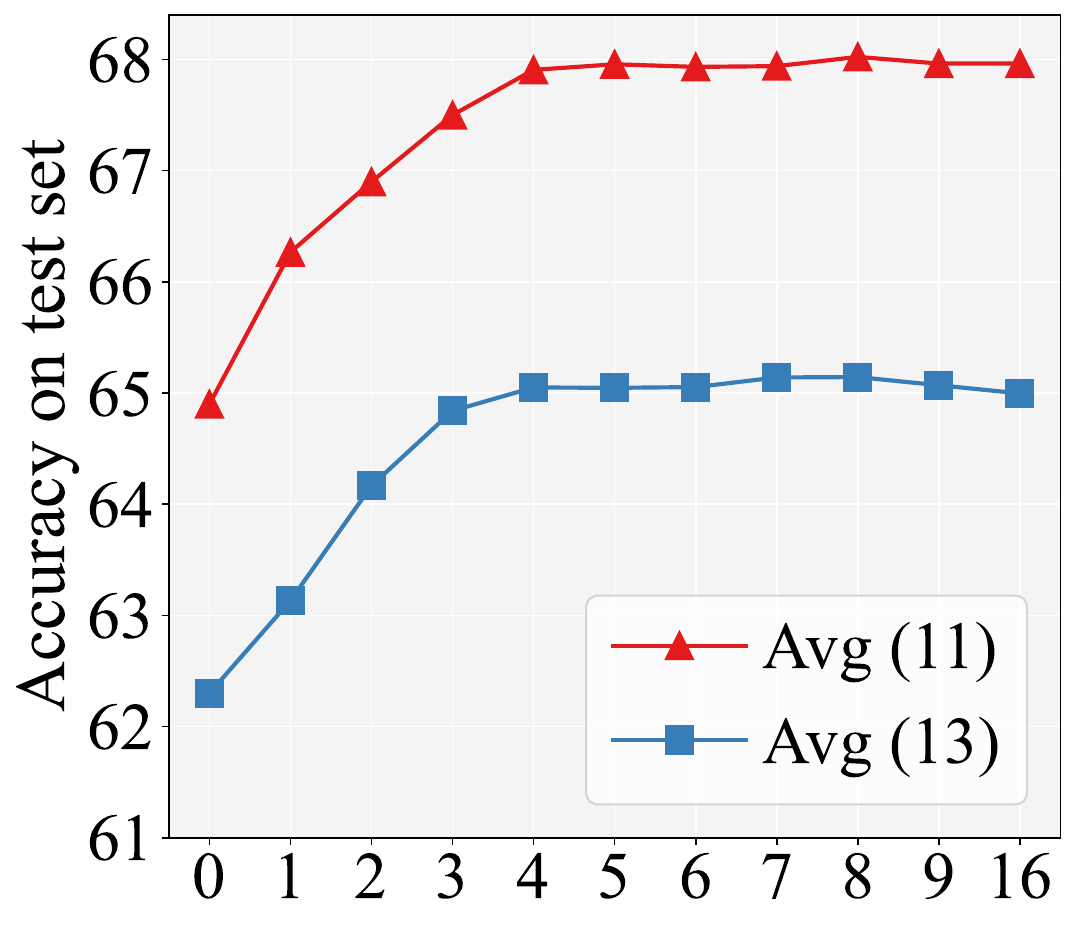}}
    \hfill
    \subfloat[Generated numbers]{\includegraphics[width=0.154\textwidth]{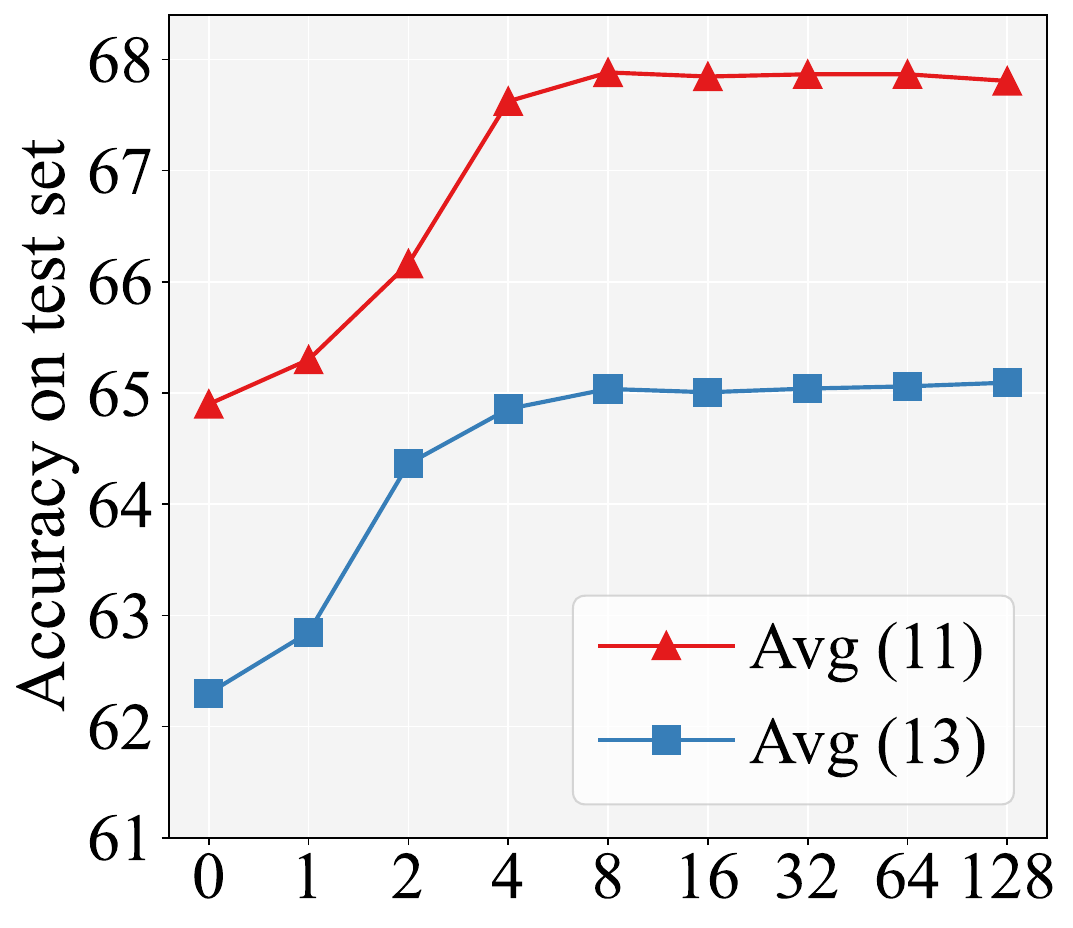}}
    \hfill
    \subfloat[Sampling groups $S$]{\includegraphics[width=0.154\textwidth]{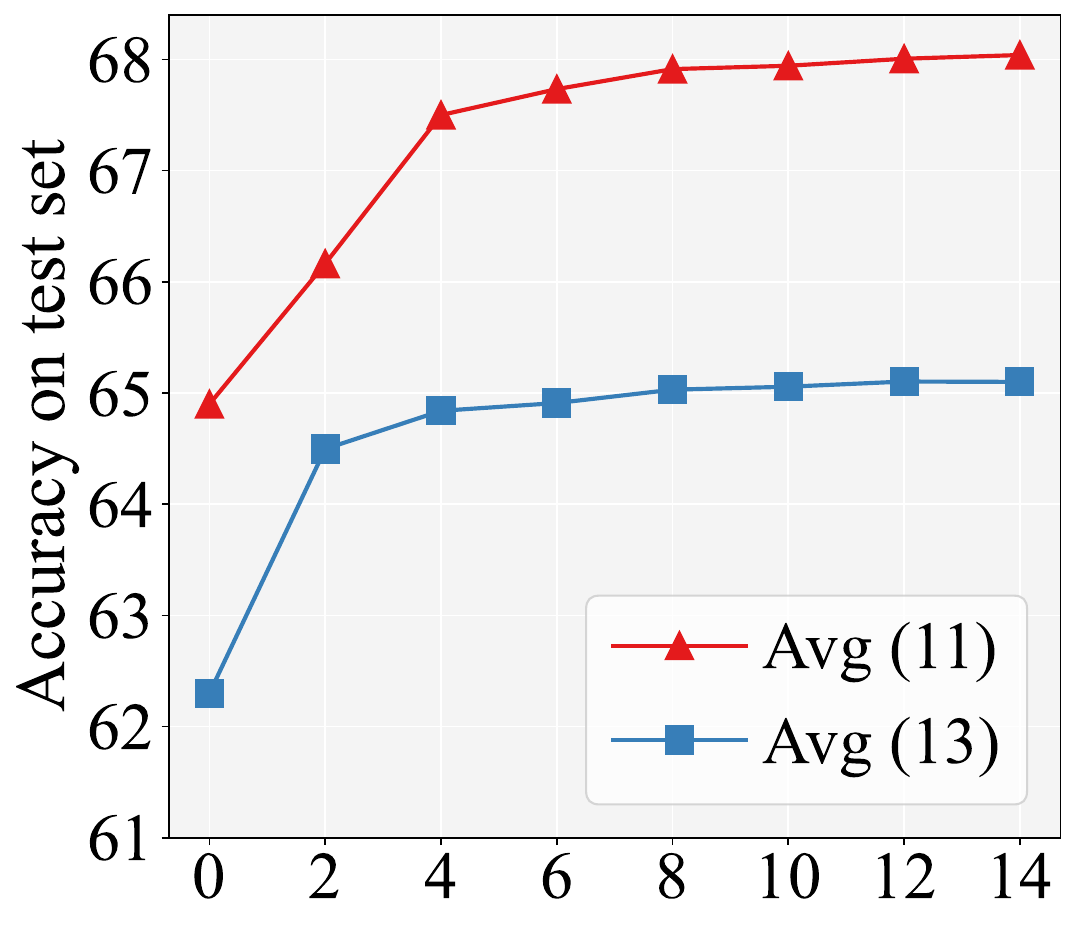}}
}
\vspace{-7pt}
\caption{\textbf{The effect of hyperparameter analysis}. The hyperparameter value set to 0 denotes the result of CuPL~\cite{CuPL}.
}
\label{fig: hyper_result}
\vspace{-5pt}
\end{figure}
}

\subsection{Hyperparameter Sensitivity}
\label{sec: hyper_ablation}

\noindent \textbf{How many iterations of Alg.~\ref{alg: APO} to use?}
In~\cref{fig: hyper_result}(a), we show the effect of iteration times $T$ in \texttt{APO} algorithm. As the number of iterations increases, we see a consistent performance improvement compared to the baseline CuPL. It demonstrates that iterative optimization improves prompt quality. Stable results are achieved when $T \geq 4$.


\noindent \textbf{How many generated prompts to use?}
In~\cref{fig: hyper_result}(b), we show the effect of generated numbers in \texttt{GEN} and \texttt{EVO} algorithm, where we set $M=N$. Similarly, progressive improvements are seen as generated numbers increase. We see a reliable result when $M = N \geq 8$.
It reveals the effectiveness of generating diverse candidates by our algorithm.


\noindent \textbf{Are more sampling groups better?}
In~\cref{fig: hyper_result}(c), we show the effect of the number of salient groups $S$ in the group sampling strategy. We see a notable improvement when $S=2$. 
As the number of groups $S$ increases, it achieves stable results when $S \geq 8$. 
It verifies that optimizing several salient categories can achieve comparable performances with all categories and save iteration costs.

\section{Conclusion}
\label{sec: conclusion}

We propose an evolution-based algorithm for VLMs to progressively refine prompts from task-specific to class-specific levels. To save generation costs, ProAPO uses several edit- and evolution-based operators to create candidate prompts with a prompt library. Results show that our fitness score mitigates overfitting in class-specific prompts. We empirically verify that two sampling strategies improve performance and save iteration times. Extensive experiments on thirteen datasets reveal that ProAPO consistently outperforms SOTA textual prompt-based methods on low-shot tasks. Moreover, our method effectively improves adapter-based and description-based methods and easily transfers across different backbones. We hope ProAPO could provide new insight into adapting VLMs from the textual side.


\section*{Acknowledgements}
This work was supported by the Central Guidance for Local Special Project (Grant No. Z231100005923044).

{
    \small
    \bibliographystyle{ieeenat_fullname}
    \bibliography{main}
}

\newpage

\appendix

We first describe detailed processes of building the prompt library and sampling strategies in our method:
\begin{itemize}
    \item \cref{sec_supp: build_prompt_library}: Details of building prompt library.
    \item \cref{sec_supp: prompt_sampling}: Details of prompt sampling strategy.
    \item \cref{sec_supp: group_sampling}: Details of group sampling strategy.
\end{itemize}

Then, we show more experiments to show the effectiveness of our ProAPO:
\begin{itemize}
    \item \cref{sec_supp: implement_details}: More implementation details.
    \item \cref{sec_supp: different_backbone_result}: Results on different backbones.
    \item \cref{supp_sec: more_comparison_with_sota_methods}: More comparisons with SOTA methods.
    \item \cref{supp_sec: ablation_template_and_description}: Ablation of progressive optimization.
    \item \cref{supp_sec: more_ablation_operator}: More ablation of operators.
    \item \cref{supp_sec: more_ablation_group_sampling}: More ablation of group sampling.
    \item \cref{supp_sec: ablation_of_cost_computation}: Ablation of cost computation.
    \item \cref{supp_sec: effect_shots}: Effect of shot numbers.
    \item \cref{supp_sec: effect_alpha}: Effect of scalar $\alpha$ in score function.
    \item \cref{supp_sec: effect_sampled_numbers}: Effect of sampled numbers in prompt sampling.
    \item \cref{supp_sec: effect_of_quality_of_prompt_library}: Effect of quality of prompt library.
    \item \cref{sec_supp: more_qualitative_result}: More qualitative results.
\end{itemize}

We also provide detailed results for experiments appearing in the main paper: 
\begin{itemize}
    \item \cref{sec_supp: transfer_to_adapter}: Results of transfering to adapter-based methods.
    \item \cref{sec_supp: transfer_to_backbones}: Results of transferring to different backbones.
    \item \cref{supp_sec: ensemble_vs_single}: Analysis of single vs ensemble prompts.
    \item \cref{sec_supp: improve_description_methods}: Improvement by iterative optimization.
    \item \cref{supp_sec: ablation_operator}: Ablation of edit and evolution operators.
    \item \cref{supp_sec: ablation_two_sampling}: Ablation of two sampling strategies.
    \item \cref{supp_sec: effect_score_func}: Ablation of different score functions.
\end{itemize}

\section{Details of Building Prompt Library}
\label{sec_supp: build_prompt_library}

\subsection{Details of Building Template Library}
The template library aims to collect a set of templates that provide task-specific contextual information, which can address issues of semantic ambiguity caused by class names. It contains processes for collecting templates, generating dataset domains, and adding dataset domains to templates.


\textbf{Collecting templates.} We utilize two ways to collect templates. First, pre-defined templates, such as Template-80~\cite{CLIP}, FILIP-8~\cite{FILIP}, and DEFILIP-6~\cite{DEFILIP} can be used. Second, similar to PN~\cite{P_N}, we query LLMs to create diverse templates by the following prompt:
\begin{quote}
    \makebox[\linewidth]{%
        \colorbox{lightblue}{%
            \hspace*{0mm} 
            \begin{minipage}{\dimexpr\linewidth+10\fboxsep\relax} 
                ``Hi, ChatGPT! I would like your help to prompt for image classification using CLIP. As a human-level prompt engineer, your task is to create a set of Templates like the following for visual classification. For example: 
                \newline \newline
                a photo of a \{\}.''
            \end{minipage}%
        }%
    }
\end{quote}

\textbf{Generating dataset domain by LLMs.} Inspired by previous description-based methods~\cite{WaffleCLIP, VDT_2023_ICCV}, we query LLMs to generate dataset domain information to provide task-specific context. For this purpose, we use the prompt:
\begin{quote}
    \makebox[\linewidth]{%
        \colorbox{lightblue}{%
            \hspace*{0mm} 
            \begin{minipage}{\dimexpr\linewidth+10\fboxsep\relax} 
                ``Hi, ChatGPT! I would like your help in generating dataset domain information for image classification based on the dataset paper. A few words are good. Please return directly without explanation. 
                \newline \newline
                \{\texttt{uploaded PDF}\}.''
            \end{minipage}%
        }%
    }
\end{quote}
Here, \{\texttt{uploaded PDF}\} represents the uploading of the paper of the dataset to LLMs. 
Generated dataset domain information is summarized in~\cref{supp_tab: domain_information}.

{
\renewcommand{\arraystretch}{1.1} 
\begin{table}[htbp]
  \centering
  \resizebox{1.0\linewidth}{!}
    {
    \begin{tabularx}{0.56\textwidth}
        {l | X }  
        \toprule
        {\textbf{Dataset}}  & \textbf{Domain Information} \\
        \midrule
        IN-1K~\cite{Imagenet} & real scenario; natural scene \\
        Caltech~\cite{caltech101} & object; everyday objects; common items \\
        Cars~\cite{Cars} & car; vehicles; auto-mobile \\ 
        CUB~\cite{CUB} & bird; wildlife; ornithology  \\
        DTD~\cite{DTD} & textures; patterns; surface; material \\
        ESAT~\cite{EuroSAT} & land cover; remote sensing; satellite photo; satellite imagery; aerial or satellite images; centered satellite photo \\
        FGVC~\cite{FGVC} & aircraft; airplane; plane; airliner \\ 
        FLO~\cite{FLO} & flower; floral; botanical; bloom \\
        Food~\cite{Food101} & food; dishes; cuisine; nourishment \\ 
        Pets~\cite{oxford_pets} & pet; domestic animals; breed; dog or cat \\ 
        Places~\cite{Places365} & place; scene \\ 
        SUN~\cite{SUN} &  place; scene \\ 
        UCF~\cite{UCF101} & action; human action; human activities; person doing \\ 
        \bottomrule
    \end{tabularx}
}
\vspace{-6pt}
  \caption{\textbf{Generated dataset domain information.}}
  \label{supp_tab: domain_information}
\end{table}
}

\textbf{Adding dataset domain to templates.} We supplement templates with dataset domain information in the following four ways: (1) Add ``a type of \{\texttt{domain}\}''. (2) Replace ``\{\texttt{class}\}'' with ``\{\texttt{domain}\}:\{\texttt{class}\}''. (3) Replace ``photo'' with ``\{\texttt{domain}\}''. (4) Replace ``photo'' with ``\{\texttt{domain}\} photo''. Taking ``a photo of a \{\texttt{class}\}'' as an example, we modify the templates with the above four ways to add dataset domain information as follows:
\begin{quote}
    \makebox[\linewidth]{%
        \colorbox{lightblue}{%
            \hspace*{0mm} 
            \begin{minipage}{\dimexpr\linewidth+10\fboxsep\relax} 
                \begin{enumerate}
                    \item a photo of a \{\texttt{class}\}, a type of \{\texttt{domain}\}.
                    \item a photo of a \{\texttt{domain}\}: \{\texttt{class}\}.
                    \item a \{\texttt{domain}\} of a \{\texttt{class}\}.
                    \item a \{\texttt{domain}\} photo of a \{\texttt{class}\}.
                \end{enumerate}
            \end{minipage}%
        }%
    }
\end{quote}
Here, \{\texttt{class}\} and \{\texttt{domain}\} denote category name and dataset domain information, respectively.

\subsection{Details of Building Description Library}
\label{supp_sec: build_description_library}

Description Library aims to provide a set of visual descriptions for each category, enhancing visual semantics for fine-grained recognition in prompts. It contains processes for generating visual descriptions and category synonyms and integrating descriptions with the best templates.


{
\renewcommand{\arraystretch}{1.1} 
\begin{table}[htbp]
  \centering
  \resizebox{1.0\linewidth}{!}
    {
    \begin{tabularx}{0.56\textwidth}
        {l | X }  
        \toprule
        {\textbf{Method}}  & \textbf{Prompts} \\
        \midrule
        DCLIP~\cite{DCLIP} & Q: What are useful visual features for distinguishing a \{\texttt{class}\} in a photo? \\
        & A: There are several useful visual features to tell there is a \{\texttt{class}\} in a photo: \\
        
        \midrule 
        CuPL-Base~\cite{CuPL} & Describe what a \{\texttt{class}\} looks like. \\ 
        & Describe a \{\texttt{class}\}. \\
        & What are the identifying characteristics of a \{\texttt{class}\}? \\
        
        \midrule

        CuPL-Full~\cite{CuPL} & Describe what a \{\texttt{class}\} looks like. \\ 
        & How can you identify a \{\texttt{class}\}? \\ 
        & What does a \{\texttt{class}\} look like? \\
        & Describe an image from the internet of a \{\texttt{class}\}\\
        & A caption of an image of a \{\texttt{class}\}: \\
        
        \midrule
        GPT4Vis~\cite{GPT4Vis} & I want you to act as an image description expert. I will give you a word and your task is to give me 20 sentences to describe the word. Your description must accurately revolve around this word and be as objective, detailed and diverse as possible. In addition, the subject of your description is a some kind of object photograph. Output the sentences in a json format which key is the the word and the value is a list composed of these sentences. Do not provide any explanations. The first word is ``\{\texttt{class}\}". \\ 

        \midrule 
        AdaptCLIP~\cite{AdaptCLIP} & What characteristics can be used to differentiate \{\texttt{class}\} from other \{\texttt{domain}\} based on just a photo? Provide an exhaustive list of all attributes that can be used to identify the \{\texttt{domain}\} uniquely. Texts should be of the form “\{\texttt{domain}\} with \{\texttt{characteristic}\}”. \\
        \bottomrule
    \end{tabularx}
}
  \caption{\textbf{Prompts for generating visual descriptions.}}
  \label{supp_tab: generate_description}
\end{table}
}

\textbf{Generating category synonym}.
Except for descriptions, we also replace class names from the dataset with their synonyms to create diverse class-specific prompts. For this purpose, we use the following prompt to ask LLMs to generate category synonyms:
\begin{quote}
    \makebox[\linewidth]{%
        \colorbox{lightblue}{%
            \hspace*{0mm} 
            \begin{minipage}{\dimexpr\linewidth+10\fboxsep\relax} 
                ``Hi, ChatGPT! I would like your help in generating category synonyms. As a \{\texttt{domain}\} expert, I will provide you with a category name. Your task is to provide synonyms for the current category. If it has subclasses, return them as well. Please return directly without explanation.
                \newline \newline 
                User: I want to give the synonyms of \{\texttt{class}\}. 
                \newline
                Assistant: ''
            \end{minipage}%
        }%
    }
\end{quote}

\textbf{Generating visual descriptions for each category}.
Similar to previous description methods~\cite{CuPL, DCLIP, GPT4Vis, AdaptCLIP}, we instruct LLM to generate visual descriptions for each category by several prompts, which are summarized in~\cref{supp_tab: generate_description}.

\textbf{Integrating descriptions with the best templates}.
We use the following prompt to integrate descriptions with templates: ``\{\texttt{template}\}. \{\texttt{description}.\}''.

After the above processes, we collect diverse visual descriptions for each category $c$, denoted as $\text{VD}(c)$. For each group iteration, we select the descriptions for categories in the specific group as the description library. Moreover, the prompt sampling strategy also utilizes these descriptions for class-specific initialization.

\section{Details of Prompt Sampling Strategy}
\label{sec_supp: prompt_sampling} 
The detailed prompt sampling strategy is summarized in Alg.~\ref{supp_alg: prompt_strategy}. Visual descriptions of each class $\text{VD}(c)$ are collected by the above process (see~\cref{supp_sec: build_description_library}). We utilize the candidate prompt $P_t^*$ with the best templates as an initial point. The $\textsc{RandomSample}(\cdot)$ operator denotes randomly selecting a set of elements from a given set. We randomly sample descriptions for each category to create multiple candidate prompts (Lines 2-8). After $T_{sample}$-times steps, we select the candidate prompt $\hat{P}_0$ with the highest score for description initialization (Line 9). It ensures that subsequent optimization is around the optimal initial point. We set $T_{sample} = 32 $ for all datasets in the default setting.


\begin{algorithm}[htbp]
\caption{Prompt Sampling Strategy.}
\label{supp_alg: prompt_strategy}
\begin{algorithmic}[1]
\REQUIRE $\mathcal{D} \leftarrow \{{(x, y)}\}_n$: training samples, $F:  \mathcal{D} \times P \to \mathbb{R}$: score function, $\mathcal{C}$: class labels, $\text{VD}(c)$: visual descriptions of class $c$, $P_t^*$: the prompt candidate with the best template
\STATE $\mathcal{U} \leftarrow \{P_t^*\} $ 
\FOR{$i=1$ to $T_{sample}$}
    \STATE $P_i \leftarrow P_t^* $
    \FORALL{class $c \in \mathcal{C}$}
        \STATE $P_i \leftarrow P_i \cup \textsc{RandomSample}(\text{VD}(c))$
    \ENDFOR
    \STATE $\mathcal{U} \leftarrow \mathcal{U} \cup \{ P_i \} $ 
\ENDFOR
\STATE $\hat{P}_0 \leftarrow \arg\max_{{P} \in \mathcal{U}} F(\mathcal{D}, {P})$ 
\RETURN the candidate prompt with the highest score $\hat{P}_0$
\end{algorithmic}
\end{algorithm}


\begin{algorithm}[htbp]
\caption{Group Sampling Strategy.}
\label{supp_alg: group_strategy}
\begin{algorithmic}[1]
\REQUIRE $\mathcal{D} \leftarrow \{{(x, y)}\}_n$: training samples, $F:  \mathcal{D} \times P \to \mathbb{R}$: score function, $\mathcal{C}$: class labels, $\text{VD}(c)$: visual descriptions of class $c$, $P_t^*$: prompt candidate with the best template, $\text{pred}(x)$: prediction for image $x$
\FORALL{class $c \in \mathcal{C}$}
    \STATE $\textsc{MisClass}(c) \leftarrow \emptyset $
\ENDFOR
\FORALL{training sample $(x, y) \in \mathcal{D}$}
    \IF{$\text{pred}(x) \neq y $}
        \STATE $\textsc{MisClass}(y) \leftarrow \textsc{MisClass}(y) \cup \{\text{pred}(x)\} $
    \ENDIF
\ENDFOR

\FORALL{class $c \in \mathcal{C}$}
    \STATE \textbf{Select Class Images}: $\textsc{Data} (c) \leftarrow \{ (x, y) \; | \; y = c\}_{(x, y) \in \mathcal{D}}$
    \STATE \textbf{Compute Accuracy}: $\textsc{Acc} (c) \leftarrow F( \textsc{Data} (c), P^*_t )$
    \STATE \textbf{Add Descriptions}: $P_{c} \leftarrow P^*_t \cup \text{VD}(c) $
    \STATE \textbf{Compute Accuracy Gain}: $\textsc{AccGain} (c) \leftarrow F( \textsc{Data} (c), P_c ) -  \textsc{Acc} (c) $
\ENDFOR
\STATE \textbf{Sort Class by Accuracy}: $\mathcal{C}_{wst}$, retaining the classes with the lowest top-$n_{wst}$ accuracy
\STATE \textbf{Sort Class by Accuracy Gain}: $\mathcal{C}_{sln}$, retaining the classes with the top-$n_{sln}$ accuracy gain
\STATE \textbf{Initialize Group Set}: $\mathcal{G} \leftarrow \emptyset$
\FORALL{class $c \in \mathcal{C}_{wst}$}
    \STATE $\mathcal{G} \leftarrow \mathcal{G} \cup \{ \textsc{MisClass}(y) \}$
\ENDFOR
\FORALL{class $c \in \mathcal{C}_{sln}$}
    \STATE $\mathcal{G} \leftarrow \mathcal{G} \cup \{ \textsc{MisClass}(y) \}$
\ENDFOR
\RETURN sampled groups $\mathcal{G}$
\end{algorithmic}
\end{algorithm}

\section{Details of Group Sampling Strategy}
\label{sec_supp: group_sampling}
The detailed group sampling strategy is summarized in Alg.~\ref{supp_alg: group_strategy}.
It contains processes of obtaining misclassified categories and selecting the worst and salient groups.

\noindent \textbf{Obtaining misclassified categories}.
In Lines 1-8 of Alg.~\ref{supp_alg: group_strategy}, we collect misclassified set for each category by $\textsc{MisClass}(\cdot)$ operator. Given an image $x$, if the prediction $\text{pred}(x)$ is not its corresponding label $y$, we will add $\text{pred}(x)$ to the misclassified set for category $y$. In fact, we also ablate the K-means clustering algorithm to group categories (in~\cref{supp_sec: more_ablation_group_sampling}). Results show that the misclassified set achieves better performance than the K-means algorithm.

\noindent \textbf{Selecting the worst groups} aims to select categories with the lowest top-$n_{wst}$ accuracy and corresponding misclassified categories. We first compute the accuracy for each category in Line 11. Then, we sort the categories by accuracy and retain the top-$n_{wst}$ worst categories in Line 15. Finally, $n_{wst}$ groups are added to the set $\mathcal{G}$ in Lines 18-20.

\noindent \textbf{Selecting the salient groups} aims to select categories with the top-$n_{sln}$ performance gains and its misclassified categories after adding descriptions. In Line 13, we compute the accuracy gains after adding the descriptions. Then, we sort the categories by accuracy gain and retain the top-$n_{sln}$ accuracy gain categories in Line 16. At last, $n_{sln}$ groups are added to the set $\mathcal{G}$ in Lines 21-23.

Finally, we collect $S = n_{wst} + n_{sln}$ groups for subsequent description optimization.

\section{More Implementation Details}
\label{sec_supp: implement_details}

\subsection{Hyperparameter Settings}
In~\cref{supp_tab: exp_details}, we show the searched hyperparameter settings for thirteen datasets. All results are average with four seeds. Except for $1, 2, 3$ as seeds like CoOp~\cite{CoOp}, we add $42$ as our fourth seed to further evaluate the stability of our method. In the default setting, we use the same LLMs as the description methods, \textit{i.e.}, GPT-3~\cite{GPT3} for CuPL~\cite{CuPL} and DCLIP~\cite{DCLIP}, GPT-4~\cite{GPT4_Tech} for GPT4Vis~\cite{GPT4Vis} and AdaptCLIP~\cite{AdaptCLIP}. 


{
\renewcommand{\arraystretch}{1.1} 
\begin{table}[htbp]
  \centering
  \resizebox{1.0\linewidth}{!}
    {
    \begin{tabular}
        {l | c | c | c | c | c | c | c  }  
        \toprule
        {\textbf{Dataset}}  & $T$ &  $M$ & $N$ & $\alpha$ & $n_{wst}$ & $n_{sln}$ & $T_{sample}$ \\
        \midrule
        IN-1K~\cite{Imagenet} & 4 & 8 & 8 & 1e3 & 4 & 4 & 32 \\
        Caltech~\cite{caltech101} & 2 & 8 & 8 & 1e2 & 2 & 2 & 32 \\
        Cars~\cite{Cars} & 4 & 8 & 8 & 1e4 & 4 & 4 & 32 \\ 
        CUB~\cite{CUB} & 4 & 8 & 8 & 1e2 & 4 & 4 & 32 \\
        DTD~\cite{DTD} & 4 &  8 & 8 & 1e3 & 4 & 4 & 32 \\
        ESAT~\cite{EuroSAT} & 4 &  8 &  8 & 1e3 & 3 & 3 & 32 \\
        FGVC~\cite{FGVC} & 4 & 8 & 8 & 1e3 & 4 & 4 & 32 \\ 
        FLO~\cite{FLO} & 4 & 8 & 8 & 1e3 & 4 & 4 & 32 \\
        Food~\cite{Food101} & 4 &  8 & 8 & 1e3 & 2 & 2 & 32 \\ 
        Pets~\cite{oxford_pets} & 2 & 8 &  8 & 1e4 & 2 & 2 & 32 \\ 
        Places~\cite{Places365} & 4  & 8 &  8 & 1e2 & 3 & 3 & 32 \\ 
        SUN~\cite{SUN} & 2 &  8 & 8 & 1e4 & 4 & 4 & 32 \\ 
        UCF~\cite{UCF101} & 4 &  8 & 8 & 1e3 & 3 & 3 & 32 \\ 
        \bottomrule
    \end{tabular}
}
\vspace{-6pt}
  \caption{\textbf{Hyperparameters settings for thirteen datasets.}}
\vspace{-10pt}
  \label{supp_tab: exp_details}
\end{table}
}

\subsection{More Related Work}
\noindent 
\textbf{Large-scale vision-language models}
like CLIP~\cite{CLIP} have shown promising performance on various tasks. They align visual and textual spaces to a joint space via training on millions of image-text pairs from the web. Other work~\cite{Align, DEFILIP, DeClip, FILIP, BLIP, Flamingo, SLIP, EVA-01, EVA-02} has furthered this paradigm to learn more accurate semantic alignment in joint space. 
In this work, we advance VLMs for downstream tasks by progressively learning optimal class-specific prompts with minimal supervision and no human intervention.

{
\renewcommand{\arraystretch}{1.1} 
\setlength{\tabcolsep}{3.8pt}

\begin{table*}[htbp]
  \centering
  \resizebox{0.98\linewidth}{!}
    {
    \begin{tabular}
        {l | ccccc ccccc ccc | c | c}
            
        \toprule
        \textbf{Module} & \rotatebox{90}{\textbf{IN-1K}} & \rotatebox{90}{\textbf{Caltech}} & \rotatebox{90}{\textbf{Cars}} & \rotatebox{90}{\textbf{CUB}} & \rotatebox{90}{\textbf{DTD}}  & \rotatebox{90}{\textbf{ESAT}} & \rotatebox{90}{\textbf{FGVC}} & \rotatebox{90}{\textbf{FLO}} & \rotatebox{90}{\textbf{Food}}  &  \rotatebox{90}{\textbf{Pets}} & \rotatebox{90}{\textbf{Places}} & \rotatebox{90}{\textbf{SUN}} & \rotatebox{90}{\textbf{UCF}} & \rotatebox{90}{\textbf{Avg (11)}} & \rotatebox{90}{\textbf{Avg (13)}} \\
        \midrule

        CLIP~\cite{CLIP} - ResNet50 &  57.9 & 84.5 & 53.9 & 44.7 & 38.8 & 28.6 & 15.9 & 60.2 & 74.0 & 83.2 & 38.2 & 58.0 & 56.9 & 55.6 & 53.4 \\ 
        CuPL~\cite{CuPL} &   61.2 & 88.3 & 55.3 & 48.7 & 49.5  & 38.2  & 18.9  & 67.0  & 80.1& 86.1& 41.2& 63.1  & 63.3   & 61.1  & 58.5 \\
        {\textbf{ProAPO} (ours)} & \textbf{61.5}  & \textbf{90.3} & \textbf{58.0} & \textbf{50.7} & \textbf{52.3} & \textbf{51.7} & \textbf{21.1} & \textbf{75.1} & \textbf{81.8} & \textbf{88.7} & \textbf{41.8} & \textbf{63.7} & \textbf{66.0} & \textbf{64.6}  & \textbf{61.8} \\

        $\Delta$ & \textcolor{retained}{+ 3.6} & \textcolor{retained}{+ 5.8} & \textcolor{retained}{+ 4.1} & \textcolor{retained}{+ 6.0} & \textcolor{retained}{+ 13.5} & \textcolor{retained}{+ 23.1} & \textcolor{retained}{+ 5.2} & \textcolor{retained}{+ 14.9} & \textcolor{retained}{+ 7.8} & \textcolor{retained}{+ 5.5} & \textcolor{retained}{+ 3.6} & \textcolor{retained}{+ 5.7} & \textcolor{retained}{+ 9.1} & \textcolor{retained}{+ 9.0} & \textcolor{retained}{+ 8.4}   \\

        \midrule

        CLIP~\cite{CLIP} - ResNet101 & 61.4  & 89.9  & 63.3  & 49.6  & 40.3  & 31.7  & 18.3  & 64.3  & 83.4  & 86.9  & 37.9  & 59.0  & 61.2  & 60.0  & 57.5  \\ 
        CuPL~\cite{CuPL} &  61.4  & 91.0  & 61.2  & 45.3  & 49.7  & 28.7  & 18.6  & 59.0  & 82.7  & 86.6  & \textbf{40.6}  & 62.3  & 56.4  & 59.8  & 57.2  \\
        {\textbf{ProAPO} (ours)} & \textbf{63.6} & \textbf{92.3} & \textbf{64.4} & \textbf{52.2} & \textbf{51.6} & \textbf{45.9} & \textbf{21.2} & \textbf{69.6} & \textbf{84.9} & \textbf{89.6} & \textbf{40.6} & \textbf{63.5} & \textbf{64.0} & \textbf{64.6}  & \textbf{61.8} \\
        $\Delta$ & \textcolor{retained}{+ 2.2} & \textcolor{retained}{+ 2.4} & \textcolor{retained}{+ 1.1} & \textcolor{retained}{+ 2.6} & \textcolor{retained}{+ 11.3} & \textcolor{retained}{+ 14.2} & \textcolor{retained}{+ 2.9} & \textcolor{retained}{+ 5.3} & \textcolor{retained}{+ 1.5} & \textcolor{retained}{+ 2.7} & \textcolor{retained}{+ 2.7} & \textcolor{retained}{+ 4.5} & \textcolor{retained}{+ 2.8} & \textcolor{retained}{+ 4.6} & \textcolor{retained}{+ 4.3} \\

        \midrule

        CLIP~\cite{CLIP} - ViT-B/32 & 62.1  & 91.2  & 60.4  & 51.7 & 42.9  & 43.9  & 20.2  & 66.0  & 83.2  & 86.8 & 39.9 & 62.1  & 60.9 & 61.8 & 59.3  \\
        CuPL~\cite{CuPL} &  {64.4}  & 92.9  & 60.7  & 53.3  & {50.6}  & 50.5  & 20.9  & 69.5  & 84.2  & 87.0  & {43.1}  & {66.3}  & 66.4  & 64.9  & 62.3  \\
        {\textbf{ProAPO} (ours)} & {\textbf{64.7}} & {\textbf{94.4}} & {\textbf{61.7}} & {\textbf{55.4}} & {\textbf{53.5}} & {\textbf{63.0}} & {\textbf{23.0}} & {\textbf{74.3}} & {\textbf{85.3}} & {\textbf{91.0}} & {\textbf{43.3}} & {\textbf{66.6}} & {\textbf{69.0}} & {\textbf{67.9}}  & {\textbf{65.0}} \\

        $\Delta$ & \textcolor{retained}{+ 2.6} & \textcolor{retained}{+ 3.2} & \textcolor{retained}{+ 1.3} & \textcolor{retained}{+ 3.7} & \textcolor{retained}{+ 10.6} & \textcolor{retained}{+ 19.1} & \textcolor{retained}{+ 2.8} & \textcolor{retained}{+ 8.3} & \textcolor{retained}{+ 2.1} & \textcolor{retained}{+ 4.2} & \textcolor{retained}{+ 3.4} & \textcolor{retained}{+ 4.5} & \textcolor{retained}{+ 8.1} & \textcolor{retained}{+ 6.1} & \textcolor{retained}{+ 5.7} \\

        \midrule

        CLIP~\cite{CLIP} - ViT-B/16 & 66.9  & 93.2  & 65.5  & 55.3  & 44.3  & 51.0  & 24.4  & 70.6  & 88.4  & 89.0  & 40.8  & 62.5  & 67.7  & 65.8  & 63.0  \\
        CuPL~\cite{CuPL} & 69.6  & 94.3  & 66.1  & 57.2  & 53.8  & 55.7  & 26.6  & 73.9  & 88.9  & 91.2  & 43.4  & \textbf{69.0}  & 70.3  & 69.0  & 66.1  \\
        {\textbf{ProAPO} (ours)} & \textbf{69.9} & \textbf{95.2} & \textbf{67.7} & \textbf{59.0} & \textbf{55.8} & \textbf{65.3} & \textbf{28.3} & \textbf{82.7} & \textbf{89.5} & \textbf{92.7} & \textbf{43.8} & {68.9} & \textbf{73.1} & \textbf{71.7}  & \textbf{68.6} \\
        $\Delta$ & \textcolor{retained}{+ 3.0} & \textcolor{retained}{+ 2.0} & \textcolor{retained}{+ 2.2} & \textcolor{retained}{+ 3.7} & \textcolor{retained}{+ 11.5} & \textcolor{retained}{+ 14.3} & \textcolor{retained}{+ 3.9} & \textcolor{retained}{+ 12.1} & \textcolor{retained}{+ 1.1} & \textcolor{retained}{+ 3.7} & \textcolor{retained}{+ 3.0} & \textcolor{retained}{+ 6.4} & \textcolor{retained}{+ 5.4} & \textcolor{retained}{+ 5.9} & \textcolor{retained}{+ 5.6} \\

        \midrule

        CLIP~\cite{CLIP} - ViT-L/14 &  73.5  & 95.1  & 76.8  & 62.5  & 52.1  & 61.5  & 33.4  & 79.5  & 93.1  & 93.3  & 40.7  & 67.6  & 75.0   & 72.8  & 69.5 \\ 
        CuPL~\cite{CuPL} & 76.7  & 96.2 & 77.6 &  61.4 & 62.6 & 62.4 & 36.1  & 79.7  & 93.4 &  93.8 & 43.8 &  73.2  & 78.3  & 75.5  & 71.9   \\
        {\textbf{ProAPO} (ours)} & \textbf{76.8} & \textbf{97.1} & \textbf{78.8} & \textbf{65.1} & \textbf{64.8} & \textbf{74.3} & \textbf{38.3} & \textbf{87.3} & \textbf{93.9} & \textbf{94.6} & \textbf{44.4} & \textbf{73.4} & \textbf{80.1} & \textbf{78.1}  & \textbf{74.5} \\
        $\Delta$ & \textcolor{retained}{+ 3.3} & \textcolor{retained}{+ 2.0} & \textcolor{retained}{+ 2.0} & \textcolor{retained}{+ 2.6} & \textcolor{retained}{+ 12.7} & \textcolor{retained}{+ 12.8} & \textcolor{retained}{+ 4.9} & \textcolor{retained}{+ 7.8} & \textcolor{retained}{+ 0.8} & \textcolor{retained}{+ 1.3} & \textcolor{retained}{+ 3.7} & \textcolor{retained}{+ 5.3} & \textcolor{retained}{+ 5.1} & \textcolor{retained}{+ 5.8} & \textcolor{retained}{+ 5.0} \\

        \midrule 

        OpenCLIP~\cite{OpenCLIP} - ViT-B/32 &  66.2  & 94.7  & 88.2  & 65.6  & 51.3  & 49.4  & 23.0  & 71.2 & 82.4 & 90.7 & 41.5 & 68.1 & 65.0  & 68.2  & 65.9  \\ 
        CuPL~\cite{CuPL} &  66.7  & 94.4  & 86.6  & 65.9  & 62.4  & 50.1  & 25.5  & 69.5  & 81.7  & 90.8  & 43.3  & 69.1  & 65.8  & 69.3  & 67.1   \\
        {\textbf{ProAPO} (ours)} & \textbf{67.0} & \textbf{95.8} & \textbf{88.7} & \textbf{67.3} & \textbf{65.1} & \textbf{66.0} & \textbf{27.5} & \textbf{81.8} & \textbf{83.2} & \textbf{91.9} & \textbf{43.4} & \textbf{69.7} & \textbf{70.2} & \textbf{73.3}  & \textbf{70.6} \\
        $\Delta$ & \textcolor{retained}{+ 0.8} & \textcolor{retained}{+ 1.1} & \textcolor{retained}{+ 0.5} & \textcolor{retained}{+ 1.7} & \textcolor{retained}{+ 13.8} & \textcolor{retained}{+ 16.6} & \textcolor{retained}{+ 4.5} & \textcolor{retained}{+ 10.6} & \textcolor{retained}{+ 0.8} & \textcolor{retained}{+ 1.2} & \textcolor{retained}{+ 1.9} & \textcolor{retained}{+ 1.6} & \textcolor{retained}{+ 5.2} & \textcolor{retained}{+ 5.1} & \textcolor{retained}{+ 4.7} \\

        \midrule 

        EVA02~\cite{EVA-02} - ViT-B/16 & 74.6  & \textbf{97.2}  & 79.2  & 60.8  & 49.7  & 68.0  & 24.6  & 75.6  & 89.5  & 92.2  & 42.9  & 70.7 & 68.6  & 71.8  & 68.7  \\ 
        CuPL~\cite{CuPL} &  75.4  & 96.7  & 79.2  & 61.8  & 59.1  & 61.7  & 27.5  & 75.2  & 89.3  & 92.1  & 44.0  & 72.5  & 71.9  & 72.8  & 69.7 \\
        {\textbf{ProAPO} (ours)} & \textbf{75.5} & 97.0 & \textbf{80.0} & \textbf{62.8} & \textbf{61.3} & \textbf{74.2} & \textbf{29.7} & \textbf{89.1} & \textbf{89.6} & \textbf{93.5} & \textbf{44.5} & \textbf{72.5} & \textbf{75.2} & \textbf{76.2}  & \textbf{72.7}  \\
        $\Delta$ & \textcolor{retained}{+ 0.9} & -0.2 & \textcolor{retained}{+ 0.8} & \textcolor{retained}{+ 2.0} & \textcolor{retained}{+ 11.6} & \textcolor{retained}{+ 6.2} & \textcolor{retained}{+ 5.1} & \textcolor{retained}{+ 13.5} & \textcolor{retained}{+ 0.1} & \textcolor{retained}{+ 1.3} & \textcolor{retained}{+ 1.6} & \textcolor{retained}{+ 1.8} & \textcolor{retained}{+ 6.6} & \textcolor{retained}{+ 4.4} & \textcolor{retained}{+ 4.0}  \\

        \midrule 

        SigLIP~\cite{SigLIP} - ViT-B/16 & 75.8  & 97.3  & 90.5  & 62.3 & 62.8  & 44.6 & 43.6 & 85.5 & 91.5  & 94.1  & 41.6  & 69.5  & 74.9  & 75.5  & 71.8 \\ 
        CuPL~\cite{CuPL} &  76.0 & 98.0 & 90.5 & 63.0 & 64.9 & 42.8 & 45.1 & 87.0 & 90.7 & 94.5 & 43.5 & 69.9 & 73.4 & 75.7 & 72.3 \\
        {\textbf{ProAPO} (ours)} & \textbf{76.4} & \textbf{98.3} & \textbf{91.7} & \textbf{66.2} & \textbf{69.1} & \textbf{55.8} & \textbf{47.1} & \textbf{93.3} & \textbf{92.2} & \textbf{94.9} & \textbf{44.3} & \textbf{71.7} & \textbf{75.9} & \textbf{78.8}  & \textbf{75.2} \\
        $\Delta$ & \textcolor{retained}{+ 0.6} & \textcolor{retained}{+ 1.0} & \textcolor{retained}{+ 1.2} & \textcolor{retained}{+ 3.9} & \textcolor{retained}{+ 6.3} & \textcolor{retained}{+ 11.2} & \textcolor{retained}{+ 3.5} & \textcolor{retained}{+ 7.8} & \textcolor{retained}{+ 0.7} & \textcolor{retained}{+ 0.8} & \textcolor{retained}{+ 2.7} & \textcolor{retained}{+ 2.2} & \textcolor{retained}{+ 1.0} & \textcolor{retained}{+ 3.3} & \textcolor{retained}{+ 3.4} \\
        
        \bottomrule
    \end{tabular}
}
\vskip -0.04in
  \caption{\textbf{Results of our ProAPO on different backbones.} \textbf{Avg (11)} and \textbf{Avg (13)} denote average results across 11 datasets (excluding CUB~\cite{CUB} and Places~\cite{Places365}) and all 13 datasets, respectively. $\Delta$ denotes performance gains compared to vanilla VLMs.}
  \label{supp_tab: results_different_backbones}
  \vskip -0.15in
\end{table*}
}

\section{Results on Different Backbones}
\label{sec_supp: different_backbone_result}

\textbf{Settings}. In~\cref{supp_tab: results_different_backbones}, we show results of our ProAPO in different backbones, including ResNet50, ResNet101, ViT-B/32, ViT-B/16, ViT-L/14 for CLIP~\cite{CLIP}, ViT-B/32 for OpenCLIP~\cite{OpenCLIP}, ViT-B/16 for EVA02~\cite{EVA-02}, and ViT-B/16 for SigLIP~\cite{SigLIP}. We compare our ProAPO with vanilla VLMs and the SOTA description method CuPL~\cite{CuPL}.

\textbf{Results}. We see that our ProAPO consistently improves vanilla CLIP and CuPL in thirteen datasets across all backbones. Compared to vanilla VLMs, our ProAPO enhances them by at least 3.4\% average accuracy in thirteen datasets. Moreover, we see notable performance improvement on several fine-grained datasets, such as DTD~\cite{DTD}, ESAT~\cite{EuroSAT}, FLO~\cite{FLO}, and UCF~\cite{UCF101}. It further verifies that class-specific descriptions provide helpful knowledge for fine-grained recognition. Besides, iterative optimization by our ProAPO also enhances the description method CuPL.

\textbf{More interesting findings}.
We find that as the backbones of VLMs become larger, the performance improvement by ProAPO gradually decreases. For example, from ViT-B/32 to ViT-B/16 to ViT-L/14, the gain for CLIP is from 5.7\% to 5.6\% to 5.0\%. Moreover, similar results appear in different models with the same backbone, \textit{i.e.}, the vanilla model with better results achieves a lower performance increase. For example, from CLIP~\cite{CLIP} to OpenCLIP~\cite{OpenCLIP} on ViT-B/32 backbone, the gain is from 5.7\% to 4.7\%, and from CLIP~\cite{CLIP} to EVA02~\cite{EVA-02} to SigLIP~\cite{SigLIP}, the gain is from 5.6\% to 4.0\% to 3.4\%. We argue that the model with the higher result has more knowledge, which may be affected less by prompt quality. Overall, our ProAPO continues to improve the performance of VLMs.

\section{More Comparisons with SOTA Methods}
\label{supp_sec: more_comparison_with_sota_methods}
In this section, we compare our ProAPO with more SOTA prompt tuning methods. These methods adapt VLMs from both visual and textual views.

{
\renewcommand{\arraystretch}{1.1} 
\setlength{\tabcolsep}{3.8pt}

\begin{table*}[htbp]
  \centering
  \resizebox{0.75\linewidth}{!}
    {
    \begin{tabular}
        {l | ccccc ccccc c | c}
            
        \toprule
        \textbf{Module} (ViT-B/16) & \rotatebox{90}{\textbf{IN-1K}} & \rotatebox{90}{\textbf{Caltech}} & \rotatebox{90}{\textbf{Cars}} & \rotatebox{90}{\textbf{DTD}}  & \rotatebox{90}{\textbf{ESAT}} & \rotatebox{90}{\textbf{FGVC}} & \rotatebox{90}{\textbf{FLO}} & \rotatebox{90}{\textbf{Food}}  &  \rotatebox{90}{\textbf{Pets}} & \rotatebox{90}{\textbf{SUN}} & \rotatebox{90}{\textbf{UCF}} & \rotatebox{90}{\textbf{Avg (11)}} \\
        \midrule

        Vanilla CLIP~\cite{CLIP} & 66.9  & 93.2  & 65.5 & 44.3  & 51.0  & 24.4  & 70.6  & 88.4  & 89.0   & 62.5  & 67.7  & 65.8  \\

        \midrule
        \multicolumn{13}{c}{\textit{\textbf{\ccol{Test-Time Prompt Tuning Methods}}}} \\
        \midrule 

        TPT~\cite{TPT} & 69.0 & 94.2 & 66.9 & 47.8 & 42.4 & 24.8 & 69.0 & 84.7 & 87.8 & 65.5 &  68.0 & 65.5   \\ 
        DiffTPT~\cite{DiffTPT} & 70.3 & 92.5 & 67.0 & 47.0 & 43.1 & 25.6 & 70.1 & 87.2 &  88.2 & 65.7 & 68.2 & 65.9    \\ 
        PromptAlign~\cite{PromptAlign} & 71.4 & 94.0 & 68.5 & 47.2 & 47.9 & 24.8 & 72.4 & 86.7 &  90.8 & 67.5 & 69.5 & 67.3  \\
        Self-TPT-v~\cite{Self_TPT_v} & \textbf{73.0} & 94.7 &  68.8 & 49.4 & 51.9 & 27.6 & 71.8 &  85.4 & 91.3 & 68.2 &  69.5 & 68.3  \\
        
        \midrule 
        \multicolumn{13}{c}{\textit{\textbf{\ccol{Vector-based Prompt Tuning Methods}}}} \\
        \midrule 
        UPT~\cite{prompt_tuning_UPT} & 69.6 & 93.7 & 67.6 & 45.0 & 66.5 & 28.4 & 75.0 & 84.2 & 82.9 & 68.8 & 72.0 & 68.5   \\
        CoCoOp~\cite{CoCoOp} & 69.4 & 93.8 & 67.2 & 48.5 & 55.3 & 12.7 & 72.1 & 85.7 & 91.3 & 68.3 & 70.3 & 66.8  \\
        MaPLe~\cite{MaPLe}  & 69.6 & 92.6 & 66.6 & 52.1 & 71.8 & 26.7 & 83.3 & 80.5 &  89.1 & 64.8 & 71.8 & 69.9   \\
        ALIGN~\cite{prompt_tuning_align} & 69.8 & 94.0 & 68.3 & 54.1 & 53.2 & 29.6 & 81.3 & 85.3 & 91.4 & 69.1 & 74.4 & 70.1   \\ 
        PromptSRC~\cite{PromptSRC} & 68.1 & 93.7 & 69.4 & 56.2 & \underline{73.1} & 27.7 & \underline{85.9} & 84.9 & 92.0 & 69.7 & 74.8 & 72.3    \\ 
        
        \midrule 
        \multicolumn{13}{c}{\textit{\textbf{\ccol{Description-Based Methods}}}} \\
        \midrule 
        
        \multicolumn{13}{l}{\textit{\textbf{w/o adapters}}} \\
        CuPL~\cite{CuPL} & 69.6  & 94.3  & 66.1   & 53.8  & 55.7  & 26.6  & 73.9  & 88.9  & 91.2  & 69.0  & 70.3  & 69.0  \\

        AWT-text~\cite{AWT} & 68.9  & 95.2  & 66.0   & 52.0  & 52.6  & 26.1  & 74.5  & 89.4  & 91.2   & 68.4  & 69.8 & 68.6  \\
        
        \highlight{\textbf{ProAPO} (ours)} & \highlight{69.9} & \highlight{95.2} & \highlight{67.7} & \highlight{55.8} & \highlight{65.3} & \highlight{28.3} & \highlight{82.7} & \highlight{89.5} & \highlight{92.7} & \highlight{68.9} & \highlight{73.1} & \highlight{71.7}  \\

        \highlight{\textbf{ProAPO} w/ AWT-text} & \highlight{69.4} & \highlight{\underline{95.3}} & \highlight{67.8} &  \highlight{54.3} & \highlight{67.1} & \highlight{27.4} & \highlight{82.1} & \highlight{\underline{89.6}} & \highlight{\underline{93.2}} & \highlight{68.5} & \highlight{73.1} & \highlight{71.6} \\

        \midrule 

         \multicolumn{13}{l}{\textit{\textbf{w/ adapters}}} \\

        AWT-Adapter~\cite{AWT} & \underline{72.1} & 95.1 & \textbf{73.4} & \underline{59.4} & \textbf{76.3} & \textbf{33.9} & 85.6 &  85.9 & 92.9 & \textbf{72.7} & \textbf{78.4} & \underline{75.1} \\
        
        \highlight{\textbf{ProAPO} w/ APE~\cite{APE}} & \highlight{71.3} & \highlight{\textbf{95.8}} & \highlight{\underline{70.9}} & \highlight{\textbf{60.6}} & \highlight{72.4} & \highlight{\underline{33.2}} & \highlight{\textbf{91.4}} & \highlight{\textbf{89.9}} & \highlight{\textbf{93.4}} & \highlight{\underline{71.0}} & \highlight{\underline{77.6}}  & \highlight{\textbf{75.2}} \\

        \bottomrule
    \end{tabular}
}
\vskip -0.04in
  \caption{\textbf{Comparison of our ProAPO with more SOTA methods under one-shot supervision.} \textbf{Avg (11)} denote average results across 11 datasets. }
  \label{supp_tab: comparison_with_more_SOTA}
\end{table*}
}

\textbf{Comparison of test-time prompt tuning methods}.
In~\cref{supp_tab: comparison_with_more_SOTA}, our ProAPO outperforms SOTA test-time prompt tuning methods on 11 datasets. Notably, we adapt VLMs solely from the textual view, while TPT methods introduce textual and visual views (\textit{i.e.}, augmented images), which further verifies the effectiveness of our method.

\textbf{Comparison of vector-based prompt-tuning methods}
 Since recent prompt-tuning methods adapt VLMs using both visual and textual views, we combine ProAPO with an adapter (\textit{i.e.}, APE~\cite{APE}) for a fair comparison. 
 \textbf{(1) Higher performance in low-shot.} In~\cref{supp_tab: comparison_with_more_SOTA}, ProAPO consistently outperforms these methods, which verifies that optimizing prompts in natural language is more effective in low-shot tasks. 
 \textbf{(2) Better transferability and interpretability}. Unlike vector-based prompt-tuning methods that search in a continuous space, ProAPO benefits from the discrete nature of natural language, leading to better interpretability and easily transfers across different backbones (shown in~\cref{tab: transfer_backbone}).
 \textbf{(3) Lower performance in high-shot}.  However, in~\cref{supp_tab: shots_influence}, ProAPO shows a sub-optimal result compared to CoOp~\cite{CoOp} in high-shot settings. This is due to the limited language search space and iteration steps.

\textbf{Comparison of AWT~\cite{AWT}}.
First, since AWT uses augmented visual and textual views to adapt VLMs, we compare ProAPO with AWT under the augmented textual view for a fair comparison. In~\cref{supp_tab: comparison_with_more_SOTA}, the result shows our ProAPO improves AWT-text by 6.1\% on average, verifying that our progressive optimization improves prompt quality. In addition, we introduce a common adapter-based method to our ProAPO and compare it with AWT-Adapter in the one-shot setting. We see that our ProAPO achieves comparable results. These results suggest that ProAPO and AWT are complementary.

\textbf{Comparison of iCM~\cite{iCM}}. iCM is somewhat similar to ours, optimizing class-specific prompts with chat-based LLMs. However, it uses the whole validation set as supervision. In~\cref{supp_tab: comparison_of_iCM}, we see that our ProAPO outperforms iCM significantly even under the one-shot supervision. 
This is because our ProAPO address challenges in class-specific prompt optimization by an offline generation algorithm to reduce LLM querying costs, an entropy-constrained fitness score to prevent overfitting, and two sampling strategies to find an optimal initial point and reduce iteration times.


{
\begin{table*}[h]
  \centering
  \resizebox{0.8\linewidth}{!}
    {
    \begin{tabular}
        {l | ccccc ccc | c}  
        \toprule
        
        \textbf{Module} (ViT-B/32) & {\textbf{IN-1K}} & {\textbf{Caltech}} & {\textbf{CUB}} & {\textbf{DTD}}  & {\textbf{ESAT}} & {\textbf{FLO}} & {\textbf{SUN}} & {\textbf{UCF}} & \textbf{Avg (8)} \\
        \midrule
        Vanilla CLIP & 62.1 & 91.2 & 51.7 & 42.9 & 43.9 &  66.0 & 62.1 & 60.9 & 60.1 \\
        \midrule
        \multicolumn{10}{c}{\textit{\textbf{\ccol{Automatic Prompt Optimization Methods}}}} \\
        \midrule
        
        iCM~\cite{iCM} (w/ validation set) & 64.5 & 92.7 & \textbf{56.1} & 51.4 & 56.3 & 72.2 & 66.2 & 67.0 & 65.8  \\
        \highlight{\textbf{ProAPO} (w/ 1-shot)} &  \highlight{\textbf{64.7}} & \highlight{\textbf{94.4}} & \highlight{55.4} & \highlight{\textbf{53.5}} & \highlight{\textbf{63.0}} & \highlight{\textbf{74.3}} & \highlight{\textbf{66.6}} & \highlight{\textbf{69.0}} & \highlight{\textbf{67.6}} \\

        \bottomrule
    \end{tabular}
}
  \caption{\textbf{Comparison of our ProAPO with iCM~\cite{iCM}.} Avg (8) denotes average results across 8 datasets.}
  \label{supp_tab: comparison_of_iCM}
\end{table*}
}

{
\renewcommand{\arraystretch}{1.1} 

\begin{table*}[htbp]
  \centering
  \resizebox{0.99\linewidth}{!}
    {
    \begin{tabular}
        {l | lllll lllll lll | l | l}

        \toprule
        \textbf{Module} (ResNet50) & \rotatebox{90}{\textbf{IN-1K}} & \rotatebox{90}{\textbf{Caltech}} & \rotatebox{90}{\textbf{Cars}} & \rotatebox{90}{\textbf{CUB}} & \rotatebox{90}{\textbf{DTD}}  & \rotatebox{90}{\textbf{ESAT}} & \rotatebox{90}{\textbf{FGVC}} & \rotatebox{90}{\textbf{FLO}} & \rotatebox{90}{\textbf{Food}}  &  \rotatebox{90}{\textbf{Pets}} & \rotatebox{90}{\textbf{Places}} & \rotatebox{90}{\textbf{SUN}} & \rotatebox{90}{\textbf{UCF}} & \rotatebox{90}{\textbf{Avg (11)}} & \rotatebox{90}{\textbf{Avg (13)}} \\
        \midrule


        Vanilla CLIP & 57.9 & 84.5 & 53.9 & 44.7 & 38.8 & 28.6 & 15.9 & 60.2 & 74.0 & 83.2 & 38.2 & 58.0 & 56.9 & 55.6 & 53.4 \\ 
        
        \midrule
        \multicolumn{16}{c}{\textit{\textbf{\ccol{Template Optimization Methods}}}} \\

        \midrule 
        PN~\cite{P_N} & 59.6 & 89.1 & 56.2 & - & 44.8 & \underline{49.0} & 18.1 & 67.2 & 78.3 & 88.1 & - & 61.0 & 60.2 & 61.1 & -  \\

        \highlight{\textbf{ATO} (w/o dataset domain)} & \highlight{60.4} & \highlight{88.9} & \highlight{56.8} & \highlight{47.0} & \highlight{45.0} & \highlight{43.7} & \highlight{17.9} & \highlight{67.4} & \highlight{79.9} & \highlight{87.8} & \highlight{40.0} & \highlight{61.2} & \highlight{61.5} & \highlight{61.0} & \highlight{58.3} \\
        
        \highlight{\textbf{ATO}} & \highlight{\underline{61.3}} & \highlight{89.4} & \highlight{57.4} & \highlight{49.2} &  \highlight{45.4} & \highlight{46.4} & \highlight{18.4} & \highlight{68.1} & \highlight{80.5} & \highlight{88.5} & \highlight{40.2} &  \highlight{61.8} & \highlight{63.9} & \highlight{61.9}  & \highlight{59.3} \\

        \midrule

        \multicolumn{16}{c}{\textit{\textbf{\ccol{Description Optimization Methods}}}} \\

        \midrule
        \highlight{\textbf{ProAPO} (w/o synonyms)} & \highlight{\textbf{61.5}} & \highlight{\underline{89.7}} & \highlight{\textbf{58.3}} & \highlight{\underline{49.7}} & \highlight{\underline{46.6}} & \highlight{46.8} & \highlight{\underline{20.5}} & \highlight{\underline{74.6}} & \highlight{\underline{81.0}} & \highlight{\textbf{88.8}} & \highlight{\underline{40.9}} & \highlight{\underline{62.3}} & \highlight{\underline{64.8}} & \highlight{\underline{63.2}} & \highlight{\underline{60.4}} \\        
        
        \highlight{\textbf{ProAPO} (ours)} & \highlight{\textbf{61.5}} & \highlight{\textbf{90.3}} & \highlight{\underline{58.0}} & \highlight{\textbf{50.7}} & \highlight{\textbf{52.3}} & \highlight{\textbf{51.7}} & \highlight{\textbf{21.1}} & \highlight{\textbf{75.1}} & \highlight{\textbf{81.8}} & \highlight{\underline{88.7}} & \highlight{\textbf{41.8}} & \highlight{\textbf{63.7}} & \highlight{\textbf{66.0}} & \highlight{\textbf{64.6}}  & \highlight{\textbf{61.8}} \\
        
        \bottomrule
    \end{tabular}
}
  
  \caption{\textbf{Ablation of template and description optimization.} 
  Avg (11) and Avg (13) denote average results across 11 datasets (excluding CUB~\cite{CUB} and Places~\cite{Places365}) and all 13 datasets, respectively. ATO denotes our automatic template optimization algorithm.}
  \vspace{-4pt}
  \label{supp_tab: ablate_template_and_description}
\end{table*}
}

\section{More Ablation Results}
\label{sec_supp: more_ablation_result}

\subsection{Ablation of Template and Description Optimization}
\label{supp_sec: ablation_template_and_description}

In~\cref{supp_tab: ablate_template_and_description}, we ablate key components in template and description optimization on the ResNet50 backbone. 

\textbf{(1) Ablation of Template Optimization}. In the main paper (Sec. 4.3), we show that prompt ensembling is better than a single prompt. Moreover, dataset domain information also plays a significant role in template optimization. Without domain information, we see a performance drop in our ATO by an average of 1.0\% (from 58.3\% to 59.3 \%) on thirteen datasets. This is because domain information provides contextual information, which can mitigate issues of semantic ambiguity caused by class names.

\textbf{(2) Ablation of Description Optimization}. Without label synonyms to increase description diversity, a performance degradation appears by an average of 1.4\% (from 60.4\% to 61.8\%) on thirteen datasets. It verifies the effectiveness of optimization class names, which are usually ignored in previous description methods~\cite{CuPL, DCLIP, AdaptCLIP, GPT4Vis}.

\textbf{(3) Template VS Description Optimization}. Compared with template optimization, we see a notable performance improvement with description optimization, especially in CUB~\cite{CUB}, DTD~\cite{DTD}, ESAT~\cite{EuroSAT}, FLO~\cite{FLO}, and UCF~\cite{UCF101} datasets. It demonstrates that optimizing class-specific prompts can find discriminative information for fine-grained classification.

\subsection{More Ablation of Operators}
\label{supp_sec: more_ablation_operator}
To further explore whether each operator has a role in searching the optimal result, we show the number of each operator causing the new optimal score during the iterations in~\cref{supp_tab: more_ablation_operator}. We see that each operator in iterative optimization may generate a better prompt. It further demonstrates that each operator is helpful in ProAPO. Notably, the crossover operator has the highest times to update the optimal score, which demonstrates that it makes the model search for the optimal prompt faster with limited iterations. 

{
\renewcommand{\arraystretch}{1.1} 
\begin{table}[t]
  \centering
  \resizebox{0.96\linewidth}{!}
    {
    \begin{tabular}
        {l | c  c  c  c  c | c  }  
        \toprule
        {\textbf{Dataset}} & \texttt{Add} & \texttt{Del} & \texttt{Rep} & \texttt{Cross} & \texttt{Mut} & \textbf{Total} \\
        \midrule
        IN-1K~\cite{Imagenet} & 3 & 4 & 5 & 5 & 2 &  19 \\
        Caltech~\cite{caltech101} & 5 & 5 & 6 & 12 & 3 & 31 \\
        Cars~\cite{Cars} & 7 & 8 & 5 & 8 & 3 & 31 \\ 
        CUB~\cite{CUB} & 9 & 4 & 10 & 6 & 2 & 31 \\
        DTD~\cite{DTD} & 5 & 3 & 8 & 8 & 2 & 26 \\
        ESAT~\cite{EuroSAT} & 2 & 4 & 6 & 8 & 1 & 21 \\
        FGVC~\cite{FGVC} & 6 & 2 & 6 & 5 & 3 & 22 \\ 
        FLO~\cite{FLO} & 5 & 3 & 11 & 5 & 4 & 28 \\
        Food~\cite{Food101} & 5 & 3 & 4 & 5 & 2 & 19 \\ 
        Pets~\cite{oxford_pets} & 4 & 2 & 5 & 6 & 2 & 19 \\ 
        Places~\cite{Places365} & 3 & 2 & 8 & 12 & 4 & 29 \\ 
        SUN~\cite{SUN} & 4 & 2 & 3 & 5 & 2 & 16 \\ 
        UCF~\cite{UCF101} & 5 & 6 & 8 & 6 & 2 & 27 \\ 
        \midrule
        \textbf{Sum} & 63 & 48 & 85 & 91 & 32 & 319 \\ 
        \bottomrule
    \end{tabular}
}
  \caption{\textbf{Number of times for each operator that update the optimal score.} \textbf{Total} denotes the total number of iterations when achieving the highest score.}
  \label{supp_tab: more_ablation_operator}
\end{table}
}

{
\renewcommand{\arraystretch}{1.1} 
\begin{table*}[t]
  \centering
  \resizebox{0.99\linewidth}{!}
    {
    \begin{tabular}
        {l | ccccc ccccc ccc | c | c | c }  
        \toprule
        {\textbf{Module} (ViT-B/32)}  & \rotatebox{90}{\textbf{IN-1K}} & \rotatebox{90}{\textbf{Caltech}} & \rotatebox{90}{\textbf{Cars}} & \rotatebox{90}{\textbf{CUB}} & \rotatebox{90}{\textbf{DTD}}  & \rotatebox{90}{\textbf{ESAT}} & \rotatebox{90}{\textbf{FGVC}} & \rotatebox{90}{\textbf{FLO}} & \rotatebox{90}{\textbf{Food}}  &  \rotatebox{90}{\textbf{Pets}} & \rotatebox{90}{\textbf{Places}} & \rotatebox{90}{\textbf{SUN}} & \rotatebox{90}{\textbf{UCF}} & \rotatebox{90}{\textbf{Avg (11)}} & \rotatebox{90}{\textbf{Avg (13)}} & \textbf{Times}  \\
        \midrule
        CuPL & 64.4  & 92.9  & 60.7  & 53.3  & {50.6}  & 50.5  & 20.9  & 69.5  & 84.2  & 87.0  & \underline{43.1}  & {66.3}  & 66.4  & 64.9  & 62.3 & - \\
        \midrule
        
        \texttt{a)} w/ all categories in one group & 64.5 & 93.3 & 60.9 & 53.5 & \underline{51.6} & 52.2 & 22.2 & 70.8 & 84.5 & 87.9 & 42.3 & \textbf{66.7} & \textbf{69.4} & 65.8  & 63.1 & \textbf{20 min} \\
        \texttt{b)} w/ random selected group & 64.3 & 93.7 & \textbf{61.8} & \underline{55.2} & 48.7 & 59.5 & 22.6 & 72.9 & \underline{85.2} & \underline{90.8} & 42.6 & 65.4 & 68.4 & 66.7  & 63.9 & 15 min  \\
        \texttt{c)} w/ performance best group & 64.1 & 93.0 & 61.2 & 54.4 & 47.4 & 56.8 & 20.7 & 68.2 & 85.1 & 88.6 & 42.4 & 65.0 & 65.4 & 65.0  & 62.5 & 15 min \\
        \texttt{d)} w/ K-Means algorithm & \underline{64.6} & \underline{93.8} & \textbf{61.8} & 55.1 & 49.4 & \underline{59.6} & \underline{22.8} & \underline{74.0} & \textbf{85.3} & 90.7 & {42.7} & 65.4 & \underline{69.0} & \underline{67.0}  & \underline{64.2} & \underline{17 min} \\
        
        \midrule

        \highlight{\textbf{ProAPO} (full model)} & \highlight{\textbf{64.7}}  & \highlight{\textbf{94.4}} & \highlight{\underline{61.7}} & \highlight{\textbf{55.4}} & \highlight{\textbf{53.5}} & \highlight{\textbf{63.0}} & \highlight{\textbf{23.0}} & \highlight{\textbf{74.3}} & \highlight{\textbf{85.3}} & \highlight{\textbf{91.0}} & \highlight{\textbf{43.3}} & \highlight{\underline{66.6}} & \highlight{\underline{69.0}} & \highlight{\textbf{67.9}}  & \highlight{\textbf{65.0}} & \highlight{15 min} \\
        \bottomrule
    \end{tabular}
}
    \vspace{-5pt}
  \caption{\textbf{More ablation of group sampling strategy.} We ablate the ways for selecting salient groups. \textbf{Times} denotes the time that ProAPO runs on ImageNet with the default setting.}
  \vspace{-7pt}
  \label{supp_tab: more_ablation_group_sampling}
\end{table*}

}

\subsection{More Ablation of Group Sampling}
\label{supp_sec: more_ablation_group_sampling}

In~\cref{supp_tab: more_ablation_group_sampling}, we ablate how to select categories in the group sampling strategy. We consider the settings for optimizing all categories in one group, selecting random categories and the best categories with their misclassified categories in groups. In rows a)-c) of~\cref{supp_tab: more_ablation_group_sampling}, we see notable performance degradation compared to full ProAPO. It further demonstrates that optimizing salient and worst groups can achieve comparable results with all categories and save iteration costs. Moreover, we also consider replacing misclassified categories with a K-Means clustering algorithm. A performance drop appears in row d), which verifies the effectiveness of selecting misclassified categories in groups.

\subsection{Ablation of Cost Computation}
\label{supp_sec: ablation_of_cost_computation}

In~\cref{supp_tab: extra_computation_cost}, we detail the time each process consumes on ImageNet. Compared to previous LLM-generated description methods, we similarly query LLMs one-time to generate descriptions (\textit{i.e.}, process of building prompt library). In addition, we introduce iterative processes to refine prompts and two sampling strategies to save costs. With a few additional costs (15 min v.s. 60 min), our ProAPO improves previous methods by at least 2.7\% on average. This further verifies the efficiency of our method.

{
\renewcommand{\arraystretch}{1.0} 
\setlength{\tabcolsep}{3.8pt}
\begin{table}[!h]
  \centering
  \resizebox{1.0\linewidth}{!}
    {
    \begin{tabular}
        {l | c | c c c c }  
        \toprule
        \textbf{Process} & \textbf{Build Library} & \highlight{\textbf{Sample Strategy}} & \highlight{\textbf{Template Optim.}} & \highlight{\textbf{Description Optim.}} \\ 
        \midrule 
        \textbf{Times} & 60 min &  \highlight{3 min}  & \highlight{1.6 min} & \highlight{10.4 min} \\
        \bottomrule
    \end{tabular}
}
\vspace{-5pt}
  \caption{\textbf{Computation cost analysis in the ImageNet dataset.}}
  \label{supp_tab: extra_computation_cost}
\end{table}
}

\section{More Hyperparameter Analysis}
\label{sec_supp: more_hyper_analysis}

{
\renewcommand{\arraystretch}{1.1} 
\setlength{\tabcolsep}{4.pt}
\begin{table}[tbp]
  \centering
  \resizebox{0.99\linewidth}{!}
    {
    \begin{tabular}
        {l | l | c | c c c c c | c }

        \toprule
        {\textbf{Dataset}} & \textbf{Module} & {\textbf{TF}} & \multicolumn{5}{c}{\textbf{Number of training samples}} & \textbf{UB} \\
        \cmidrule(lr){4-8}
         & (RN50) & &  1 & 2 & 4 & 8 & 16\\
        \midrule
        \multirow{2}{*}{\textbf{Avg (11)}} & CoOp~\cite{CoOp} & \xmark & 59.6 & 62.3 & \textbf{66.8} & \textbf{69.9} & \textbf{73.4} & -  \\
         & {\textbf{ProAPO}} & \cmark & \textbf{64.6} & \textbf{65.0} & 65.4 & 65.8 & 66.1 & 67.2 \\
        \midrule
        
        \multirow{2}{*}{\textbf{IN-1K}} & CoOp~\cite{CoOp} & \xmark  & 57.2 & 57.8 & 60.0 & \textbf{61.6} & \textbf{63.0} & -  \\
        & {\textbf{ProAPO}} & \cmark & \textbf{61.5} & \textbf{61.6} & \textbf{61.5} & \textbf{61.6} & 61.6 & 61.7 \\
        \midrule
        
        \multirow{2}{*}{\textbf{Caltech}} & CoOp~\cite{CoOp} & \xmark  & 87.5 & 87.9 & 89.6 & 90.2 & 91.8 & -  \\
        & {\textbf{ProAPO}} & \cmark & \textbf{90.3} & \textbf{90.4} & \textbf{90.6} & \textbf{90.7} & \textbf{91.0} & 91.1 \\        
        \midrule
        
        \multirow{2}{*}{\textbf{Cars}} & CoOp~\cite{CoOp} & \xmark & 55.6 & 58.3 & \textbf{62.6} & \textbf{68.4} & \textbf{73.4} & -  \\
        & {\textbf{ProAPO}} & \cmark & \textbf{58.0} & \textbf{58.5} & 58.8 & 58.9 & 59.1 & 60.8  \\
        \midrule
        
        \multirow{2}{*}{\textbf{DTD}} & CoOp~\cite{CoOp} & \xmark & 44.4 & 45.2 & \textbf{53.5} & \textbf{60.0} & \textbf{63.6} & -  \\
        & {\textbf{ProAPO}} & \cmark & \textbf{52.3} & \textbf{52.7} & 53.0 & 53.4 & 53.6 \\
        \midrule
        
        \multirow{2}{*}{\textbf{ESAT}} & CoOp~\cite{CoOp} & \xmark & 50.6 & \textbf{61.5} & \textbf{70.2} & \textbf{76.7} & \textbf{83.5} & -  \\
        & {\textbf{ProAPO}} & \cmark & \textbf{51.7} & 53.5 & 55.6 & 57.4 & 58.3 & 62.2 \\
        \midrule
        
        \multirow{2}{*}{\textbf{FGVC}} & CoOp~\cite{CoOp} & \xmark & 9.6 & 18.7 & \textbf{21.9} & \textbf{26.1} & \textbf{31.3} & -  \\
        & {\textbf{ProAPO}} & \cmark & \textbf{21.1} & \textbf{21.0} & 21.2 & 21.2 & 21.3 & 21.5 \\
        \midrule
        
        \multirow{2}{*}{\textbf{FLO}} & CoOp~\cite{CoOp} & \xmark & 68.1 & \textbf{77.5} & \textbf{86.2} & \textbf{91.2} & \textbf{94.5} & -  \\
        & {\textbf{ProAPO}} & \cmark & \textbf{75.1} & 75.6 & 76.4 & 76.7 & 77.8 & 79.1 \\
        \midrule
        
        \multirow{2}{*}{\textbf{Food}} & CoOp~\cite{CoOp} & \xmark  & 74.3 & 72.5 & 73.3 & 71.8 & 74.7 & -  \\
        & {\textbf{ProAPO}} & \cmark & \textbf{81.8} & \textbf{82.0} & \textbf{82.1} & \textbf{82.2} & \textbf{82.3} & {82.9} \\
        \midrule
        
        \multirow{2}{*}{\textbf{Pets}} & CoOp~\cite{CoOp} & \xmark  & 85.9 & 82.6 & 86.7 & 85.3 & 87.0 & -  \\
        & {\textbf{ProAPO}} & \cmark & \textbf{88.7} & \textbf{89.4} & \textbf{89.5} & \textbf{89.8} & \textbf{89.9} & {91.0} \\
        \midrule
        
        \multirow{2}{*}{\textbf{SUN}} & CoOp~\cite{CoOp} & \xmark  & 60.3 & 59.5 & 63.5 & \textbf{65.5} & \textbf{69.3} & -  \\
        & {\textbf{ProAPO}} & \cmark & \textbf{63.7} & \textbf{63.8} & \textbf{63.8} & 63.8 & 63.9 & 64.5 \\
        \midrule
        
        \multirow{2}{*}{\textbf{UCF}} & CoOp~\cite{CoOp} & \xmark  & 61.9 & 64.1 & 67.0 & \textbf{71.9} & \textbf{75.7} & -  \\
        & {\textbf{ProAPO}} & \cmark &  \textbf{66.0} & \textbf{66.8} & \textbf{67.1} & 68.1 & 68.9 & 71.4 \\
        \bottomrule
    \end{tabular}
}
  \vspace{-5pt}
  \caption{\textbf{Scaling up to more shots.} \textbf{Avg (11)} denotes average results across 11 datasets. \textbf{TF} denotes training-free approaches. \textbf{UB} denotes upper bound evaluated on the test set.}
  \label{supp_tab: shots_influence}
\end{table}
}

\subsection{Effect of Shot Numbers}
\label{supp_sec: effect_shots}
In~\cref{supp_tab: shots_influence}, we show the effect of the number of training samples per category.
Specifically, we conduct experiments with 1, 2, 4, 8, and 16 shots.
Moreover, we introduce the performance of the optimal prompt searched in the test set as the upper bound of ProAPO.
Compared with CoOp~\cite{CoOp}, ProAPO achieves remarkable performance when shots $\leq 2$, which demonstrates the effectiveness of our method under low-shot settings. Since we only adapt VLMs in a training-free way, the performance increases finitely as the training samples increase. We attribute two key directions for further performance improvement in high-shot settings. First, our result is still far from the upper bound (66.1 \% in 16 shots VS 67.2 \% for the upper bound). We need to improve the prompt generation algorithm and the score function to find better candidate prompts within the limited iterations. Second, the upper bound of our ProAPO is much smaller than the prompt tuning method. We need to use a larger natural language search space (\textit{e.g.}, more diverse descriptions, or more query times of LLMs) to further increase the upper bound of the optimal result.

\subsection{Effect of Scalar in Score Function}
\label{supp_sec: effect_alpha}

In~\cref{supp_tab: ablation_alpha},  we show the effect of $\alpha$ in~\cref{eq: score_function}. We see that performance improves as the $\alpha$ increases. This is because the entropy constraint provides more information to select better candidate prompts. We see a stable result when $\alpha \in [5e2, 5e3]$, which means a better trade-off between accuracy and entropy constraint. However, a high $\alpha$ may be biased to the train set, thus harming the performance. 

{
\begin{table}[h]
  \centering
  \resizebox{1.0\linewidth}{!}
    {
    \begin{tabular}
        {l | c c c c c c c c}  
        \toprule
        $\alpha$ & $0$ & $1e1$ & $1e2$ & $5e2$ & $1e3$ & $5e3$ & $1e4$ & $1e5$ \\
        \midrule
        \textbf{Avg (13)} & 62.3 & 63.4 & 64.4 & 64.9 & \textbf{65.0} & 64.8 & 63.7 & 63.1 \\
        \bottomrule
    \end{tabular}
}
  \caption{\textbf{Effect of $\alpha$ value in Eq.6 across 13 datasets.}}
  \label{supp_tab: ablation_alpha}
\end{table}
}

\subsection{Effect of Sampled Numbers in Prompt Sampling Strategy}
\label{supp_sec: effect_sampled_numbers}
\begin{figure}[htbp]
\centering
\includegraphics[width=0.8\linewidth]{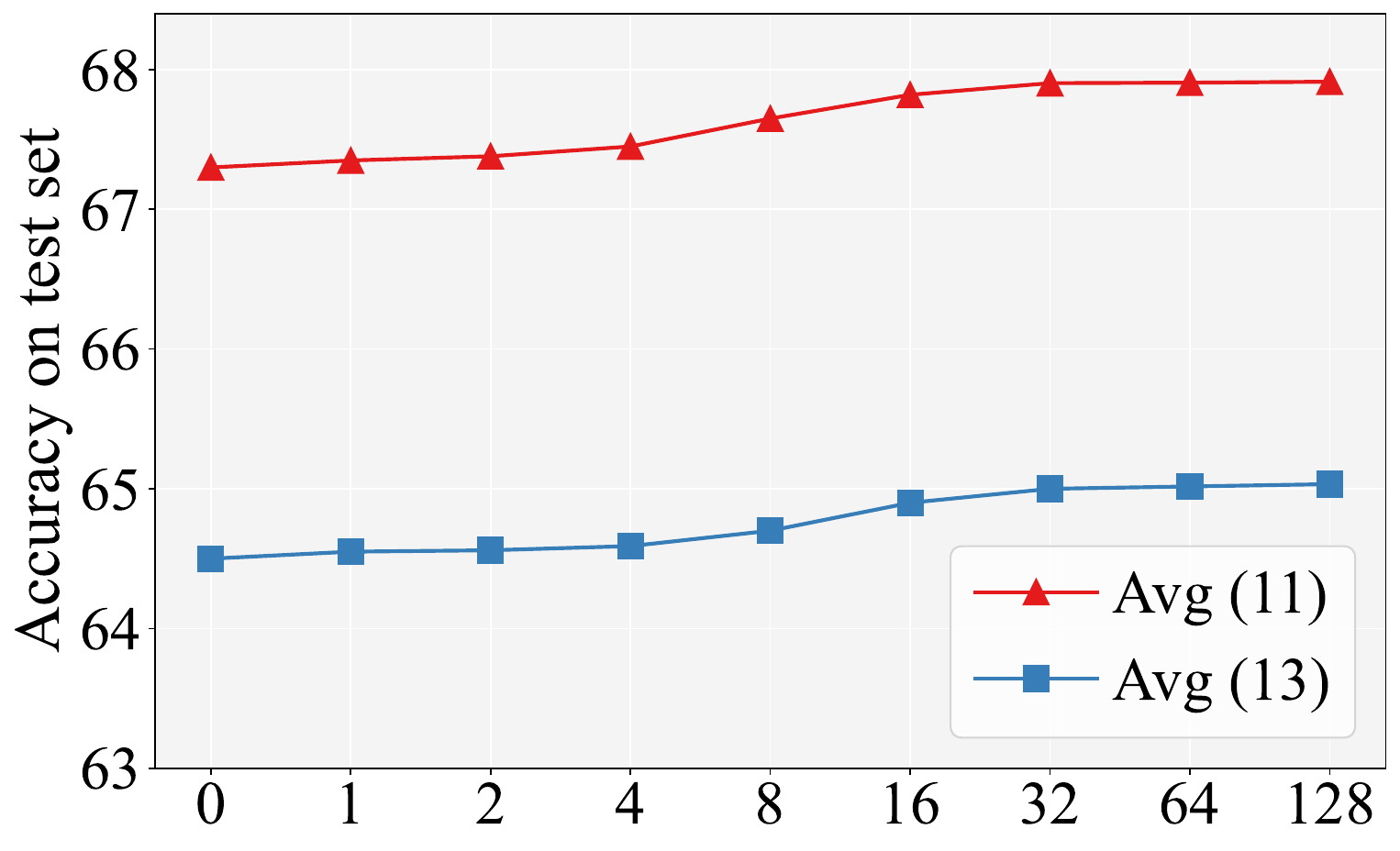}
\vspace{-8pt}
\caption{\textbf{Effect of sampled numbers $T_{sample}$.} 
}
\label{supp_fig: effect_sampled_numbers}
\vspace{-6pt}
\end{figure}

In~\cref{supp_fig: effect_sampled_numbers}, we show the effect of sampled numbers $T_{sample}$ of Alg.~\ref{supp_alg: prompt_strategy}. The $T_{sample} = 0$ means that the prompt sampling strategy is not used. As the number of $T_{sample}$ increases, we see a slight performance gain when $T_{sample} < 4$. After $T_{sample} \geq 4$, a consistent improvement appears because the initial search point achieves a higher score than the baseline. We achieve stable results when $T_{sample} \geq 32$.

\subsection{Effect of Quality of Prompt Library}
\label{supp_sec: effect_of_quality_of_prompt_library}
\begin{figure}[htbp]
\centering
\includegraphics[width=0.8\linewidth]{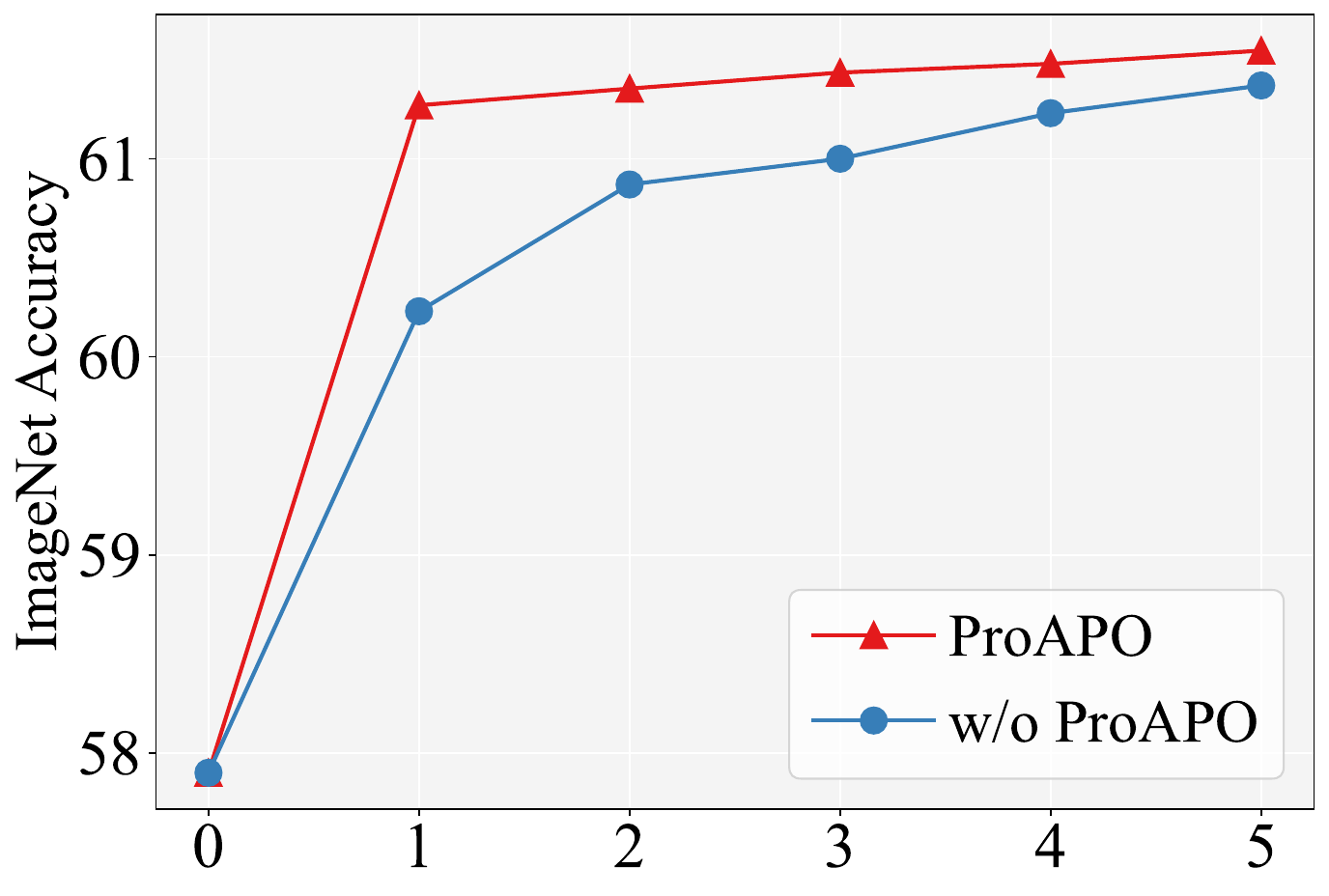}
\vspace{-8pt}
\caption{\textbf{Effect of Number of LLM-query Prompts.} 
}
\label{supp_fig: effect_LLM_Query}
\end{figure}

\begin{figure}[htbp]
\centering
\includegraphics[width=0.8\linewidth]{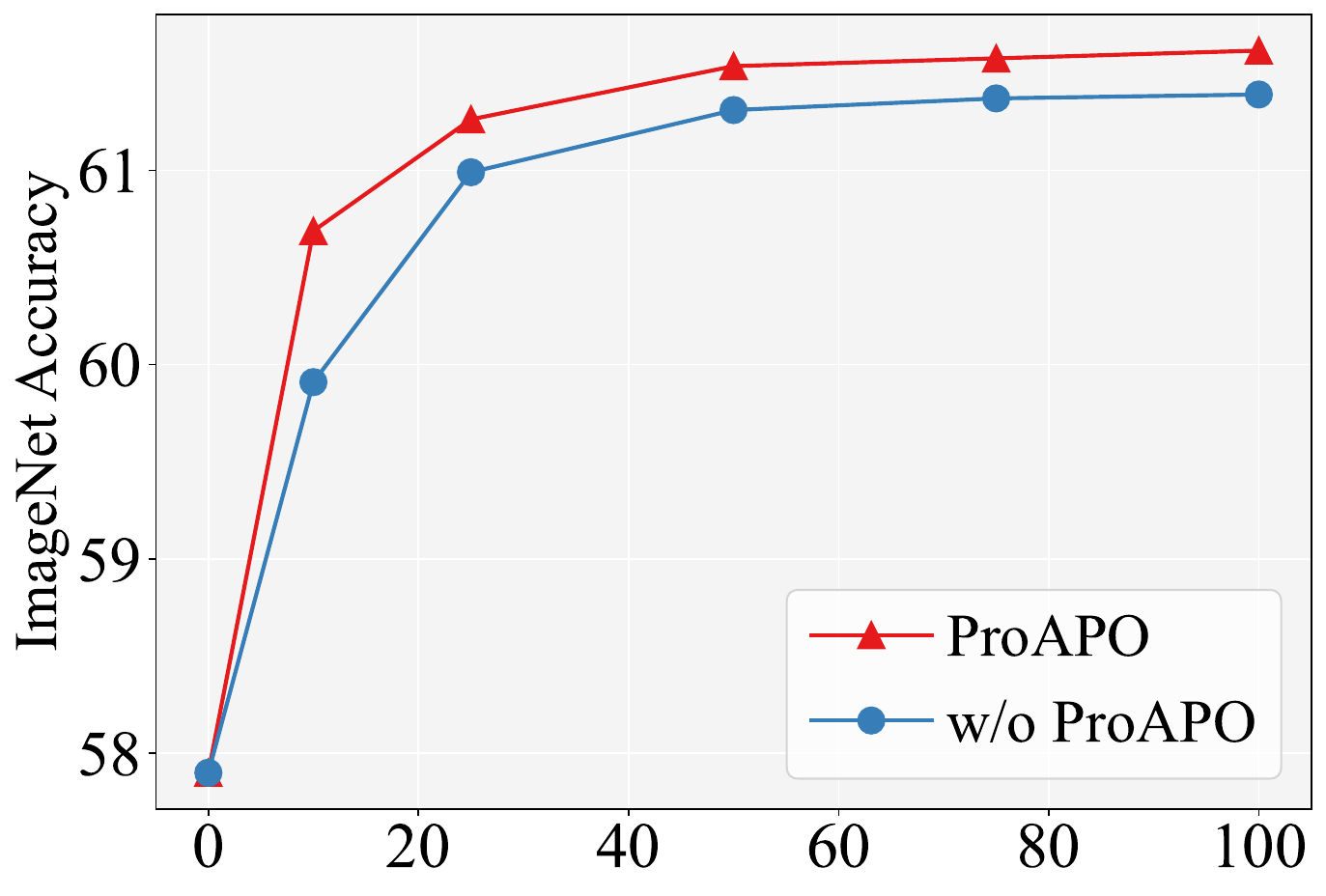}
\vspace{-8pt}
\caption{\textbf{Effect of Number of Generated Descriptions.} 
}
\label{supp_fig: effect_generated_descriptions}
\end{figure}

In~\cref{supp_fig: effect_LLM_Query} and~\cref{supp_fig: effect_generated_descriptions}, we analyze two key factors affecting the prompt library: LLM-query prompts and generated descriptions. Our ProAPO improves prompt quality even under a small number of query prompts and descriptions, demonstrating its effectiveness in a limited prompt library.

\section{More Qualitative Results}
\label{sec_supp: more_qualitative_result}
\begin{figure*}[htbp]
\centering
\includegraphics[width=0.8\linewidth]{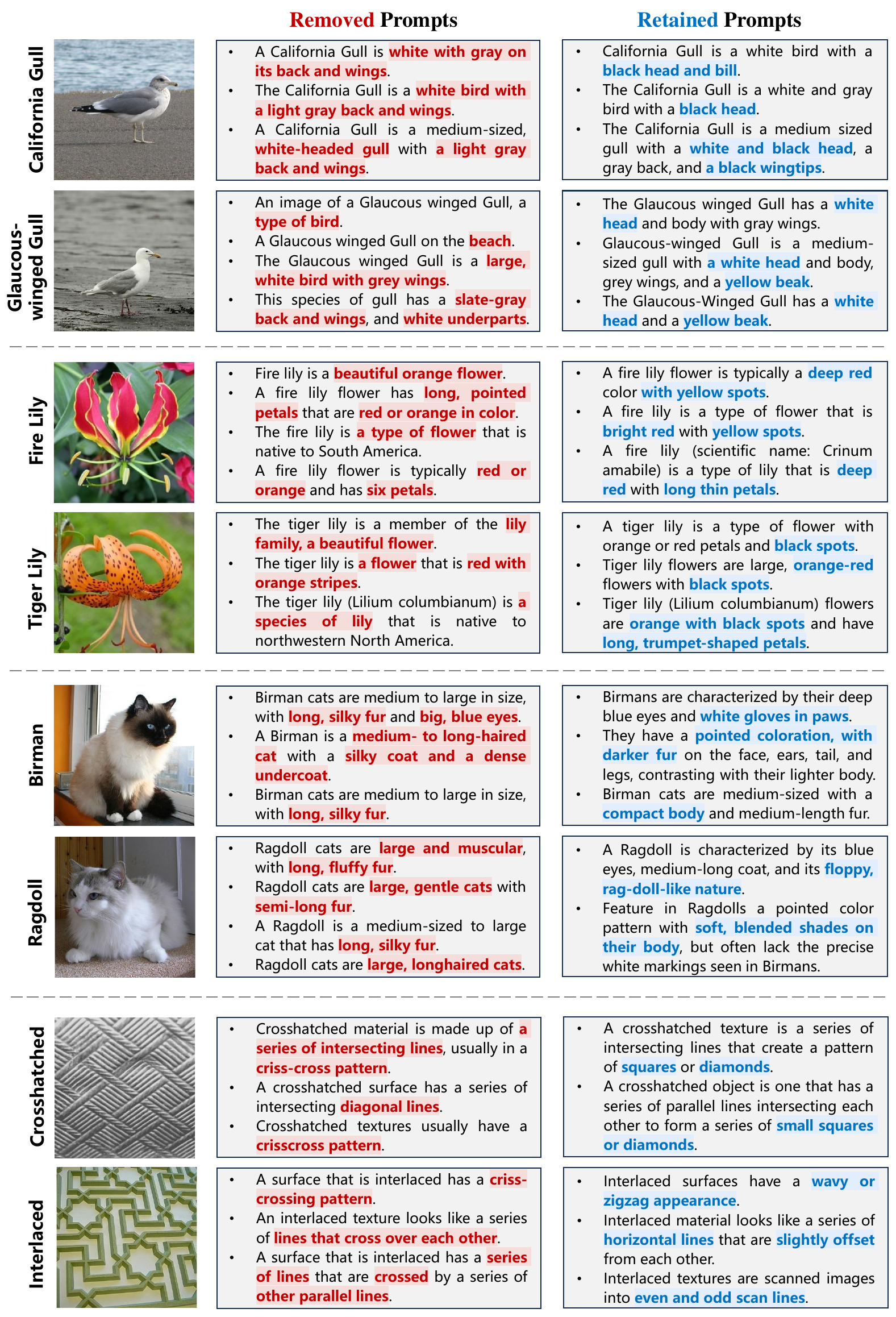}
\caption{\textbf{Qualitative analysis of class-specific prompt optimization by ProAPO.} Shaded \textbf{\textcolor{removed}{red}} and \textbf{\textcolor{retained}{blue}} words denote common and discriminative descriptions in two confused categories.
}
\label{supp_fig: qualitative_result}
\vspace{20pt}
\end{figure*}

In~\cref{supp_fig: qualitative_result}, we show more examples of the changes in descriptions with our ProAPO, including images of animals, flowers, and textures.
Similarly, we see that common descriptions are removed and discriminative ones are retained for fine-grained categories, which further verifies the effectiveness of our progressive optimization.

\newpage

\section{Detailed Results of More Benefits by Optimal Prompts}
\label{sec_supp: detailed_results_of_main_paper}

\subsection{Transfer to Adapter-based Methods}
\label{sec_supp: transfer_to_adapter}
In~\cref{supp_fig: improve_train_free_adapter} and ~\cref{supp_fig: improve_train_adapter}, we show the detailed results of popular training-free and training adapter-based methods~\cite{Tip, Tip-X, APE, CLIP_Adapter} with different prompt initialization, \textit{i.e.}, SOTA method CuPL~\cite{CuPL} and our ProAPO. Adapter-based methods with ProAPO (solid lines) consistently surpass those with CuPL (dotted lines). It reveals that high-quality prompts make adapters perform better. Even in low shots, training with ProAPO achieves notable performance gains, which further verifies its effectiveness.

{
\begin{figure*}[ht]
\centering
\begin{adjustbox}{minipage=\textwidth,scale=0.88}
\begin{subfigure}{0.33\textwidth}
    \includegraphics[width=\textwidth]{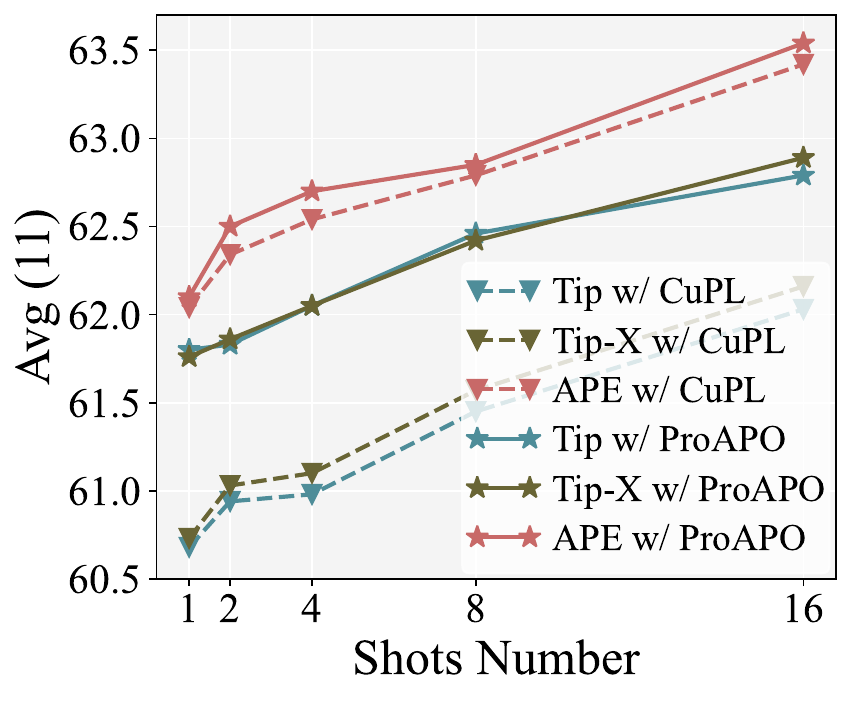}
    \caption{ImageNet.}
\end{subfigure}
\hfill
\begin{subfigure}{0.33\textwidth}
    \includegraphics[width=\textwidth]{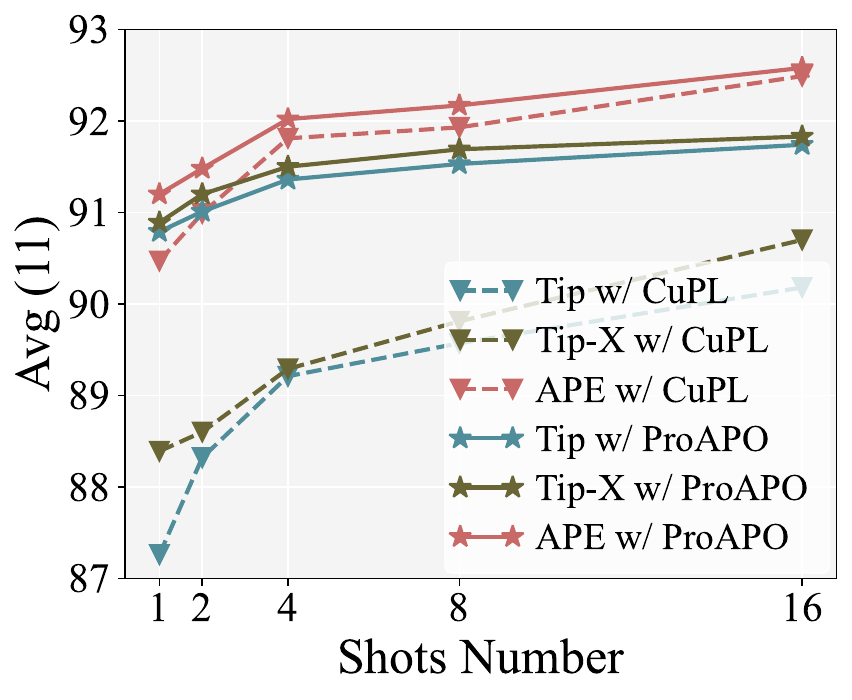}
    \caption{Caltech.}
\end{subfigure}
\hfill
\begin{subfigure}{0.33\textwidth}
    \includegraphics[width=\textwidth]{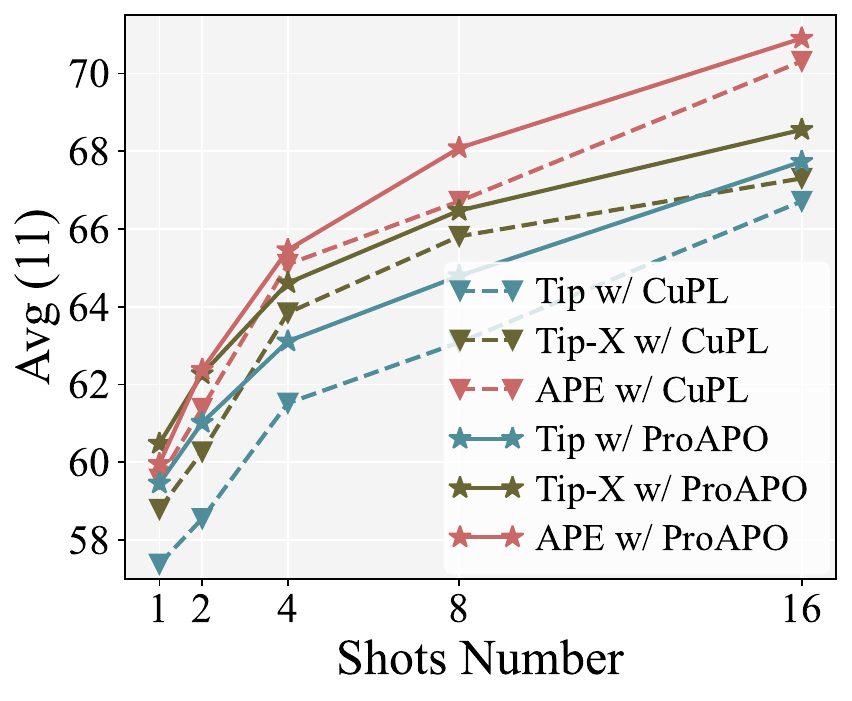}
    \caption{Cars.}
\end{subfigure}
\begin{subfigure}{0.33\textwidth}
    \includegraphics[width=\textwidth]{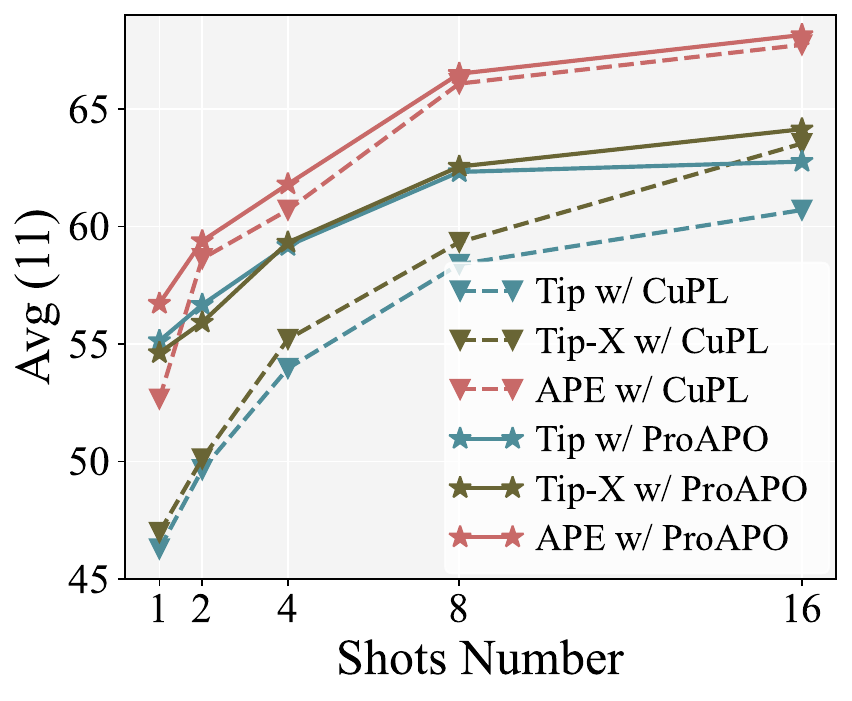}
    \caption{DTD.}
\end{subfigure}
\hfill
\begin{subfigure}{0.33\textwidth}
    \includegraphics[width=\textwidth]{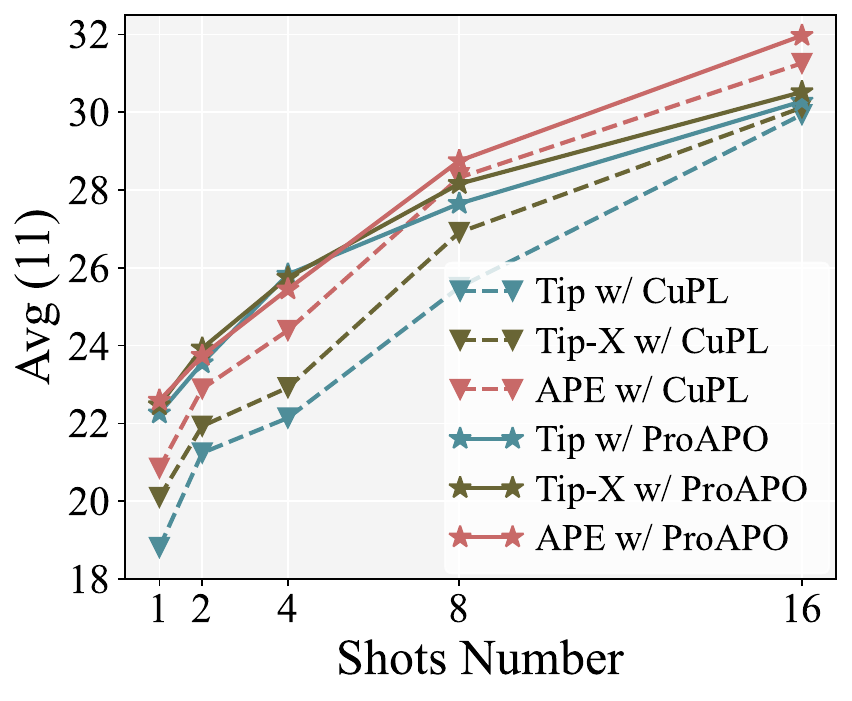}
    \caption{FGVC.}
\end{subfigure}
\hfill
\begin{subfigure}{0.33\textwidth}
    \includegraphics[width=\textwidth]{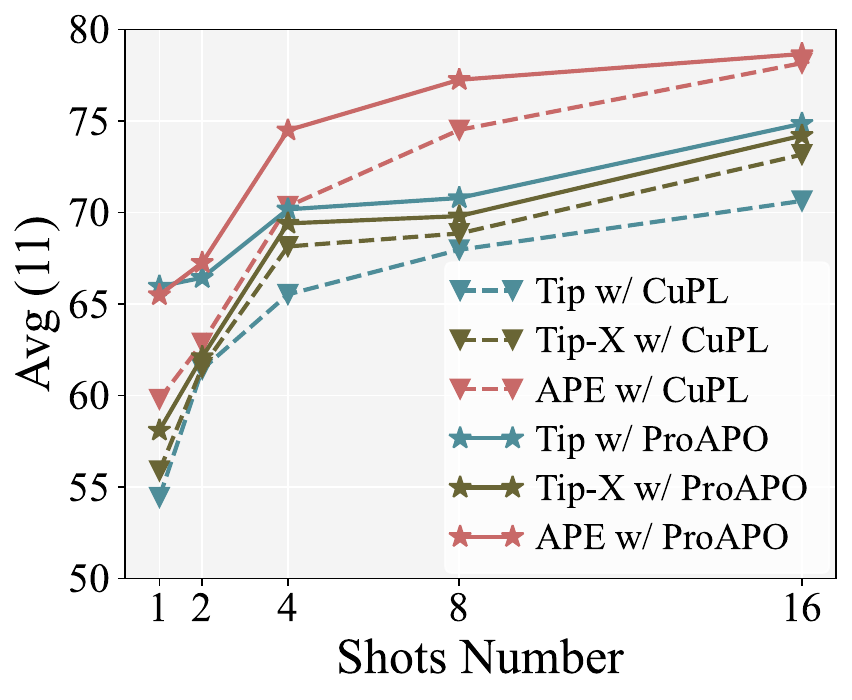}
    \caption{ESAT.}
\end{subfigure}

\begin{subfigure}{0.33\textwidth}
    \includegraphics[width=\textwidth]{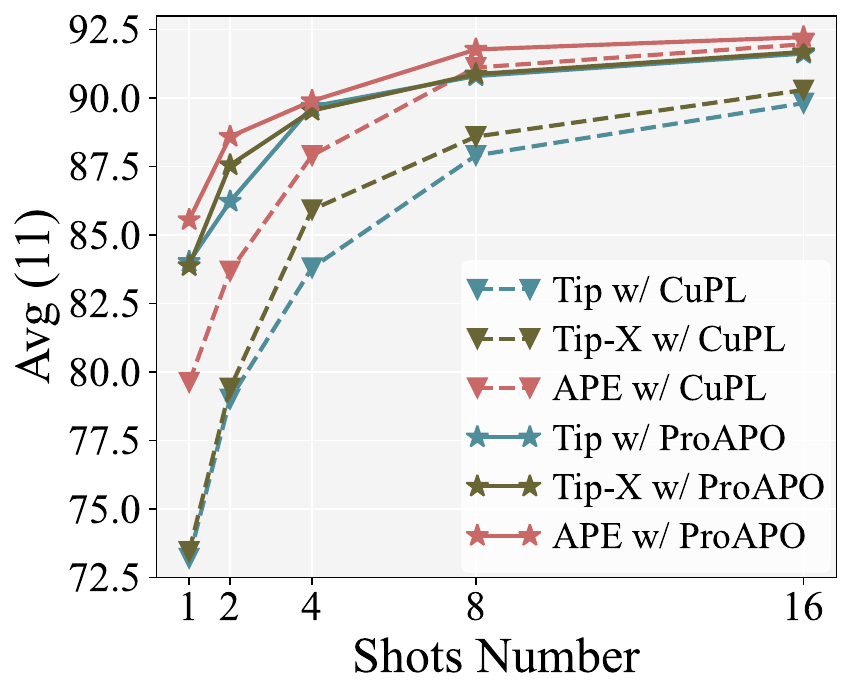}
    \caption{FLO.}
\end{subfigure}
\hfill
\begin{subfigure}{0.33\textwidth}
    \includegraphics[width=\textwidth]{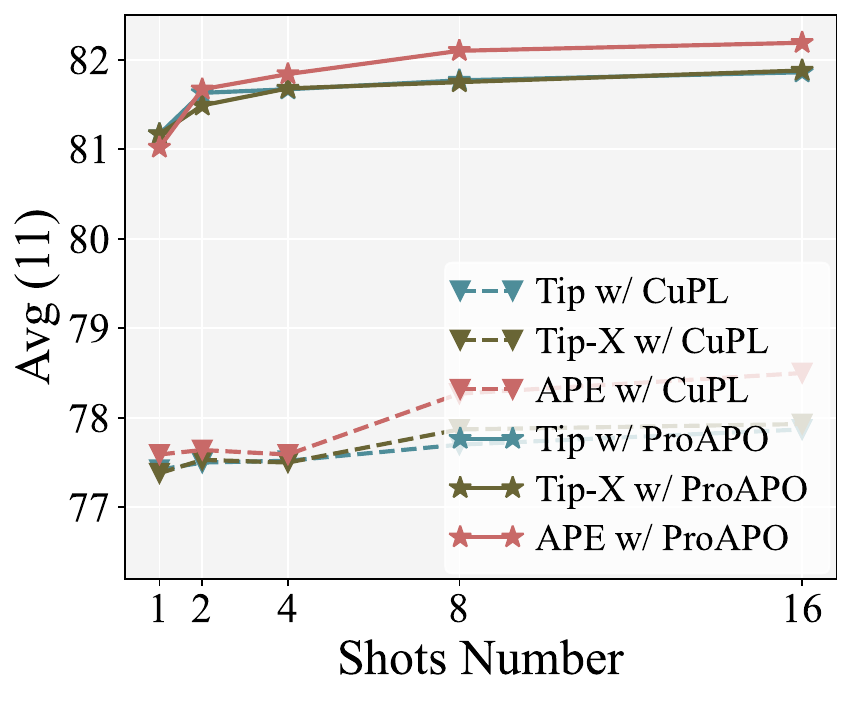}
    \caption{Food.}
\end{subfigure}
\hfill
\begin{subfigure}{0.33\textwidth}
    \includegraphics[width=\textwidth]{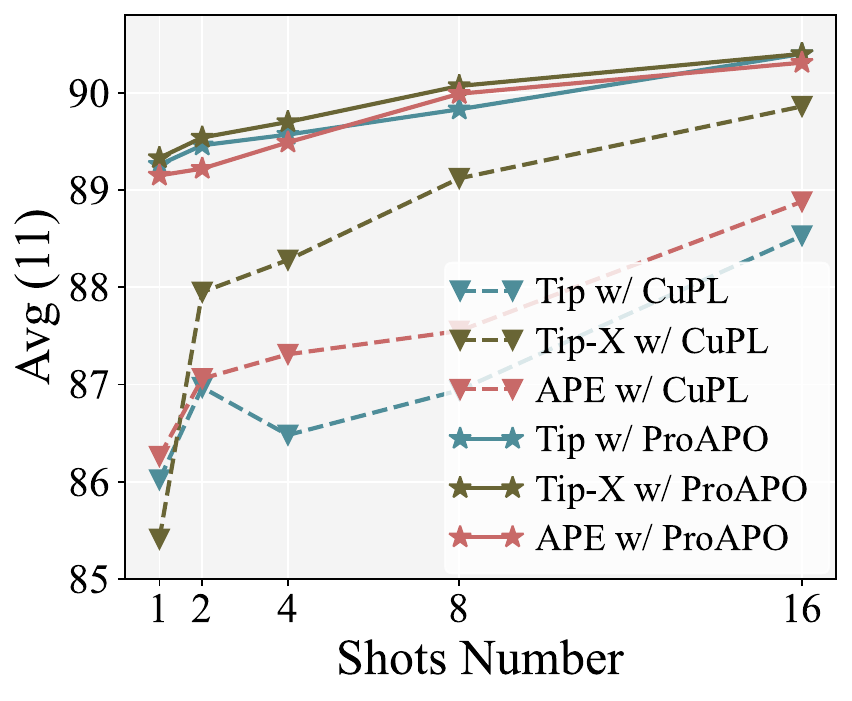}
    \caption{Pets.}
\end{subfigure}

\begin{subfigure}{0.33\textwidth}
    \includegraphics[width=\textwidth]{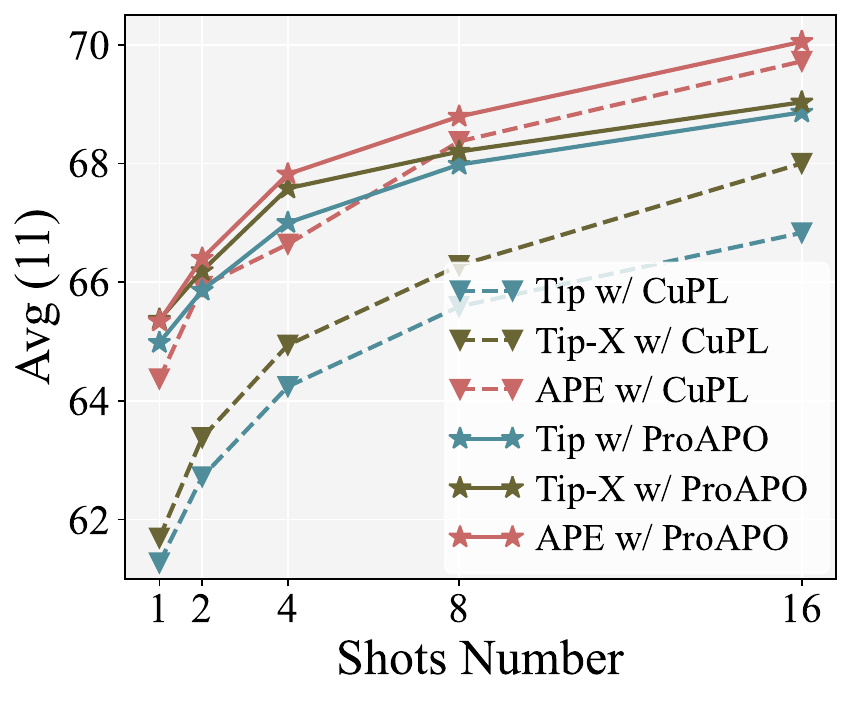}
    \caption{SUN.}
\end{subfigure}
\begin{subfigure}{0.33\textwidth}
    \includegraphics[width=\textwidth]{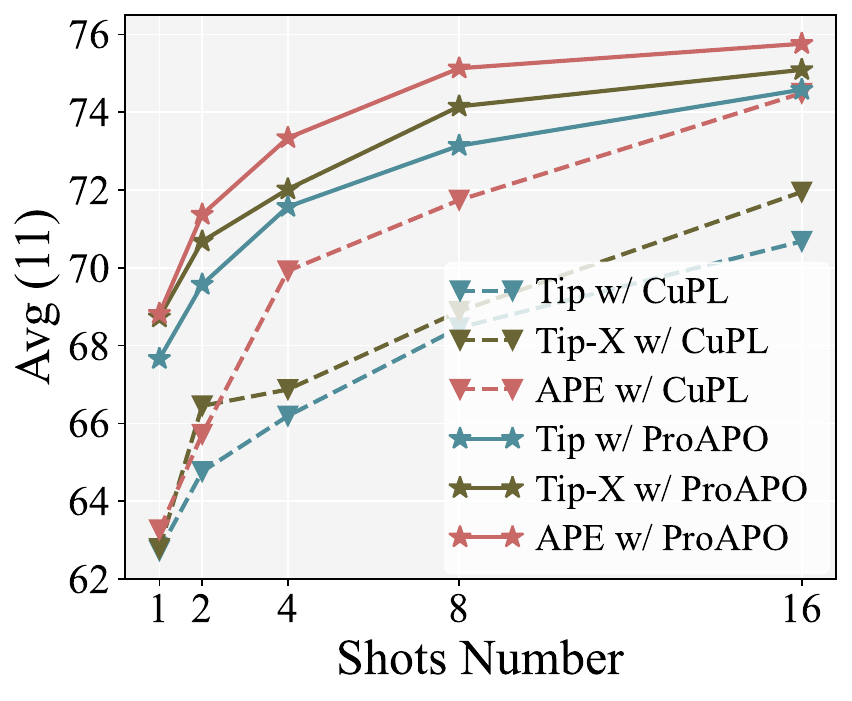}
    \caption{UCF.}
\end{subfigure}
\end{adjustbox}
\caption{\textbf{Results of training-free adapter-based methods with different initial prompts.} Solid and dotted lines denote prompt initialization with ProAPO and CuPL, respectively. We see that our ProAPO consistently improves adapter-based methods.}
\label{supp_fig: improve_train_free_adapter}
\vspace{60pt}
\end{figure*}
}

{
\begin{figure*}[ht]
\centering
\begin{adjustbox}{minipage=\textwidth,scale=0.88}
\begin{subfigure}{0.33\textwidth}
    \includegraphics[width=\textwidth]{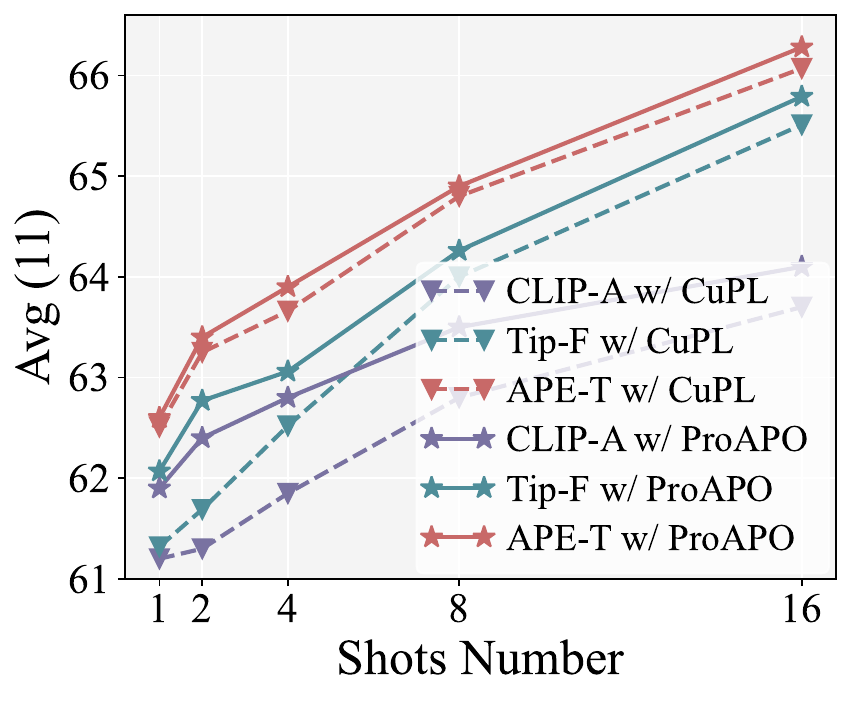}
    \caption{ImageNet.}
\end{subfigure}
\hfill
\begin{subfigure}{0.33\textwidth}
    \includegraphics[width=\textwidth]{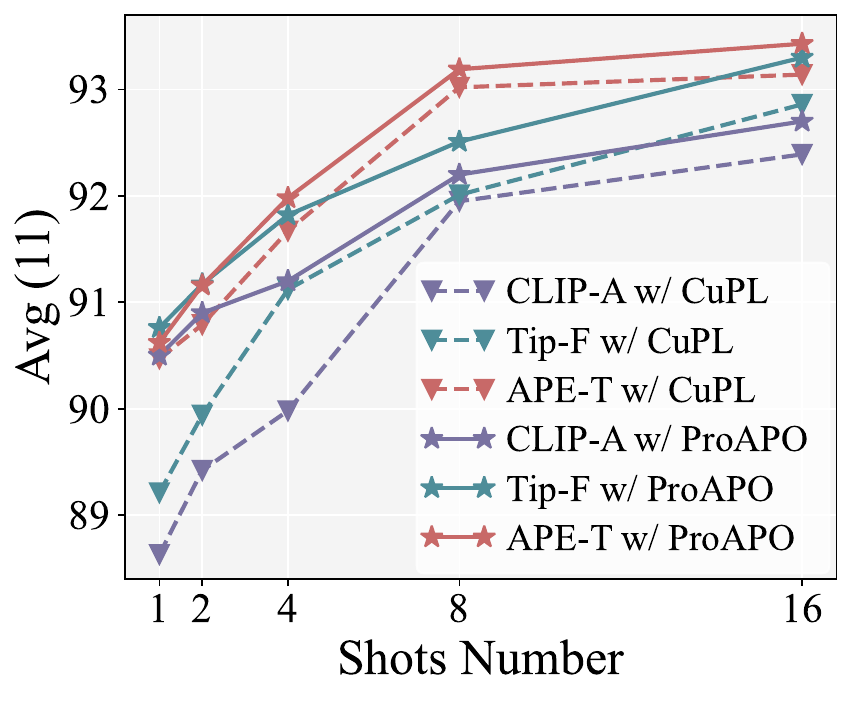}
    \caption{Caltech.}
\end{subfigure}
\hfill
\begin{subfigure}{0.33\textwidth}
    \includegraphics[width=\textwidth]{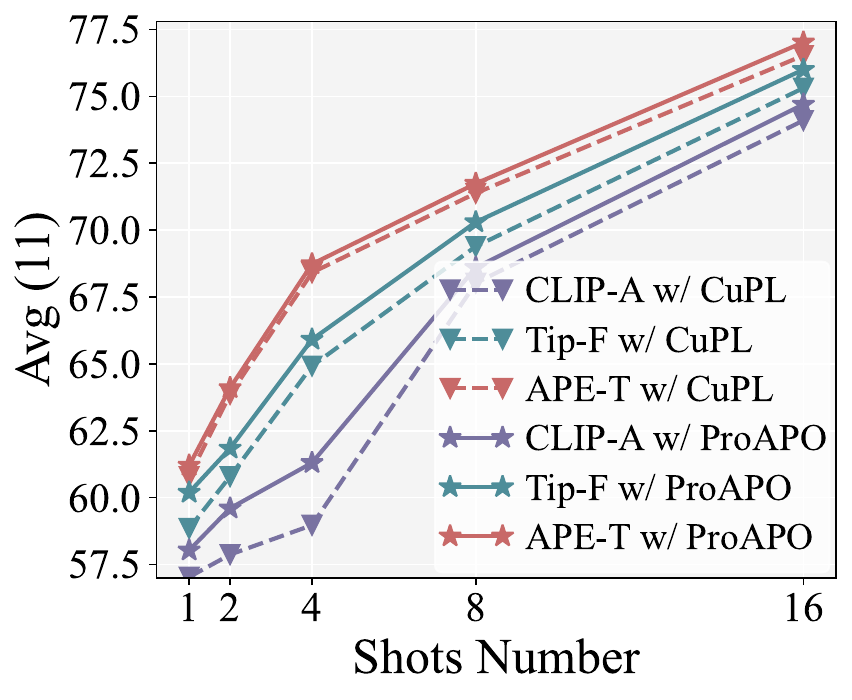}
    \caption{Cars.}
\end{subfigure}
\begin{subfigure}{0.33\textwidth}
    \includegraphics[width=\textwidth]{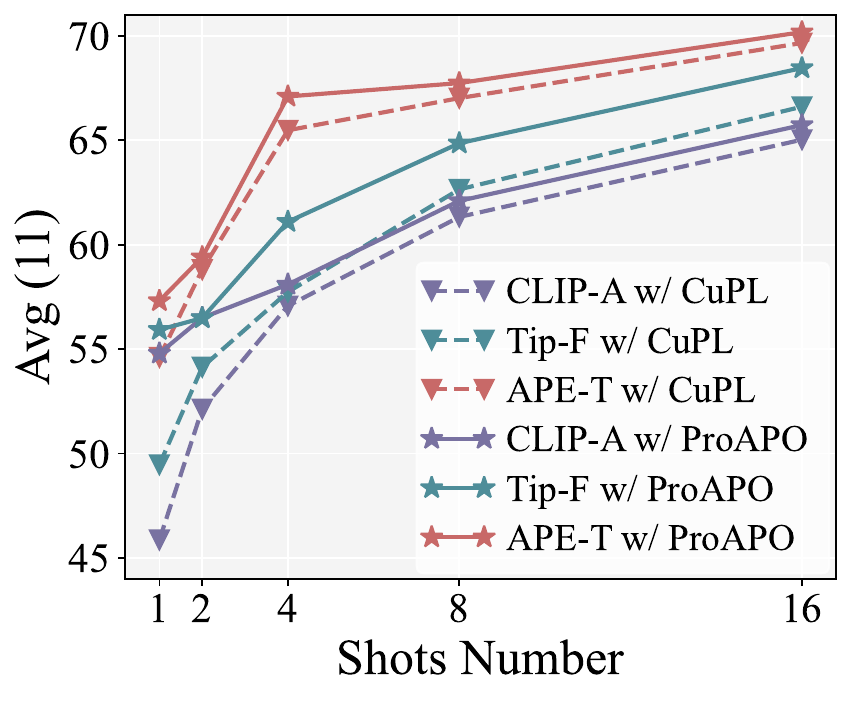}
    \caption{DTD.}
\end{subfigure}
\hfill
\begin{subfigure}{0.33\textwidth}
    \includegraphics[width=\textwidth]{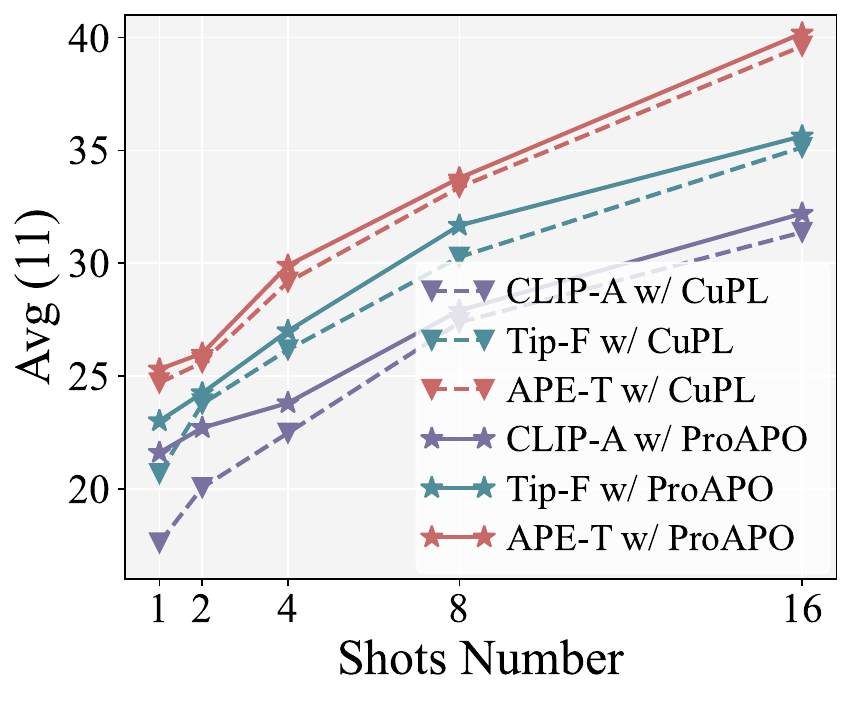}
    \caption{FGVC.}
\end{subfigure}
\hfill
\begin{subfigure}{0.33\textwidth}
    \includegraphics[width=\textwidth]{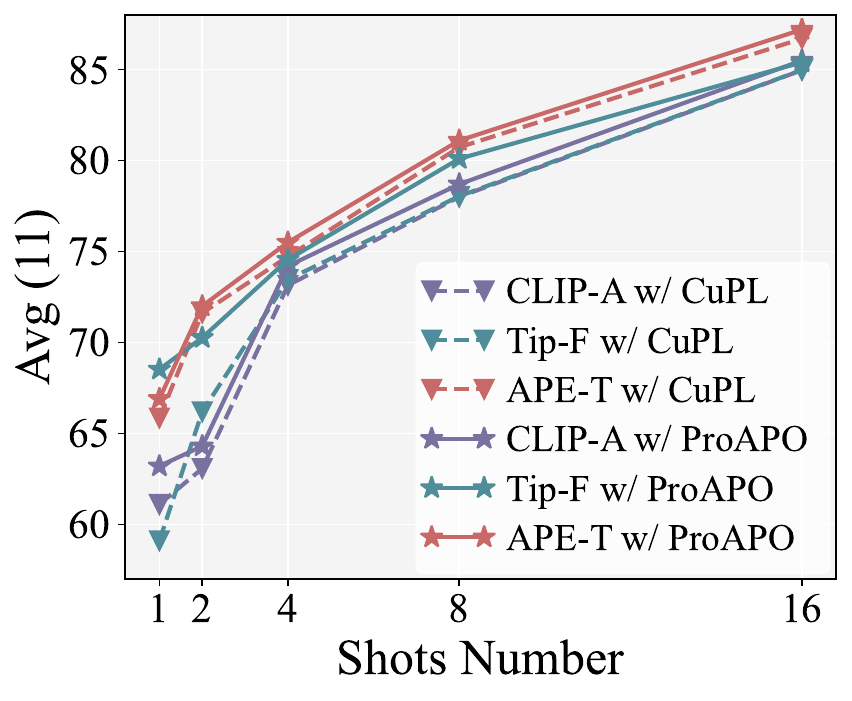}
    \caption{ESAT.}
\end{subfigure}

\begin{subfigure}{0.33\textwidth}
    \includegraphics[width=\textwidth]{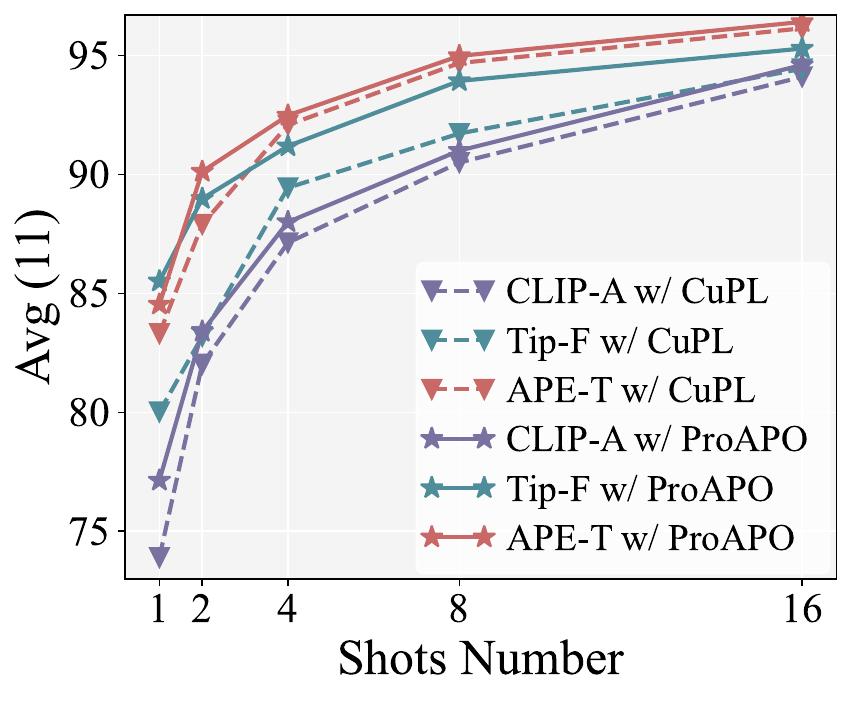}
    \caption{FLO.}
\end{subfigure}
\hfill
\begin{subfigure}{0.33\textwidth}
    \includegraphics[width=\textwidth]{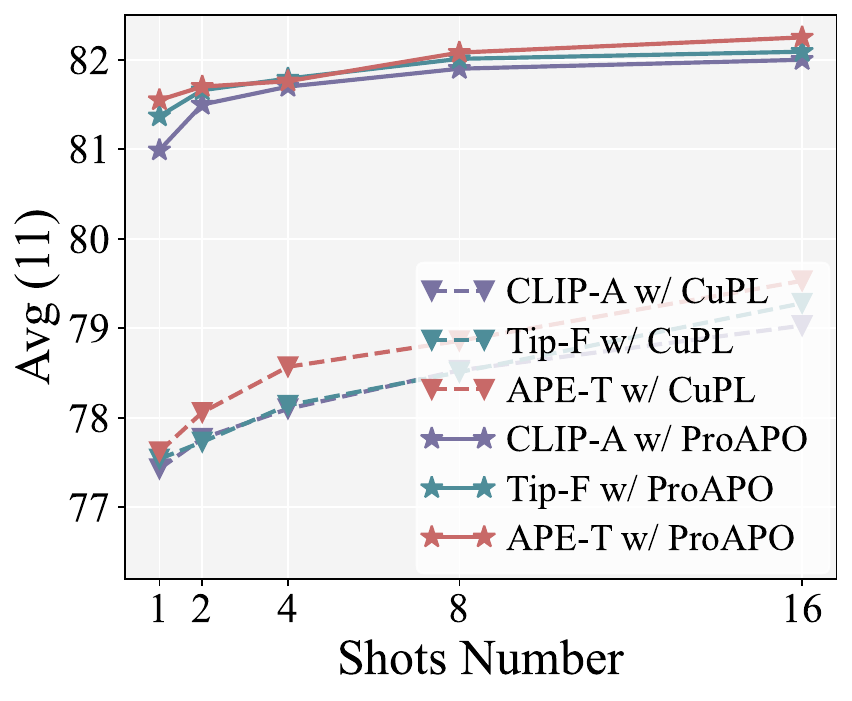}
    \caption{Food.}
\end{subfigure}
\hfill
\begin{subfigure}{0.33\textwidth}
    \includegraphics[width=\textwidth]{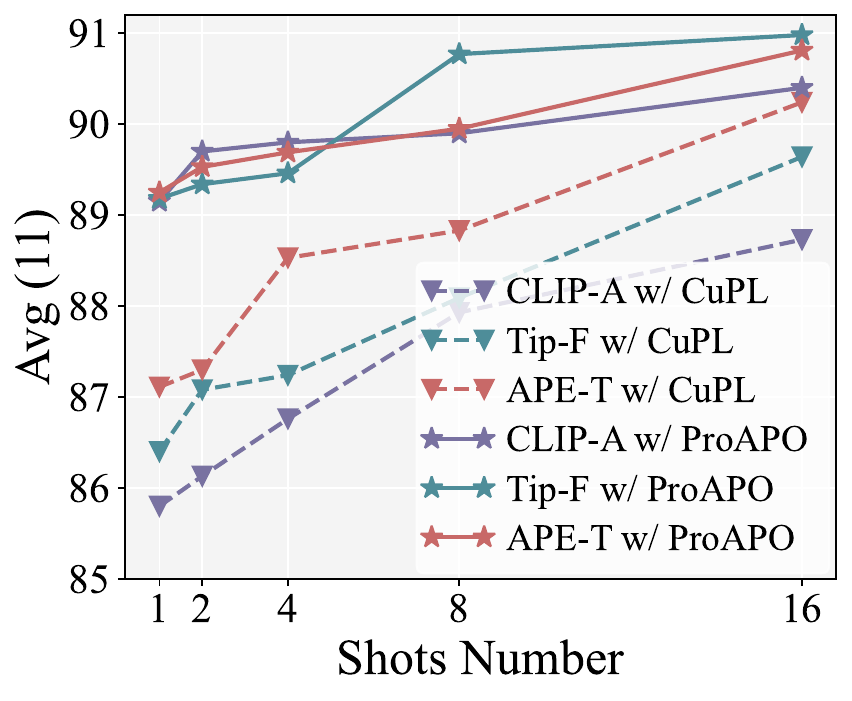}
    \caption{Pets.}
\end{subfigure}

\begin{subfigure}{0.33\textwidth}
    \includegraphics[width=\textwidth]{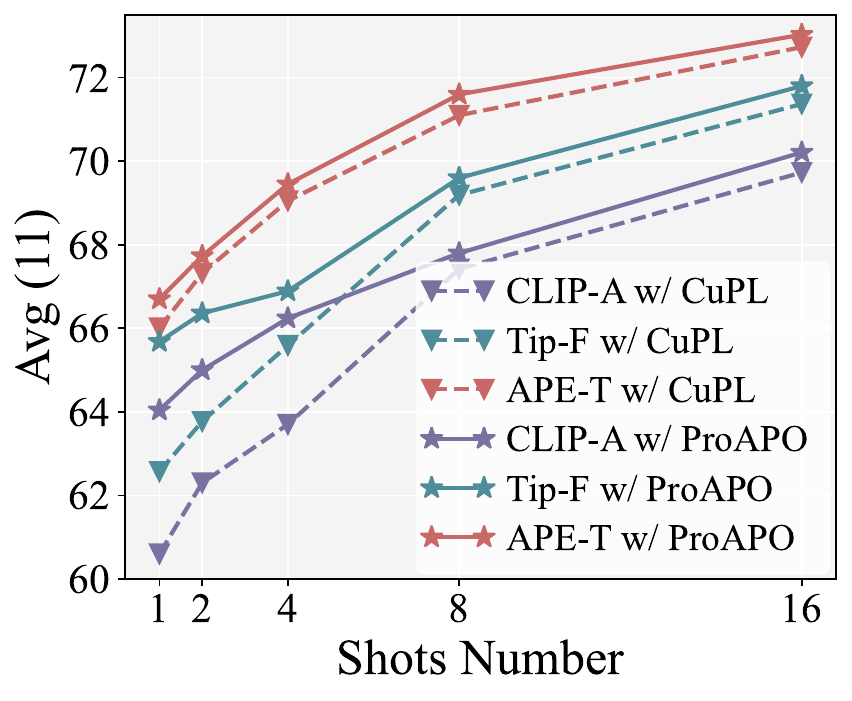}
    \caption{SUN.}
\end{subfigure}
\begin{subfigure}{0.33\textwidth}
    \includegraphics[width=\textwidth]{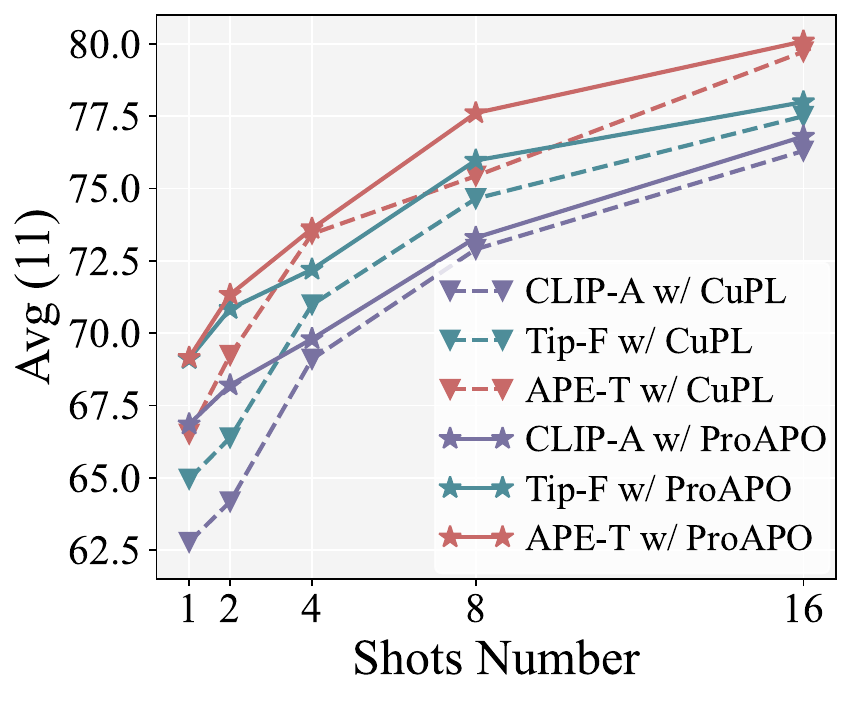}
    \caption{UCF.}
\end{subfigure}
\end{adjustbox}
\caption{\textbf{Results of training adapter-based methods with different initial prompts.} Solid and dotted lines denote prompt initialization with ProAPO and CuPL, respectively. We see that our ProAPO consistently improves adapter-based methods.}
\label{supp_fig: improve_train_adapter}
\vspace{60pt}
\end{figure*}
}

\subsection{Transfer to Different Backbones}
\label{sec_supp: transfer_to_backbones}

In~\cref{supp_fig: transfer_backbones}, we show detailed results of transferring prompts from source to target models in thirteen datasets. Our optimized prompts of ResNet50 and ViT-B/32 are reported.
We see that ProAPO achieves stable performance gains compared to CuPL~\cite{CuPL}, which verifies that ProAPO transfers easily across different backbones.

{
\begin{figure*}[htbp]
\centering
{
    \hfill
    \subfloat[ImageNet]{\includegraphics[width=0.32\textwidth]{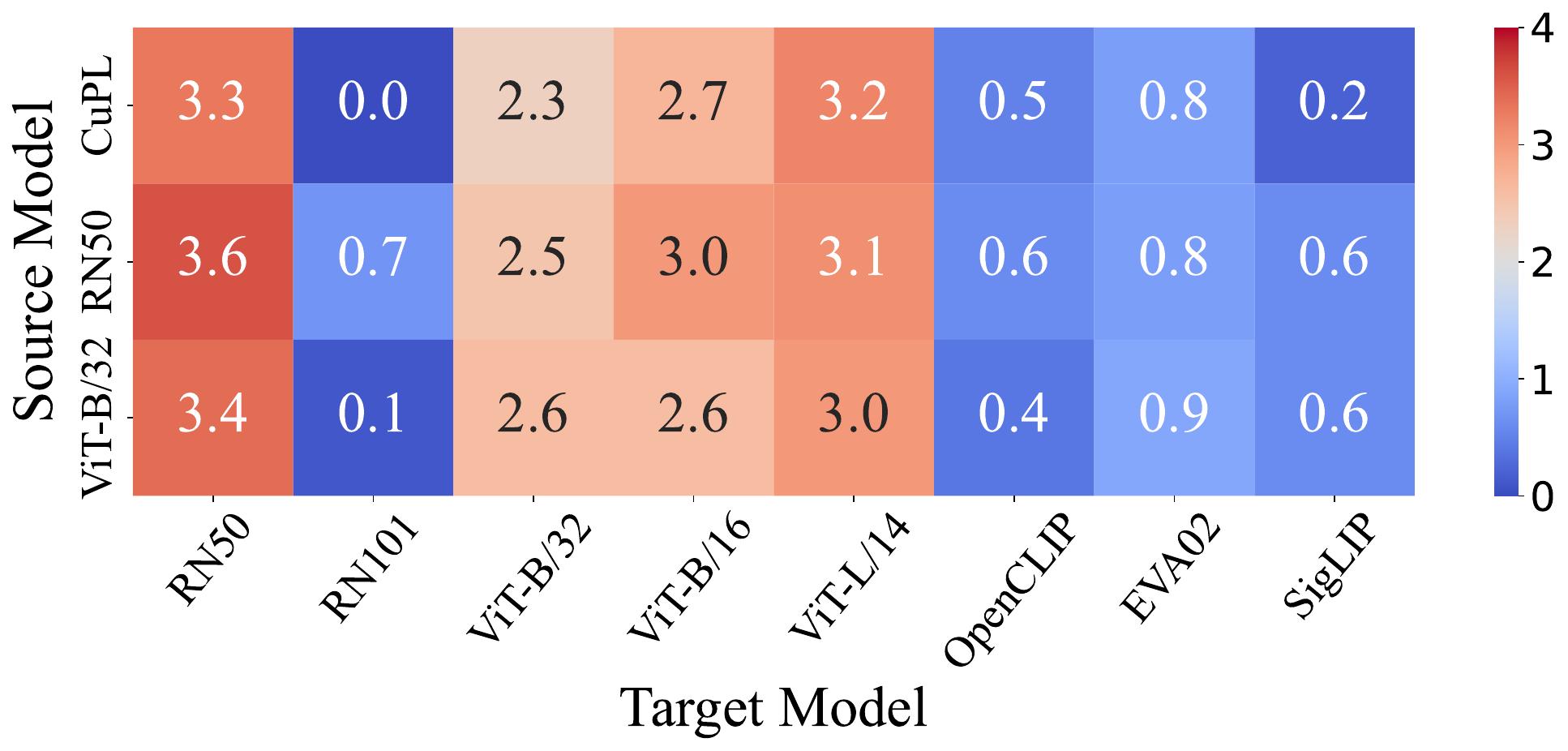}}
    \hfill
    \subfloat[Caltech]{\includegraphics[width=0.32\textwidth]{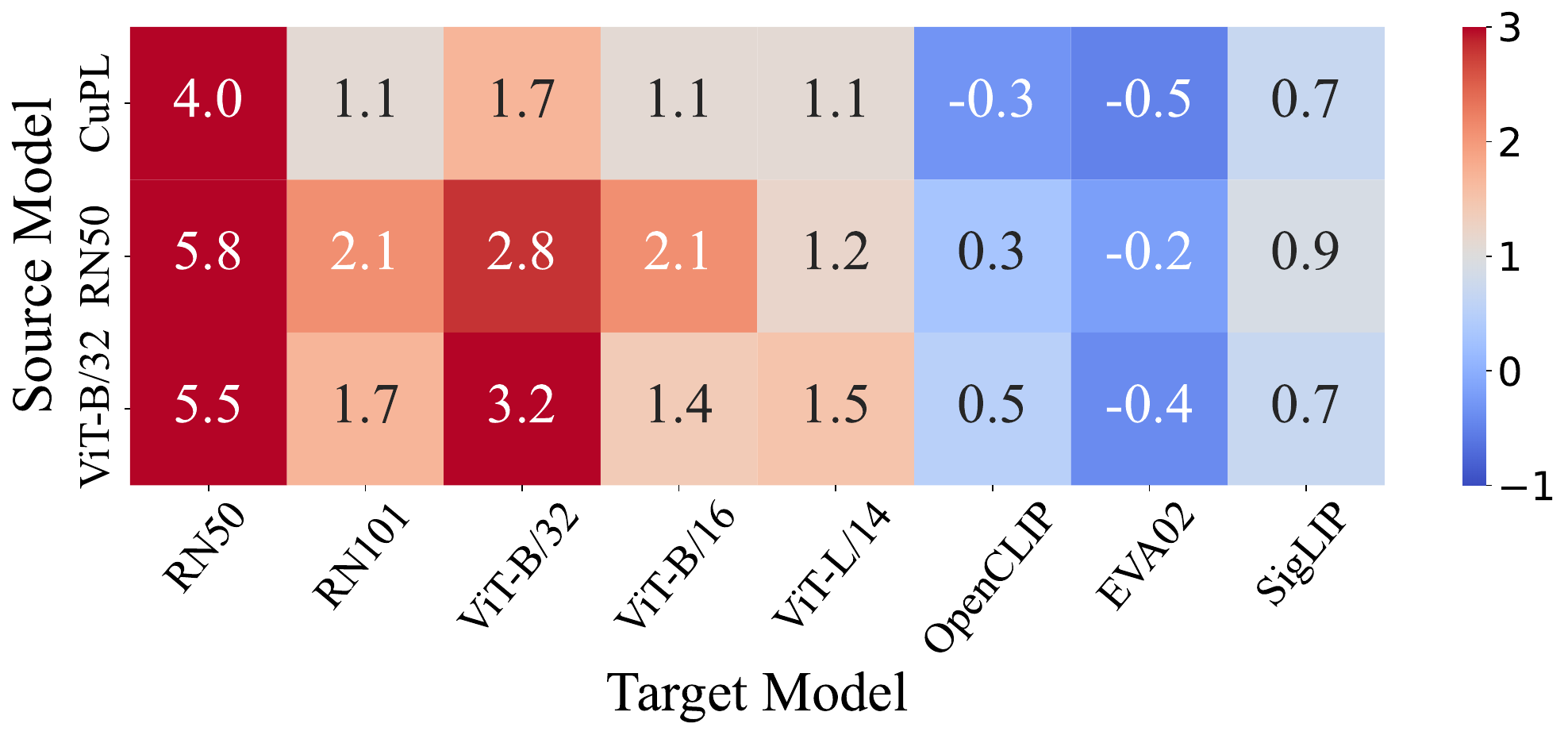}}
    \hfill
    \subfloat[Cars]{\includegraphics[width=0.32\textwidth]{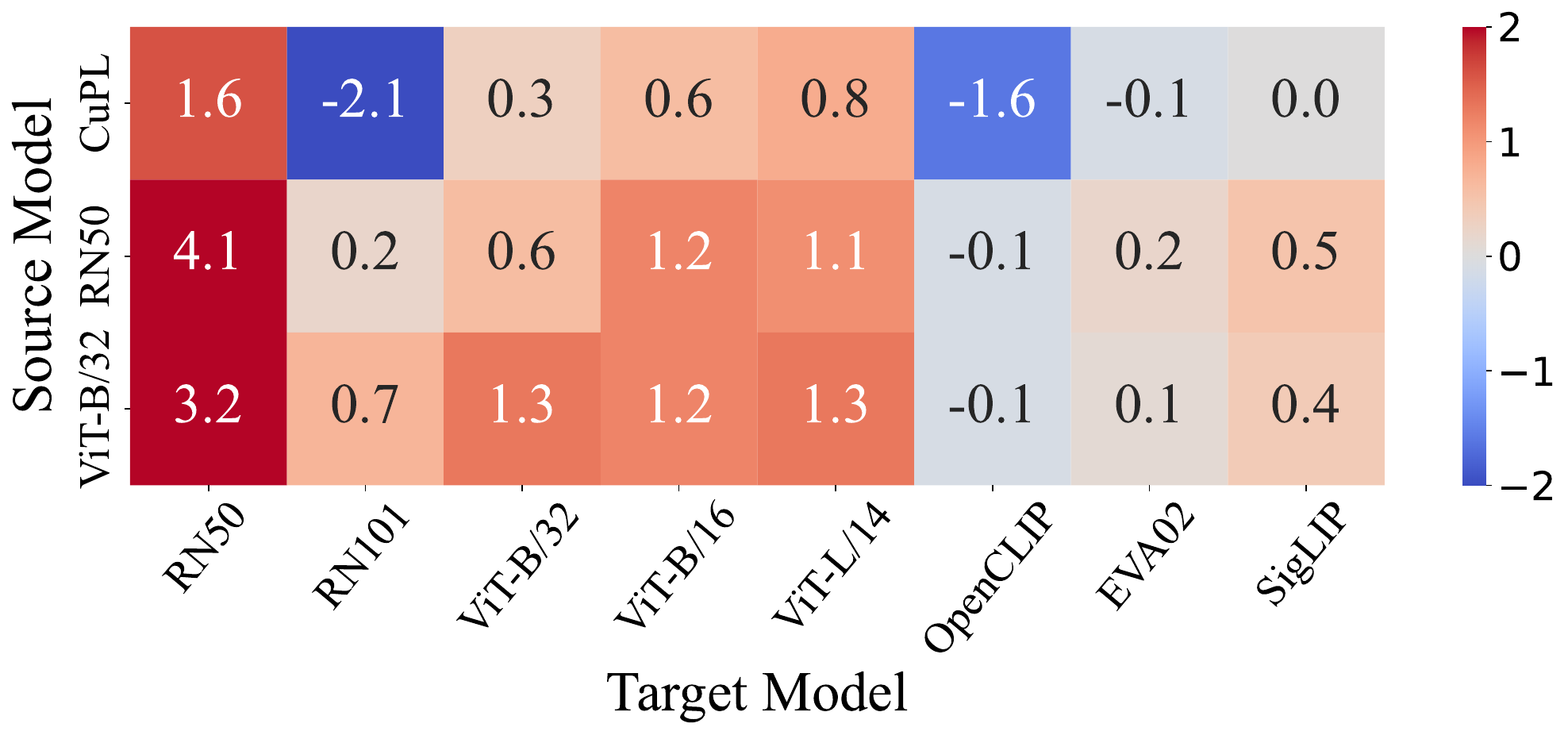}}
    \hfill
    \subfloat[CUB]{\includegraphics[width=0.32\textwidth]{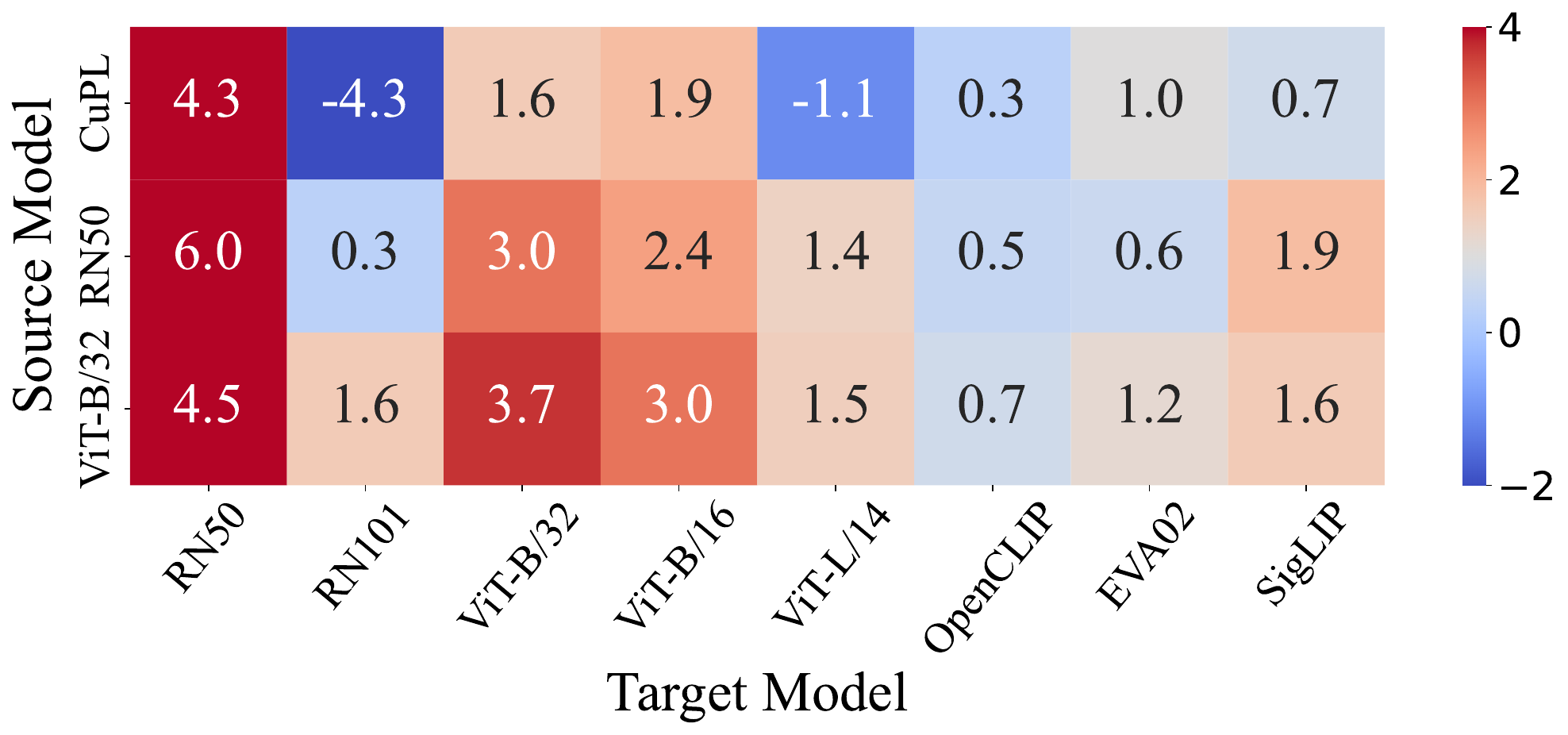}}
    \hfill
    \subfloat[DTD]{\includegraphics[width=0.32\textwidth]{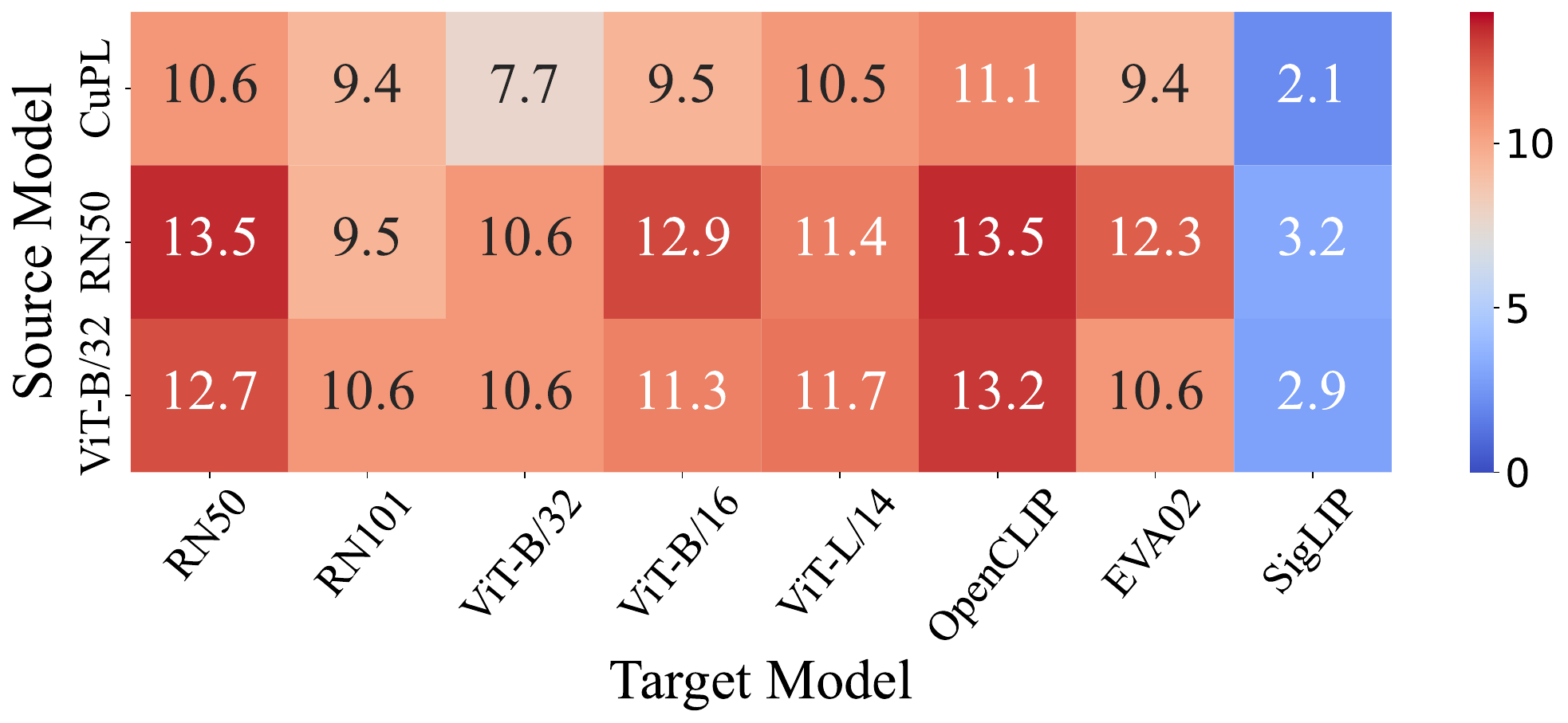}}
    \hfill
    \subfloat[FGVC]{\includegraphics[width=0.32\textwidth]{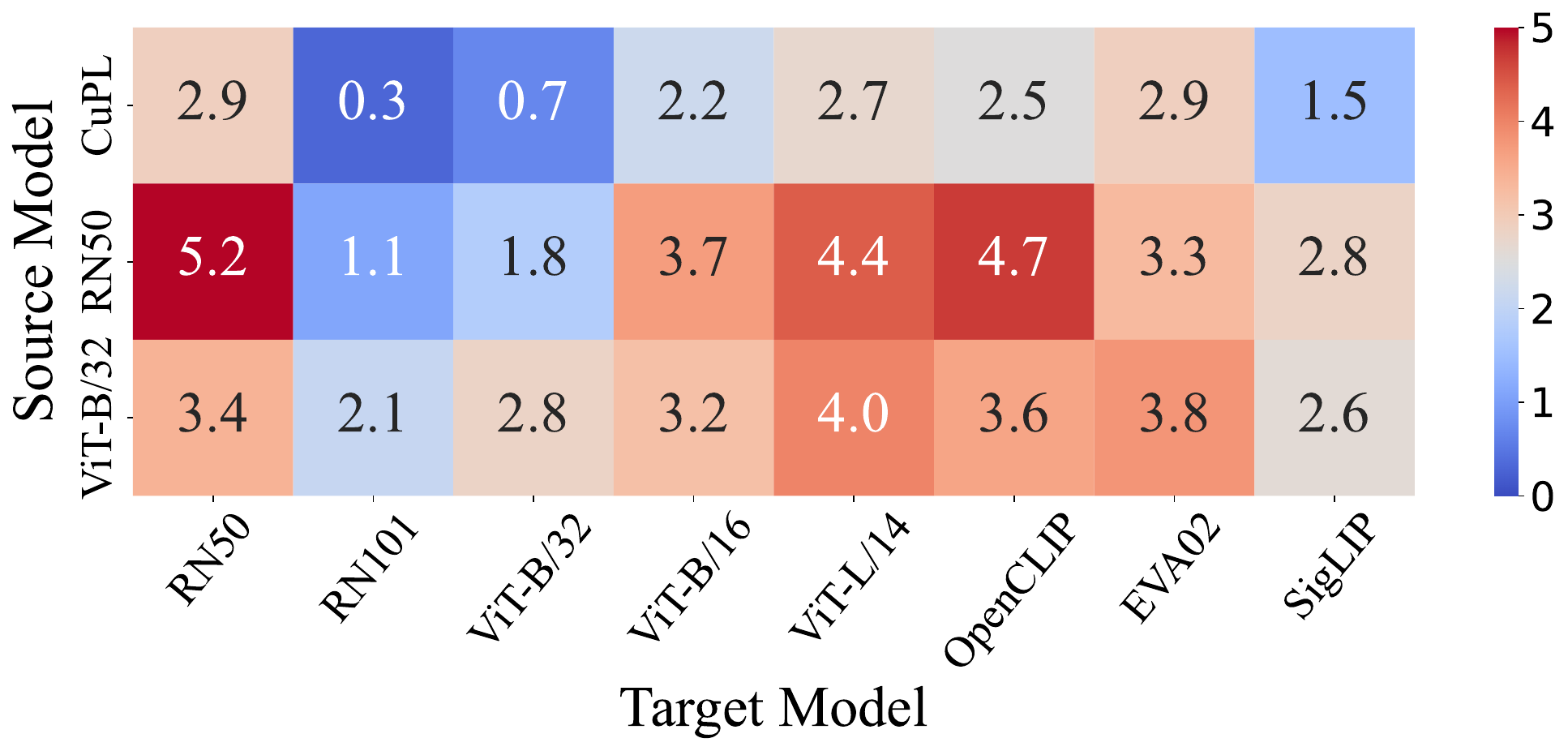}}
    \hfill
    \subfloat[ESAT]{\includegraphics[width=0.32\textwidth]{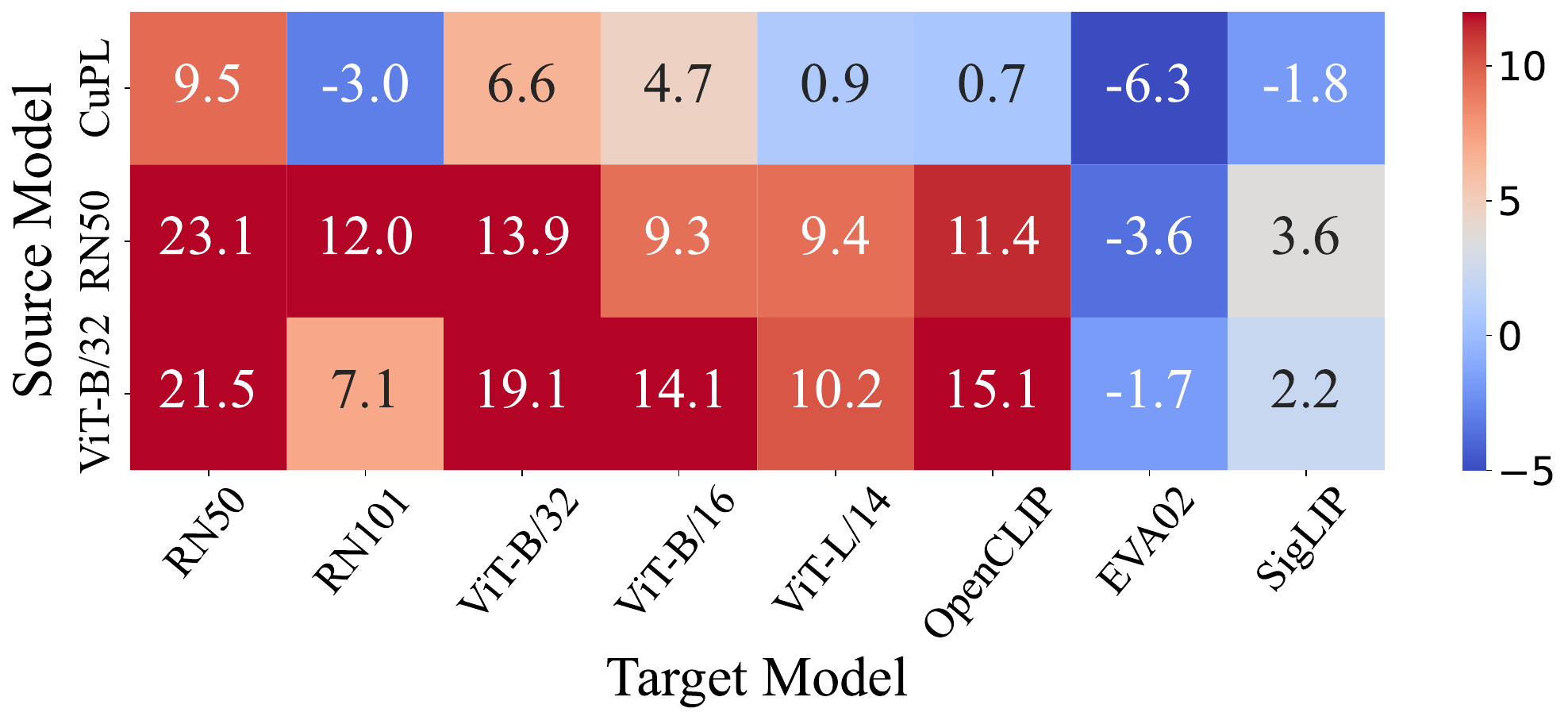}}
    \hfill
    \subfloat[FLO]{\includegraphics[width=0.32\textwidth]{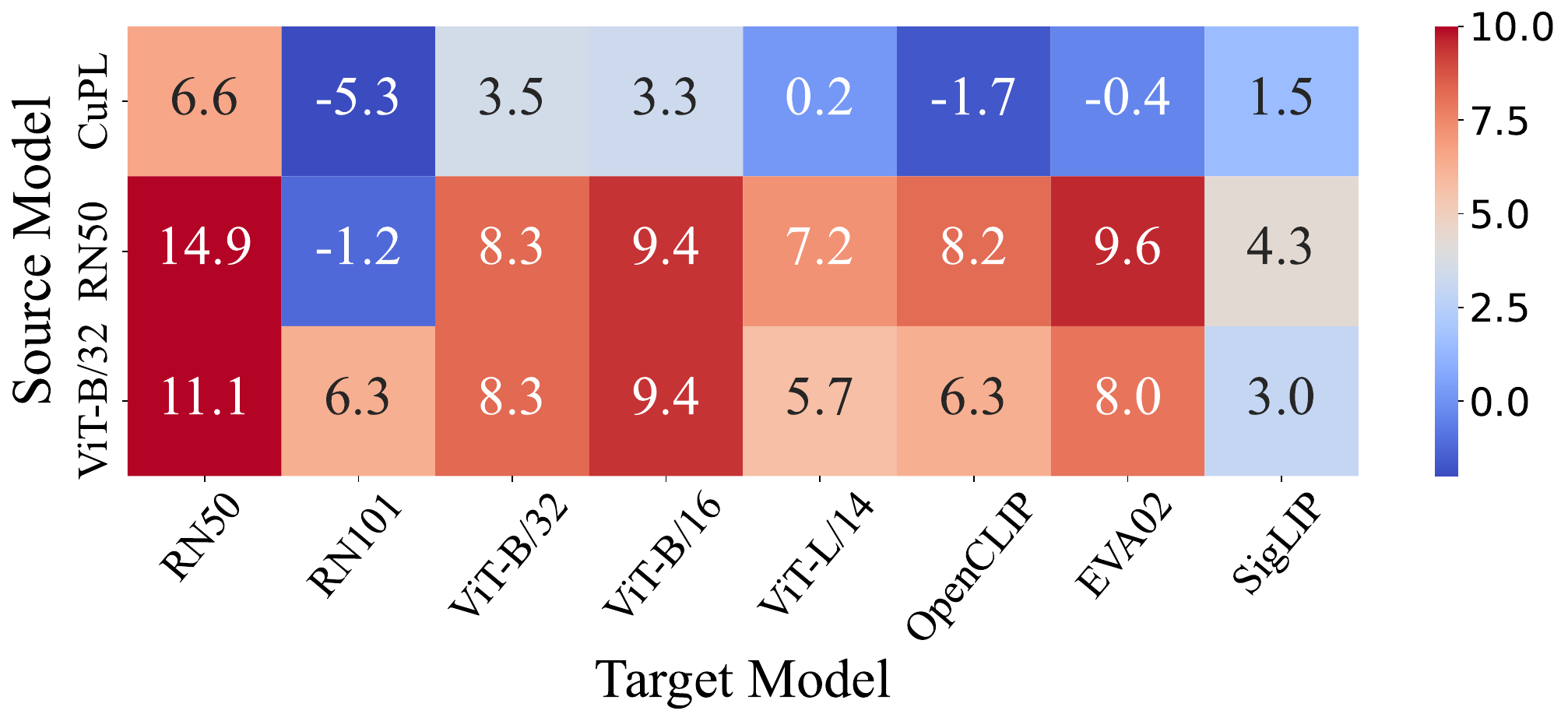}}
    \hfill
    \subfloat[Food]{\includegraphics[width=0.32\textwidth]{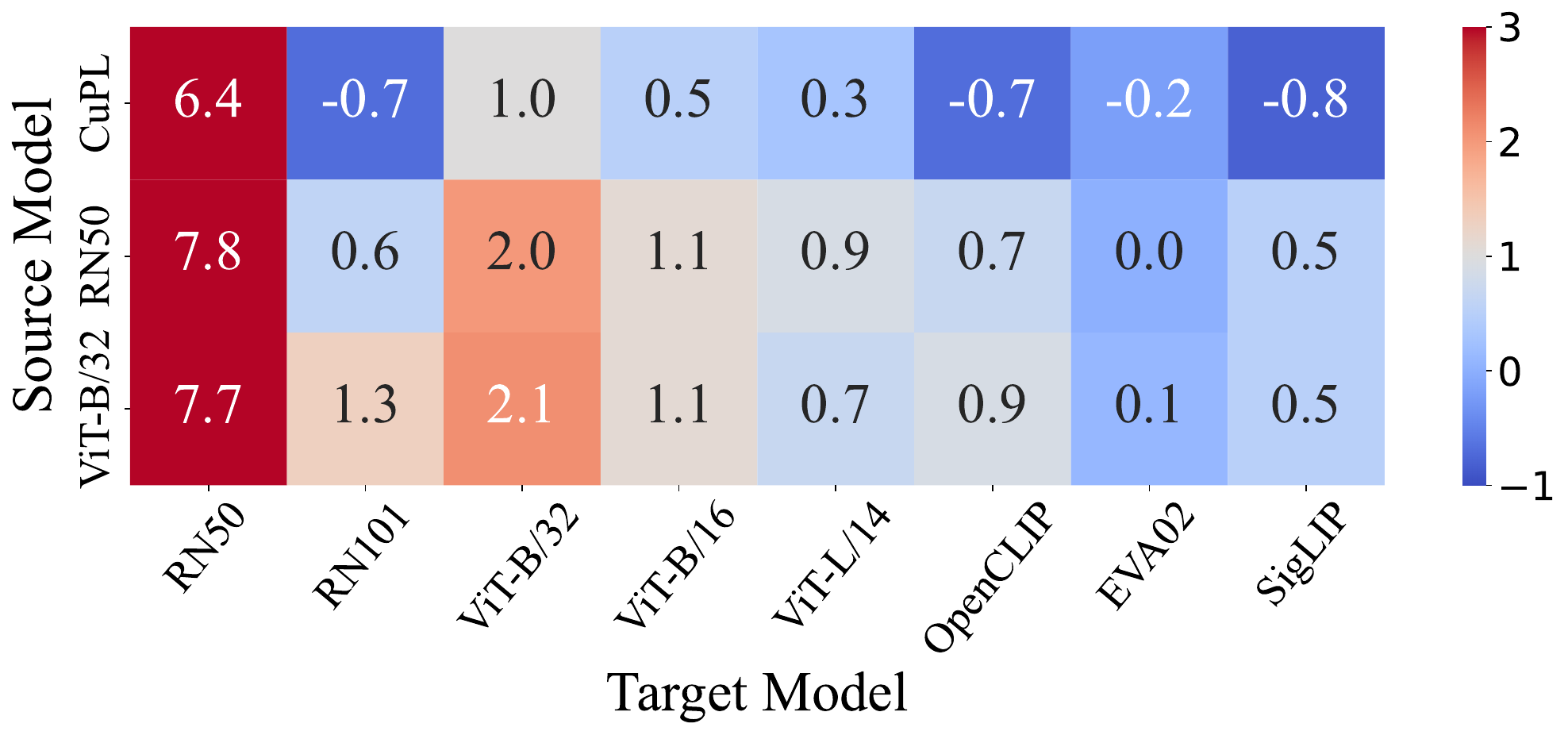}}
    \hfill
    \subfloat[Pets]{\includegraphics[width=0.32\textwidth]{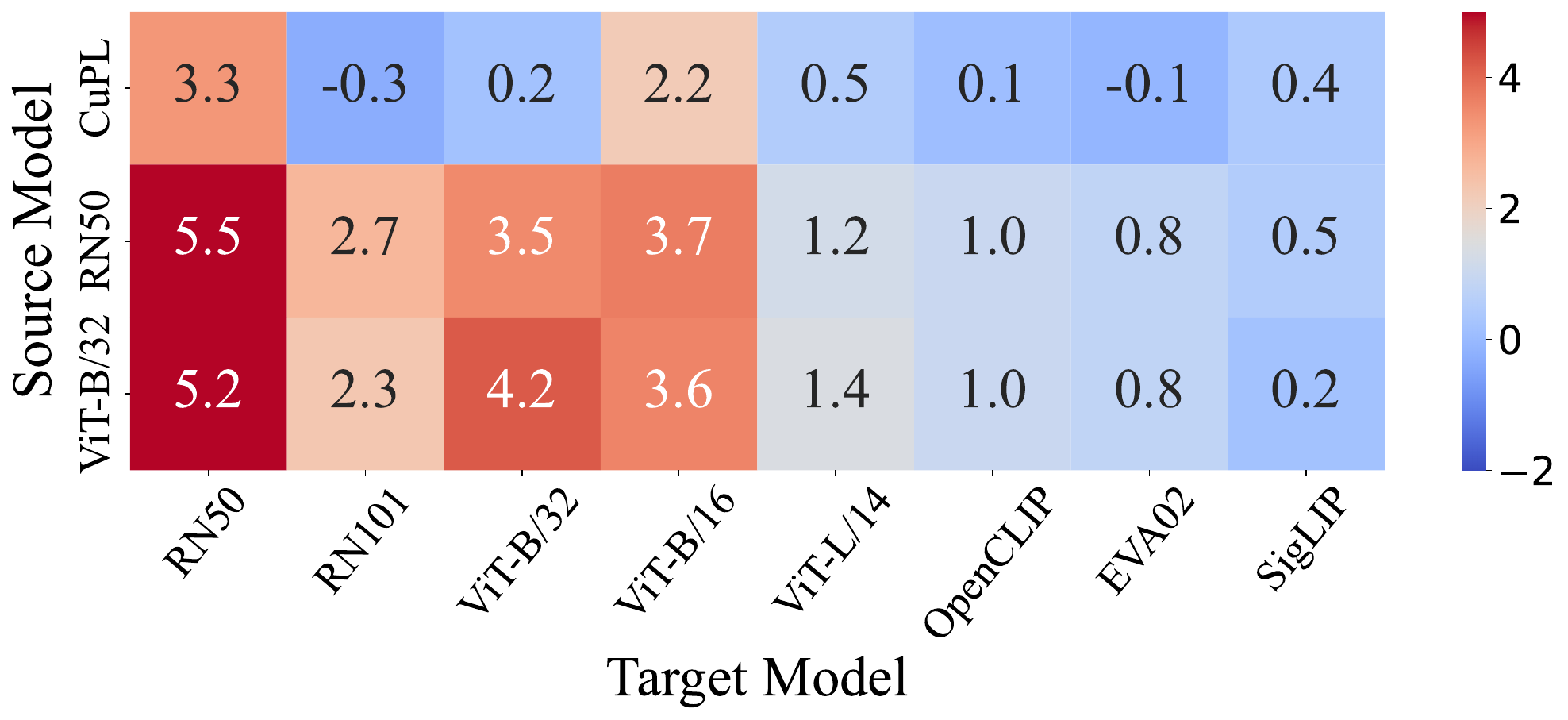}}
    \hfill
    \subfloat[Places]{\includegraphics[width=0.32\textwidth]{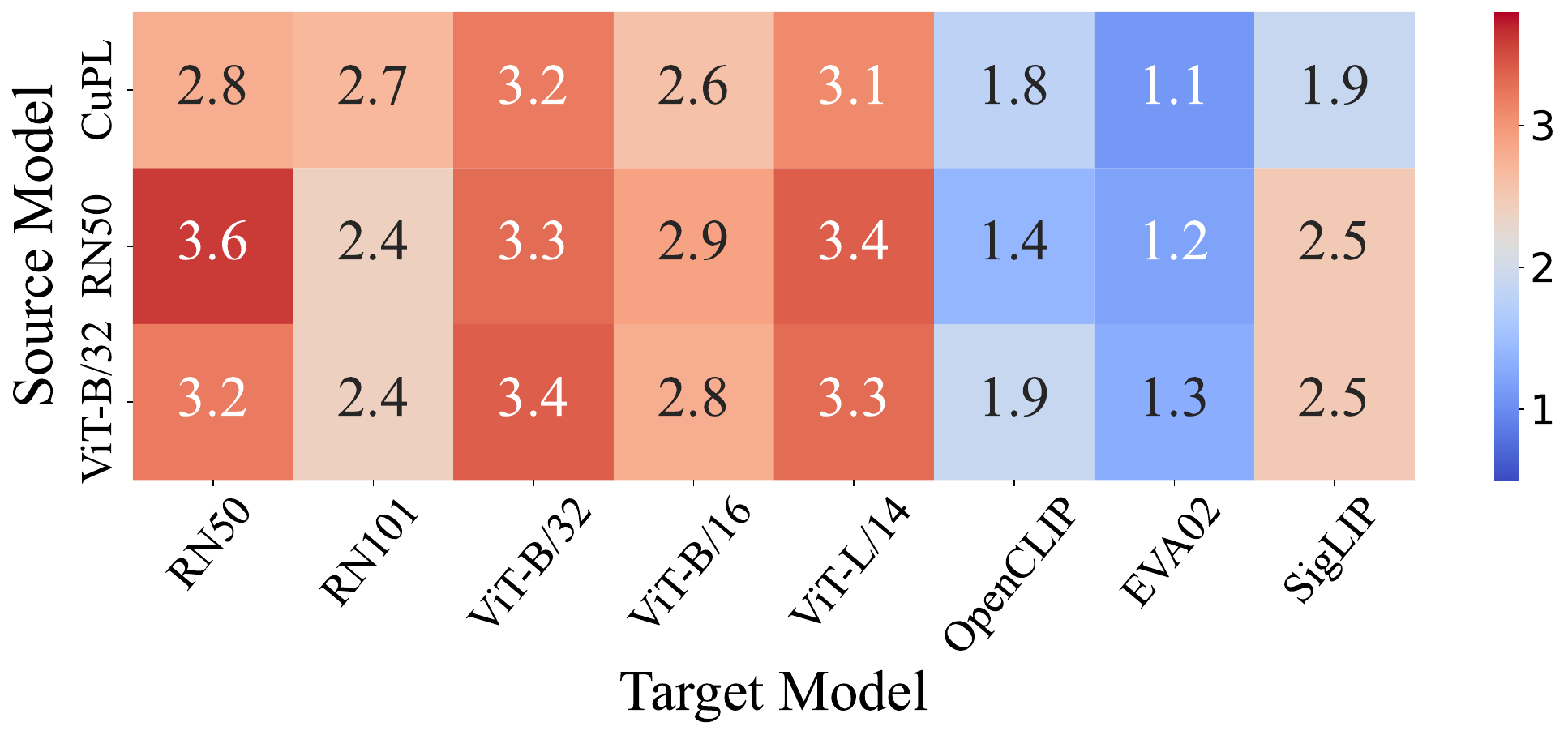}}
    \hfill
    \subfloat[SUN]{\includegraphics[width=0.32\textwidth]{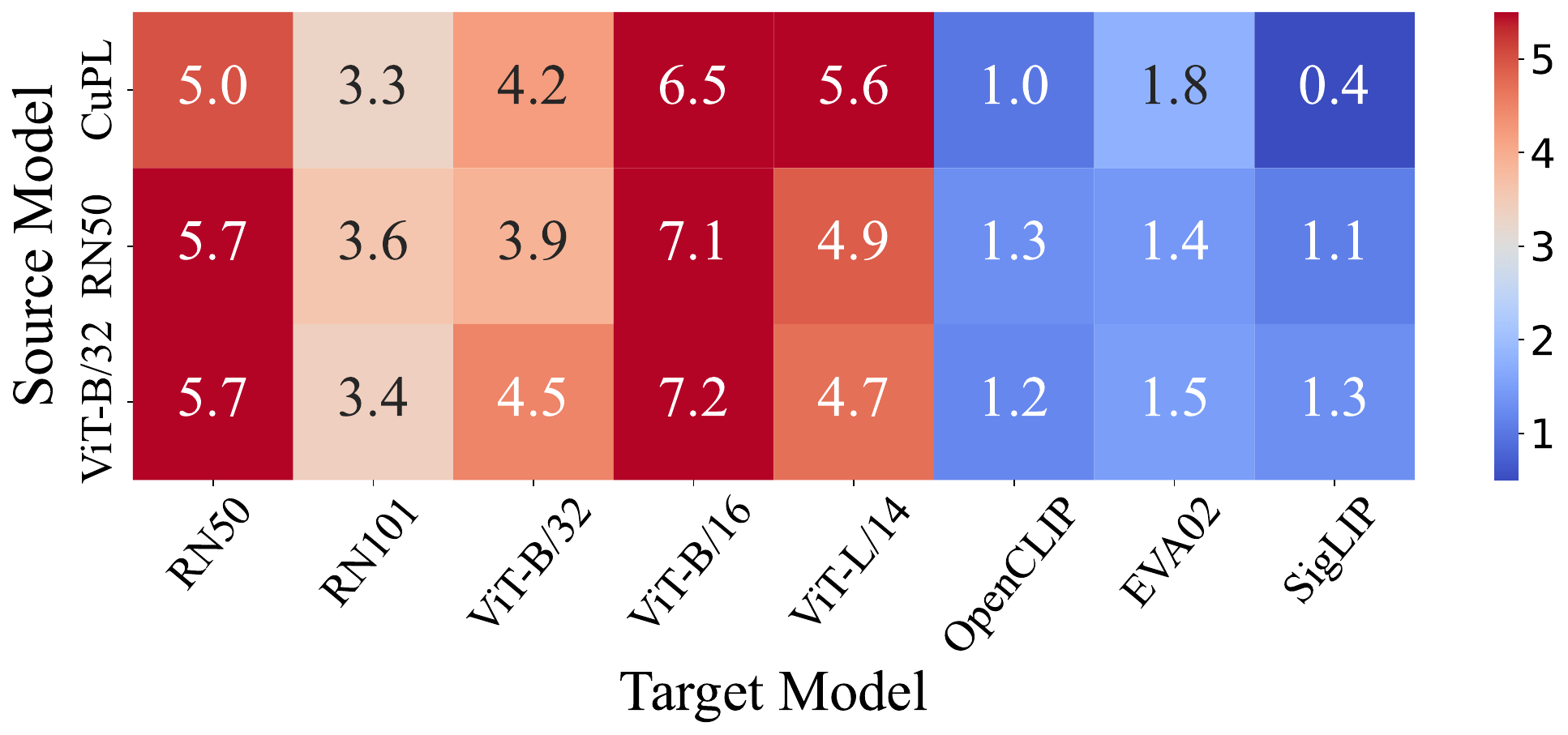}}
    \hfill
    \subfloat[UCF]{\includegraphics[width=0.32\textwidth]{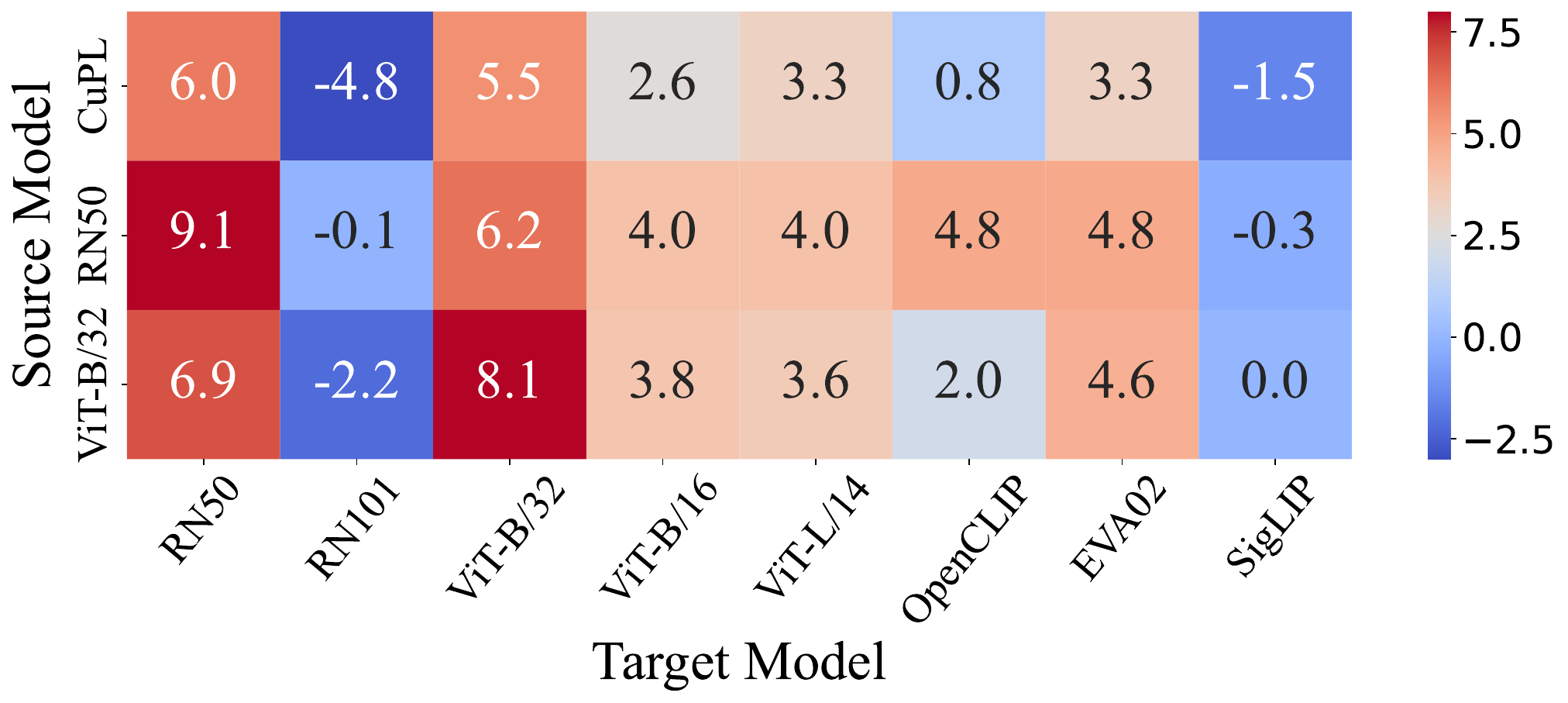}}
    \subfloat[Avg (11)]{\includegraphics[width=0.32\textwidth]{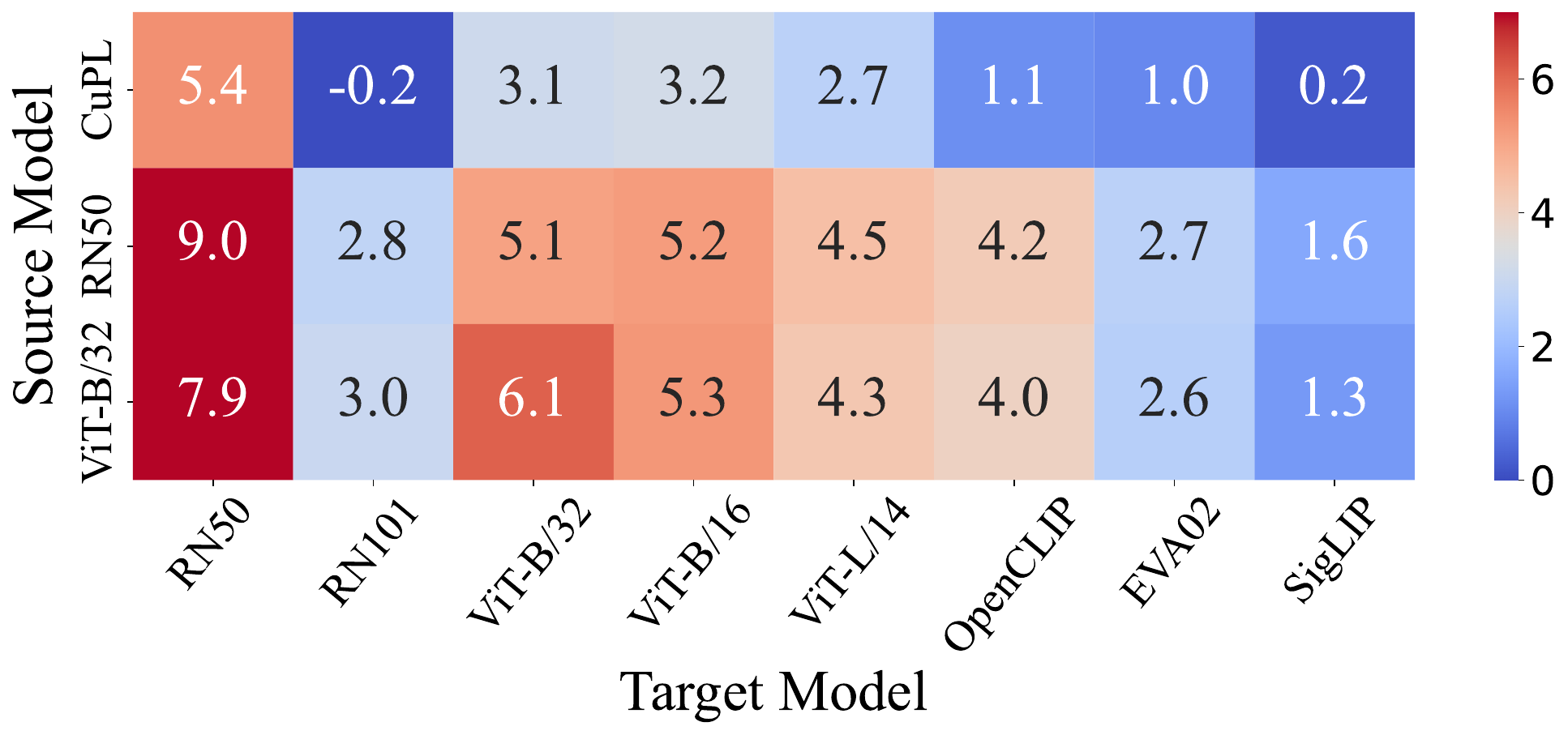}}
}
\vspace{-7pt}
\caption{\textbf{Results of prompt transfer to different backbones.} The value denotes performance gains compared to vanilla VLMs. Our optimized prompts of ResNet50 and ViT-B/32 are reported. We see that we achieve stable performance gains compared to CuPL~\cite{CuPL}. 
}
\label{supp_fig: transfer_backbones}
\vspace{-5pt}
\end{figure*}
}

\section{Detailed Results of Performance Improvement Analysis}


{
\renewcommand{\arraystretch}{1.1} 
\begin{table*}[htbp]
  \centering
  \resizebox{0.75\linewidth}{!}
    {
    \begin{tabular}
        {l |  ccccc ccccc c | c }
        \toprule
        \textbf{Module} (ResNet50) & \rotatebox{90}{\textbf{IN-1K}} & \rotatebox{90}{\textbf{Caltech}} & \rotatebox{90}{\textbf{Cars}}  & \rotatebox{90}{\textbf{DTD}}  & \rotatebox{90}{\textbf{ESAT}} & \rotatebox{90}{\textbf{FGVC}} & \rotatebox{90}{\textbf{FLO}} & \rotatebox{90}{\textbf{Food}}  &  \rotatebox{90}{\textbf{Pets}}  & \rotatebox{90}{\textbf{SUN}} & \rotatebox{90}{\textbf{UCF}} & \rotatebox{90}{\textbf{Avg (11)}}  \\
        \midrule

        CLIP (a photo of a \{\})  & 57.9 & 84.5 & 53.9 & 38.8 & 28.6 & 15.9 & 60.2 & 74.0 & 83.2 & 58.0 & 56.9 & 55.6 \\ 

        \midrule
        \multicolumn{13}{c}{\textit{\textbf{\ccol{Single Prompt}}}} \\
        \midrule
        
        PN~\cite{P_N} & {59.6} & 89.1 & 56.2  & 44.8 & \underline{49.0} & 18.1 & 67.2 & {78.3} & 88.1  & 61.0 & 60.2 & 61.1 \\
        
        Best Single* & 60.2 & \underline{89.2} & \underline{57.9}  & 45.0 & 46.0 & \underline{18.3} & \underline{68.1} & \underline{81.8} & 88.3 & 61.5 & 62.6 & 61.7 \\

        \midrule
        \multicolumn{13}{c}{\textit{\textbf{\ccol{Ensemble Prompt}}}} \\
        \midrule 
        
        \highlight{\textbf{ATO} (ours)} & \highlight{\underline{61.3}} & \highlight{\underline{89.2}} & \highlight{\underline{57.9}} & \highlight{\underline{45.4}} & \highlight{44.7} & \highlight{18.2} & \highlight{\underline{68.1}} & \highlight{\underline{81.8}} & \highlight{\underline{88.5}} & \highlight{\underline{61.8}} & \highlight{\underline{63.9}} & \highlight{\underline{61.9}}  \\
        
        Best Ensemble* & \textbf{61.5} & \textbf{90.0} & \textbf{58.4}  & \textbf{47.0} & \textbf{49.1} & \textbf{18.7} & \textbf{69.9} & \textbf{82.2} & \textbf{89.4} & \textbf{62.1} & \textbf{64.8} & \textbf{63.0}  \\

        
        \bottomrule
    \end{tabular}
}
  \vspace{-6pt}
  \caption{\textbf{Analysis of the effect of single vs ensemble prompts.} * denotes results evaluated in the test set. ATO is our automatic template optimization algorithm. We see that our optimized templates achieve higher results than PN~\cite{P_N}, even better than the best single template.}
  \vspace{-6pt}
  \label{supp_tab: ensemble_vs_single}
\end{table*}
}

\subsection{Analysis of the Effect of Single VS Ensemble Prompts}
\label{supp_sec: ensemble_vs_single}
In~\cref{supp_tab: ensemble_vs_single}, we show detailed results of the effect of single vs ensemble prompts. Compared to PN~\cite{P_N}, we utilize prompt ensembling instead of a single prompt to optimize the template and description. We observe that ensemble templates have a higher upper bound than the single template. Similarly, our optimized templates achieve higher performance than PN~\cite{P_N}, even better than the best single template, further verifying the effectiveness of our method.

\subsection{Performance Improvement of Description Methods by ProAPO}
\label{sec_supp: improve_description_methods}
In~\cref{tab:description_add_APO}, we show detailed results of description methods~\cite{DCLIP, CuPL, GPT4Vis, AdaptCLIP} with our ATO and ProAPO. We see a notable improvement in description methods by at least 2.7\% average in thirteen datasets. It further verifies the effectiveness of our progressive optimization. 

{
\renewcommand{\arraystretch}{1.1} 

\begin{table*}[tbp]
  \centering
  \resizebox{0.99\linewidth}{!}
    {
    \begin{tabular}
        {l | lllll lllll lll | l | l}

        \toprule
        \textbf{Module} (ViT-B/32) & \rotatebox{90}{\textbf{IN-1K}} & \rotatebox{90}{\textbf{Caltech}} & \rotatebox{90}{\textbf{Cars}} & \rotatebox{90}{\textbf{CUB}} & \rotatebox{90}{\textbf{DTD}}  & \rotatebox{90}{\textbf{ESAT}} & \rotatebox{90}{\textbf{FGVC}} & \rotatebox{90}{\textbf{FLO}} & \rotatebox{90}{\textbf{Food}}  &  \rotatebox{90}{\textbf{Pets}} & \rotatebox{90}{\textbf{Places}} & \rotatebox{90}{\textbf{SUN}} & \rotatebox{90}{\textbf{UCF}} & \rotatebox{90}{\textbf{Avg (11)}} & \rotatebox{90}{\textbf{Avg (13)}} \\
        \midrule


        Vanilla CLIP & 62.1  & 91.2  & 60.4  & 51.7 & 42.9  & 43.9  & 20.2  & 66.0  & 83.2  & 86.8 & 39.9 & 62.1  & 60.9 & 61.8 & 59.3 \\ 
        
        \midrule
        DCLIP~\cite{DCLIP} & 63.3  & 92.7  & 59.4  & 52.7  & 44.1  & 38.4  & 19.4  & 66.1  & 83.9  & 88.1  & 41.2  & 65.0  & 65.8  & 62.4  & 60.0 \\
        + \textbf{ATO} & 63.8& 93.0& 60.3& 52.5& 46.5& 54.1& 21.8& 68.9& 84.0& 88.4& 41.5& 65.4& 66.0 & 64.7 & 62.0 \\
        + \textbf{ProAPO} & \textbf{64.1} & \textbf{93.2} & \textbf{60.6} & \textbf{53.6} & \textbf{48.2} & \textbf{59.4} & \textbf{22.6} & \textbf{71.5} & \textbf{84.2} & \textbf{88.7} & \textbf{42.7} & \textbf{66.0} & \textbf{68.0} & \textbf{66.0}  & \textbf{63.3}  \\
        $\Delta$ & \textcolor{retained}{+ 0.8} & \textcolor{retained}{+ 0.5} & \textcolor{retained}{+ 1.2} & \textcolor{retained}{+ 0.9} & \textcolor{retained}{+ 4.1} & \textcolor{retained}{+ 21.0} & \textcolor{retained}{+ 3.2} & \textcolor{retained}{+ 5.4} & \textcolor{retained}{+ 0.3} & \textcolor{retained}{+ 0.6} & \textcolor{retained}{+ 1.5} & \textcolor{retained}{+ 1.0} & \textcolor{retained}{+ 2.2} & \textcolor{retained}{+ 3.6} & \textcolor{retained}{+ 3.3} \\

        \midrule
        
        CuPL-base~\cite{CuPL} & 64.0  & 92.3  & 60.1  & 54.3  & 47.2  & 42.4  & 21.7  & 68.7  & 84.3  & 88.8  & 42.0  & \textbf{66.2}  & 66.7  & 63.8  & 61.4 \\
        + \textbf{ATO} & 64.2 & 93.3 & 60.9 & 54.8 & 47.8 & 53.1 & 22.2  & 70.4  & 84.9 & 89.2  & 42.3 & 65.5 & 67.4 & 65.3 & 62.8 \\
        + \textbf{ProAPO} & \textbf{64.4} & \textbf{94.2} & \textbf{61.8} & \textbf{55.9} & \textbf{48.1} & \textbf{62.1} & \textbf{23.2} & \textbf{74.4} & \textbf{85.4} & \textbf{91.0} & \textbf{42.7} & 65.6 & \textbf{68.6} & \textbf{67.2}  & \textbf{64.4} \\
        $\Delta$ & \textcolor{retained}{+ 0.4} & \textcolor{retained}{+ 1.9} & \textcolor{retained}{+ 1.7} & \textcolor{retained}{+ 1.6} & \textcolor{retained}{+ 0.9} & \textcolor{retained}{+ 19.7} & \textcolor{retained}{+ 1.5} & \textcolor{retained}{+ 5.7} & \textcolor{retained}{+ 1.1} & \textcolor{retained}{+ 2.2} & \textcolor{retained}{+ 0.7} & -0.6 & \textcolor{retained}{+ 1.9} & \textcolor{retained}{+ 3.4} & \textcolor{retained}{+ 3.0} \\

        \midrule
        CuPL-full~\cite{CuPL} & 64.4  & 92.9  & 60.7  & 53.3  & 50.6  & 50.5  & 20.9  & 69.5  & 84.2  & 87.0  & 43.1  & 66.3  & 66.4  & 64.9  & 62.3 \\
        + \textbf{ATO} & 64.5 & 93.7 & 61.0 & 54.0 & 52.0 & 58.7 & 22.1 & 70.5 & 84.6 & 89.2 & 43.2 & 66.4 & 67.5 & 66.4 &  63.6 \\
        + \textbf{ProAPO} &  \textbf{64.7} & \textbf{94.4} & \textbf{61.7} & \textbf{55.4} & \textbf{53.5} & \textbf{63.0} & \textbf{23.0} & \textbf{74.3} & \textbf{85.3} & \textbf{91.0} & \textbf{43.3} & \textbf{66.6} & \textbf{69.0} & \textbf{67.9}  & \textbf{65.0} \\
        $\Delta$ & \textcolor{retained}{+ 0.3} & \textcolor{retained}{+ 1.5} & \textcolor{retained}{+ 1.0} & \textcolor{retained}{+ 2.1} & \textcolor{retained}{+ 2.9} & \textcolor{retained}{+ 12.5} & \textcolor{retained}{+ 2.1} & \textcolor{retained}{+ 4.8} & \textcolor{retained}{+ 1.1} & \textcolor{retained}{+ 4.0} & \textcolor{retained}{+ 0.2} & \textcolor{retained}{+ 0.3} & \textcolor{retained}{+ 2.6} & \textcolor{retained}{+ 3.0} & \textcolor{retained}{+ 2.7} \\

        \midrule
        GPT4Vis~\cite{GPT4Vis} & 63.5  & 93.1  & 61.4  & 52.7  & 48.5  & 47.0  & 21.4  & 69.8  & 84.3  & 88.1  & 42.7  & 64.2  & 65.7  & 64.3  & 61.7 \\ 
        + \textbf{ATO} & 63.8 & 93.4 & 61.2 & 53.8 & 49.0 & 54.0  & 22.4 & 70.8  & 84.7  & 88.1  & 42.6 & 64.7 & 66.8 & 65.3 & 62.7 \\
        + \textbf{ProAPO} & \textbf{64.4} & \textbf{93.7} & \textbf{61.8} & \textbf{55.4} & \textbf{49.3} & \textbf{62.6} & \textbf{23.9} & \textbf{73.8} & \textbf{85.4} & \textbf{90.7} & \textbf{42.8} & \textbf{65.5} & \textbf{68.2} & \textbf{67.2}  & \textbf{64.4} \\
        $\Delta$ & \textcolor{retained}{+ 0.9} & \textcolor{retained}{+ 0.6} & \textcolor{retained}{+ 0.4} & \textcolor{retained}{+ 2.7} & \textcolor{retained}{+ 0.8} & \textcolor{retained}{+ 15.6} & \textcolor{retained}{+ 2.5} & \textcolor{retained}{+ 4.0} & \textcolor{retained}{+ 1.1} & \textcolor{retained}{+ 2.6} & \textcolor{retained}{+ 0.1} & \textcolor{retained}{+ 1.3} & \textcolor{retained}{+ 2.5} & \textcolor{retained}{+ 2.9} & \textcolor{retained}{+ 2.7} \\

        \midrule 
        AdaptCLIP~\cite{AdaptCLIP} & 63.3  & 92.7  & 59.7  & 53.6  & 47.4  & 51.3  & 20.8  & 67.2  & 84.2  & 87.6  & 41.9  & 66.1  & 66.5  & 64.2  & 61.7 \\
        + \textbf{ATO} & 63.9 & 93.2 & 60.4 & 54.2 & 47.9 & 55.5 & \textbf{22.4} & 69.1 & 84.7  & 88.8  & 42.3 & 66.3 & 67.6 & 65.4 & 62.8 \\
        + \textbf{ProAPO} & \textbf{64.4} & \textbf{93.7} & \textbf{61.8} & \textbf{55.5} & \textbf{49.6} & \textbf{61.6} & {23.3} & \textbf{73.8} & \textbf{85.4} & \textbf{91.0} & \textbf{42.6} & \textbf{66.5} & \textbf{68.6} & \textbf{67.2}  & \textbf{64.5} \\
        $\Delta$ & \textcolor{retained}{+ 1.1} & \textcolor{retained}{+ 1.0} & \textcolor{retained}{+ 2.1} & \textcolor{retained}{+ 1.9} & \textcolor{retained}{+ 2.2} & \textcolor{retained}{+ 10.3} & \textcolor{retained}{+ 2.5} & \textcolor{retained}{+ 6.6} & \textcolor{retained}{+ 1.2} & \textcolor{retained}{+ 3.4} & \textcolor{retained}{+ 0.7} & \textcolor{retained}{+ 0.4} & \textcolor{retained}{+ 2.1} & \textcolor{retained}{+ 3.0} & \textcolor{retained}{+ 2.8}  \\
        
        \bottomrule
    \end{tabular}
}
  \caption{\textbf{Performance improvement of description methods by our ProAPO.} 
  Avg (11) and Avg (13) denote average results across 11 datasets (excluding CUB~\cite{CUB} and Places~\cite{Places365}) and all 13 datasets, respectively. $\Delta$ denotes performance gains compared to baseline.}
  \vspace{40pt}
  \label{tab:description_add_APO}
\end{table*}
}

{
\renewcommand{\arraystretch}{1.1} 
\begin{table*}[t]
  \centering
  \resizebox{0.99\linewidth}{!}
    {
    \begin{tabular}
        {l c c c c c | ccccc ccccc ccc | c | c }    
        \toprule
        \multicolumn{6}{c}{\textbf{Component}}  &   \\
        \cmidrule(lr){1-6} 
        & \texttt{Add} & \texttt{Del} & \texttt{Rep} & \texttt{Cross} & \texttt{Mut}  & \rotatebox{90}{\textbf{IN-1K}} & \rotatebox{90}{\textbf{Caltech}} & \rotatebox{90}{\textbf{Cars}} & \rotatebox{90}{\textbf{CUB}} & \rotatebox{90}{\textbf{DTD}}  & \rotatebox{90}{\textbf{ESAT}} & \rotatebox{90}{\textbf{FGVC}} & \rotatebox{90}{\textbf{FLO}} & \rotatebox{90}{\textbf{Food}}  &  \rotatebox{90}{\textbf{Pets}} & \rotatebox{90}{\textbf{Places}} & \rotatebox{90}{\textbf{SUN}} & \rotatebox{90}{\textbf{UCF}} & \rotatebox{90}{\textbf{Avg (11)}} & \rotatebox{90}{\textbf{Avg (13)}}  \\
        \midrule
       \multicolumn{5}{l}{Vanilla CLIP (ViT-B/32)} & & 62.1  & 91.2  & 60.4  & 51.7 & 42.9  & 43.9  & 20.2  & 66.0  & 83.2  & 86.8 & 39.9 & 62.1  & 60.9 & 61.8 & 59.3  \\ 
       \midrule
       \multicolumn{5}{l}{\textbf{\textit{edit-based generation}}} \\
       \texttt{a)} & \cmark & & & & & 63.8 & 93.6 & 60.0 & 54.6 & 51.8 & 59.0 & 21.8 & 74.0 & 82.2 & 86.7 & 43.0 & 65.7 & 66.8 & 66.0  & 63.3  \\ 
       \texttt{b)} & \cmark & \cmark & & & & \underline{64.6} & 94.0 & 60.9 & 55.0 & 52.6 & 59.3 & 21.8 & 72.0 & 83.2 & \underline{88.0} & 43.2 & 66.4 & 68.0 & 66.4  & 63.8 \\
       \texttt{c)} & \cmark &  & \cmark & & & 64.4 & 94.0 & 61.0 & \underline{55.2} & 52.3 & 59.7 & 22.4 & 71.9 & 84.0 & 87.7 & 43.2 & 66.4 & 67.8 & 66.5  & 63.8 \\
       \texttt{d)} & \cmark & \cmark & \cmark & & & \underline{64.6}  & 93.6 & 60.8 & 54.4 & 53.1 & 60.1 & 22.2 & \textbf{74.7} & 82.4 & 87.2 & \textbf{43.4} & 66.5 & \underline{68.6} & 66.7  & 64.0  \\ 
        \midrule
        \multicolumn{5}{l}{\textbf{\textit{evolution-based generation}}} \\
        \texttt{e)} & \cmark & \cmark & \cmark & \cmark & & \underline{64.6} & \underline{94.3} & 61.2 & 55.0 & \underline{53.2} & \underline{62.6} & \underline{22.9} & 73.9 & \underline{84.3} & \underline{88.0} & 43.1 & \textbf{66.8} & 68.5 & \underline{67.3}  & \underline{64.5} \\ 
        \texttt{f)} & \cmark & \cmark & \cmark & & \cmark &  \textbf{64.7} & \underline{94.3} & \underline{61.4} & 55.1 & 52.9 & 61.4 & 22.6 & 74.0 & 83.6 & 87.7 & \textbf{43.4} & \underline{66.7} & 68.3 & 67.1  & 64.3   \\ 
        \highlight{\texttt{g)}} & \highlight{\cmark} & \highlight{\cmark} & \highlight{\cmark} & \highlight{\cmark} & \highlight{\cmark} & \highlight{\textbf{64.7}}  & \highlight{\textbf{94.4}} & \highlight{\textbf{61.7}} & \highlight{\textbf{55.4}} & \highlight{\textbf{53.5}} & \highlight{\textbf{63.0}} & \highlight{\textbf{23.0}} & \highlight{\underline{74.3}} & \highlight{\textbf{85.3}} & \highlight{\textbf{91.0}} & \highlight{\underline{43.3}} & \highlight{{66.6}} & \highlight{\textbf{69.0}} & \highlight{\textbf{67.9}}  & \highlight{\textbf{65.0}}   \\
        \bottomrule
    \end{tabular}
}
\caption{\textbf{Ablation of edit- and evolution-based operators.}}
\vspace{50pt}
\label{supp_tab: ablation_generate}
\end{table*}
}

{
\renewcommand{\arraystretch}{1.1} 
\begin{table*}[tbp]
  \centering
  \resizebox{0.99\linewidth}{!}
    {
    \begin{tabular}
        {l | ccccc ccccc ccc | c | c | c}  
        \toprule
        {\textbf{Module} (ViT-B/32)}  & \rotatebox{90}{\textbf{IN-1K}} & \rotatebox{90}{\textbf{Caltech}} & \rotatebox{90}{\textbf{Cars}} & \rotatebox{90}{\textbf{CUB}} & \rotatebox{90}{\textbf{DTD}}  & \rotatebox{90}{\textbf{ESAT}} & \rotatebox{90}{\textbf{FGVC}} & \rotatebox{90}{\textbf{FLO}} & \rotatebox{90}{\textbf{Food}}  &  \rotatebox{90}{\textbf{Pets}} & \rotatebox{90}{\textbf{Places}} & \rotatebox{90}{\textbf{SUN}} & \rotatebox{90}{\textbf{UCF}} & \rotatebox{90}{\textbf{Avg (11)}} & \rotatebox{90}{\textbf{Avg (13)}} & \textbf{Times} \\
        \midrule
         
        \texttt{a)} w/o prompt sampling & 64.4 & 93.8 & \textbf{61.8} & \underline{55.4} & 51.8 & 60.0 & \underline{23.2} & 74.0 & 85.1 & 90.7 & \underline{43.0} & 66.0 & 69.3 & 67.3 & 64.5 & 12 min \\
        
        \texttt{b)} w/o group sampling & \textbf{64.8} & \textbf{94.5} & \underline{61.7} & \textbf{55.5} & \textbf{53.6} & \textbf{63.5} & \underline{23.2} & \underline{75.3} & \textbf{85.4} & 90.8 & \textbf{43.3} & \textbf{66.7} & \textbf{69.8} & \textbf{68.1}  & \textbf{65.2} & \textbf{306 min} \\ 
        
        \texttt{c)} w/o sampling strategies & 64.5 & 93.4 & 57.4 & 54.8 & \textbf{53.6} & \underline{63.2} & \textbf{23.4} & \textbf{76.8} & 83.8 & 86.9 & \textbf{43.3} & 66.1 & \underline{69.7} & 67.2  & 64.4 & \underline{302 min} \\

        \midrule
        
        \highlight{\textbf{ProAPO} (full model)} & \highlight{\underline{64.7}}  & \highlight{\underline{94.4}} & \highlight{\underline{61.7}} & \highlight{\underline{55.4}} & \highlight{\underline{53.5}} & \highlight{{63.0}} & \highlight{{23.0}} & \highlight{{74.3}} & \highlight{\underline{85.3}} & \highlight{\textbf{91.0}} & \highlight{\textbf{43.3}} & \highlight{\underline{66.6}} & \highlight{{69.0}} & \highlight{\underline{67.9}}  & \highlight{\underline{65.0}} & \highlight{15 min} \\
        \bottomrule
    \end{tabular}
}
  \caption{\textbf{Ablation of two sampling strategies.}}
  \label{supp_tab: ablation_sample}
\end{table*}
}

{
\renewcommand{\arraystretch}{1.1} 
\begin{table*}[tbp]
  \centering
  \resizebox{0.99\linewidth}{!}
    {
    \begin{tabular}
        {l | ccccc ccccc ccc | c | c }  
        \toprule
        {\textbf{Module} (ViT-B/32)}  & \rotatebox{90}{\textbf{IN-1K}} & \rotatebox{90}{\textbf{Caltech}} & \rotatebox{90}{\textbf{Cars}} & \rotatebox{90}{\textbf{CUB}} & \rotatebox{90}{\textbf{DTD}}  & \rotatebox{90}{\textbf{ESAT}} & \rotatebox{90}{\textbf{FGVC}} & \rotatebox{90}{\textbf{FLO}} & \rotatebox{90}{\textbf{Food}}  &  \rotatebox{90}{\textbf{Pets}} & \rotatebox{90}{\textbf{Places}} & \rotatebox{90}{\textbf{SUN}} & \rotatebox{90}{\textbf{UCF}} & \rotatebox{90}{\textbf{Avg (11)}} & \rotatebox{90}{\textbf{Avg (13)}} \\
        \midrule
         
        \texttt{a)} w/ only accuracy & 64.0 & 93.0 & 60.8 & 54.2 & 49.1 & 55.5 & 20.4 & 68.3 & 84.8 & 88.4 & 41.9 & 64.6 & 65.1 & 64.9 & 62.3 \\
        
        \texttt{b)} w/ only entropy constrain & 64.3 & 93.4 & 61.6 & 54.8 & 49.3 & 56.7 & 22.3 & 69.9 & 85.2 & 89.1 & 42.4 & 65.1 & 66.7 & 65.8 & 63.1 \\ 
        
        \midrule
        
        \highlight{\textbf{ProAPO} (full model)} & \highlight{\textbf{64.7}}  & \highlight{\textbf{94.4}} & \highlight{\textbf{61.7}} & \highlight{\textbf{55.4}} & \highlight{\textbf{53.5}} & \highlight{\textbf{63.0}} & \highlight{\textbf{23.0}} & \highlight{\textbf{74.3}} & \highlight{\textbf{85.3}} & \highlight{\textbf{91.0}} & \highlight{\textbf{43.3}} & \highlight{\textbf{66.6}} & \highlight{\textbf{69.0}} & \highlight{\textbf{67.9}}  & \highlight{\textbf{65.0}}  \\
        \bottomrule
    \end{tabular}
}
  \caption{\textbf{Ablation of different score functions.}}
  \label{supp_tab: ablation_score_function}
\end{table*}
}

\section{Detailed Results of Ablation Study}

\subsection{Ablation of Edit- and Evolution-based Operators}
\label{supp_sec: ablation_operator}
In~\cref{supp_tab: ablation_generate}, we show detailed ablation results of edit- and evolution-based operators. For edit-based operators, we observe that the model with add, delete, and replace operations achieves a higher result in row d). After introducing evolution-based operators, \textit{i.e.}, crossover operator to combine advantages of high-scoring candidates, and mutation operator to avoid locally optimal solutions, we see an increase in performance in rows e)-g). It confirms that evolution-based operators make the model search the optimal prompt faster with limited iterations.

\subsection{Ablation of Two Sampling Strategies}
\label{supp_sec: ablation_two_sampling}

In~\cref{supp_tab: ablation_sample}, we show detailed ablation results of two sampling strategies. Without the prompt sampling, we see a slight decrease in times while results drop in row a). It verifies the effectiveness of the prompt sampling. Without the group sampling to select salient categories for optimization, we observe a notable increase in time costs (from 15 min to 300+ min, 20 times) yet similar results in row b) and the full model. It reveals that group sampling simultaneously improves performance and efficiency.

\subsection{Ablation of Different Score Functions}
\label{supp_sec: effect_score_func}

In~\cref{supp_tab: ablation_score_function}, we show detailed ablation results of score functions. Accuracy obtains the worst result as the score function due to the overfitting problem. Our full model with accuracy and entropy constraints as the score function achieves the SOTA result. The score function with only accuracy or entropy constraint achieves suboptimal results, suggesting a trade-off process between them.


\end{document}